\documentclass[10pt,journal,compsoc]{IEEEtran}
\usepackage{import}

\usepackage{graphicx}
\usepackage{amsmath}
\usepackage{amssymb}
\usepackage{xspace}
\usepackage{graphbox}
\usepackage{tabu}
\usepackage{ragged2e}

\usepackage{import}
\usepackage{multirow}
\usepackage{bm}
\usepackage{makecell}
\usepackage{booktabs}
\usepackage[pagebackref=true,breaklinks=true,letterpaper=true,colorlinks,bookmarks=false]{hyperref}
\usepackage{cite}
\usepackage{rotating}
\usepackage{tabularx}

\usepackage{array}
\usepackage{subfigure}
\usepackage{fixltx2e}
\usepackage{stfloats}
\usepackage{url}
\usepackage{color}
\usepackage{caption}

\usepackage{epsfig}
\usepackage{graphicx}
\usepackage{amsmath}
\usepackage{amssymb}

\usepackage{array}
\usepackage{multirow}
\usepackage{soul}
\usepackage{siunitx}

\usepackage{colortbl}
\usepackage{color}
\usepackage[dvipsnames,table]{xcolor}
\usepackage{tabularx}
\usepackage{hyperref}
\usepackage{wrapfig}
\usepackage{comment}
\usepackage[switch]{lineno}

\usepackage[frozencache,cachedir=.]{minted}
\usepackage{multirow}
\usepackage{colortbl}
\usepackage{titlesec}

\hypersetup{
    colorlinks=true,
    linkcolor=blue,
    filecolor=magenta,      
    urlcolor=cyan,
    pdftitle={Overleaf Example},
    pdfpagemode=FullScreen,
    }

\makeatletter
\g@addto@macro{\endtabular}{\rowfont{}}
\makeatother
\newcommand{\rowfonttype}{}
\newcommand{\rowfont}[1]{
\gdef\rowfonttype{#1}#1\ignorespaces%
}
\makeatother
\makeatletter
\DeclareRobustCommand\onedot{\futurelet\@let@token\@onedot}
\def\@onedot{\ifx\@let@token.\else.\null\fi\xspace}

\def\eg{\emph{e.g}\onedot} 
\def\ie{\emph{i.e}\onedot}

\makeatother

\ifCLASSINFOpdf
\else
\fi
\definecolor{MyDarkBlue}{rgb}{0,0.08,1}
\definecolor{MyDarkGreen}{rgb}{0.02,0.6,0.02}
\definecolor{MyDarkRed}{rgb}{0.8,0.02,0.02}
\definecolor{MyDarkOrange}{rgb}{0.40,0.2,0.02}
\definecolor{MyPurple}{RGB}{111,0,255}
\definecolor{MyRed}{rgb}{1.0,0.0,0.0}
\definecolor{MyGold}{rgb}{0.75,0.6,0.12}
\definecolor{MyDarkgray}{rgb}{0.66, 0.66, 0.66}
\newcommand{\revise}[1]{{#1}}
\newcommand{\red}[1]{\textcolor{red}{#1}}
\newcommand{\blue}[1]{\textcolor{blue}{#1}}

\makeatletter
\DeclareRobustCommand\onedot{\futurelet\@let@token\@onedot}
\def\@onedot{\ifx\@let@token.\else.\null\fi\xspace}
\def\ie{i.e\onedot}
\def\eg{e.g\onedot}

\newcommand{\dataset}{ComPhy\xspace}
\newcommand{\model}{CPL\xspace}

\newcommand{\modelNew}{PCR\xspace}
\newcommand{\modelNewFull}{ Physical Concept Reasoner\xspace}

\newcommand{\datasetNew}{ComPhy-DIV\xspace}
\newcommand{\datasetNewReal}{ComPhy-REAL\xspace}

\newcommand{\Frst}[1]{\textcolor{red}{\textbf{#1}}}
\newcommand{\Scnd}[1]{\textcolor{blue}{\textbf{#1}}}

\begin{document}
\title{Compositional Physical Reasoning of Objects and Events from Videos}

\author{~\IEEEmembership{Zhenfang Chen*}
        ~\IEEEmembership{Shilong Dong*}
        ~\IEEEmembership{Kexin Yi}
        ~\IEEEmembership{Yunzhu Li}
        ~\IEEEmembership{Mingyu Ding}
        ~\IEEEmembership{Antonio Torralba} \\
        ~\IEEEmembership{Joshua B. Tenenbaum}
        ~\IEEEmembership{Chuang Gan}
                
\IEEEcompsocitemizethanks{
\IEEEcompsocthanksitem Z. Chen and S. Dong contribute equally.
\IEEEcompsocthanksitem Z. Chen is with MIT-IBM Watson AI lab.
E-mail: zfchenzfc@gmail.com
\IEEEcompsocthanksitem S. Dong is with New York University. E-mail: shilongdong00@gmail.com
\IEEEcompsocthanksitem K. Yi is with Harvard University. E-mail: kyi@g.harvard.edu
\IEEEcompsocthanksitem Y. Li is with UIUC. E-mail: yunzhuli@illinois.edu
\IEEEcompsocthanksitem M. Ding is with UC Berkeley. E-mail: myding@berkeley.edu
\IEEEcompsocthanksitem A. Torralba and J. B. Tenenbaum  are with MIT. E-mail: \{torralba, jbt\}@mit.edu  
\IEEEcompsocthanksitem C. Gan is with MIT-IBM Watson AI lab and UMass Amherst. E-mail: ganchuang1990@gmail.com   
}
}

\IEEEtitleabstractindextext{%

\begin{abstract}
\justifying
{
Understanding and reasoning about objects' physical properties in the natural world is a fundamental challenge in artificial intelligence. While some properties like colors and shapes can be directly observed, others, such as mass and electric charge, are hidden from the objects' visual appearance.
This paper addresses the unique challenge of inferring these hidden physical properties from objects' motion and interactions and predicting corresponding dynamics based on the inferred physical properties.
We first introduce the Compositional Physical Reasoning (\dataset) dataset.
For a given set of objects, \dataset includes limited videos of them moving and interacting under different initial conditions. The model is evaluated based on its capability to unravel the compositional hidden properties, such as mass and charge, and use this knowledge to answer a set of questions.
{Besides the synthetic videos from simulators, we also collect a real-world dataset to show further test physical reasoning abilities of different models.}
We evaluate state-of-the-art video reasoning models on \dataset and reveal their limited ability to capture these hidden properties, which leads to inferior performance.
We also propose a novel neuro-symbolic framework, \modelNewFull (\modelNew), that learns and reasons about both visible and hidden physical properties from question answering. Leveraging an object-centric representation, \modelNew utilizes videos and the associated natural language to infer objects' physical properties without dense object annotations. Furthermore, It incorporates property-aware graph networks to approximate the dynamic interactions among objects. \modelNew also employs a semantic parser to convert questions into semantic programs, and a program executor to execute the programs based on the learned physical properties and dynamics.
After training, \modelNew demonstrates remarkable capabilities. It can detect and associate objects across frames, ground visible and hidden physical properties, make future and counterfactual predictions, and utilize these extracted representations to answer challenging questions. 
We hope the proposed \dataset dataset and the \modelNew model present a promising step towards more comprehensive physical reasoning in AI systems.
}

\end{abstract}

}
\maketitle

\begin{figure*}[t]
    \centering
    \includegraphics[width=.9\linewidth]{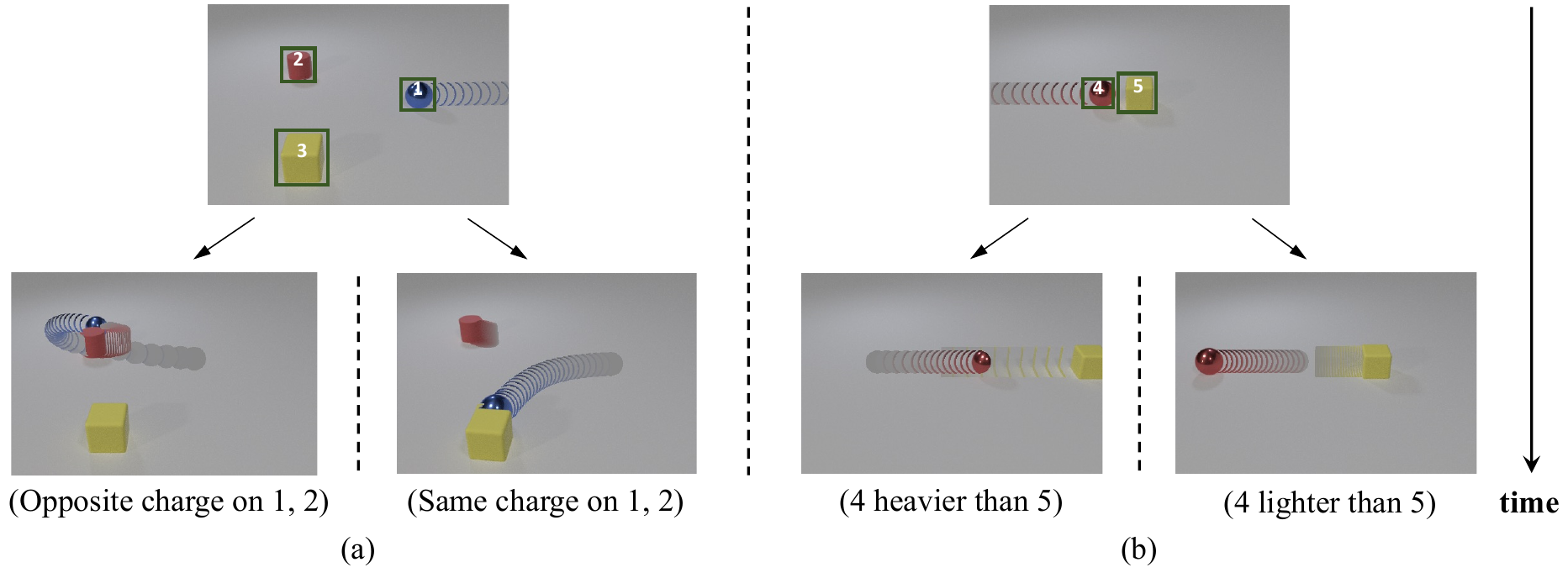}
    \vspace{-1em}
    \caption{Non-visual properties like mass and charge govern the interaction between objects and lead to different motion trajectories. a) Objects attract and repel each other according to the (sign of) charge they carry. b) Mass determines how much an object's trajectory is perturbed during an interaction. Heavier objects have more stable motion.
    } 
    \vspace{-1em}
    \label{fig:teaser}
\end{figure*}

\IEEEraisesectionheading{\section{Introduction}\label{sec:intro}}
\IEEEPARstart{W}{hat} causes apples to float in water while bananas sink? What is the underlying reason for magnets attracting on one side and repelling on the other? Objects in nature frequently manifest complex properties, which delineate their interaction schema within the physical world.
For humans, deciphering these \textit{intrinsic} physical properties often represents pivotal milestones in fostering a more profound and precise comprehension of nature.
The majority of these properties are intrinsic in nature, as they are not readily apparent through objects' static visual attributes and are only detectable from objects' interactions.
Furthermore, these properties influence object motion in a \textit{compositional} manner, where the causal relationships and mathematical laws governing these properties can often be complex. 

As depicted in Figure~\ref{fig:teaser}, various \textit{intrinsic} physical properties, such as charge and inertia, often result in significantly divergent future trajectories.
Objects bearing identical or opposite \textit{charges} will exert either repulsive or attractive forces on one another.
The resultant motion is not only related to the magnitude of the charge each object possesses but also to their respective signs, as illustrated in Figure~\ref{fig:teaser}-(a).
\textit{Inertia} governs the degree of sensitivity of an object's motion to external forces.
In scenarios, where a massive object interacts with a lighter one through attraction, repulsion, or collision, the lighter object experiences more substantial alterations in its motion relative to the trajectory of the massive object, as depicted in Figure~\ref{fig:teaser}-(b).

Recent research has introduced a suite of benchmarks aimed at assessing and diagnosing machine learning systems across a range of physics-related settings~\cite{bakhtin2019phyre,yi2019clevrer,Baradel2020CoPhy}.
These benchmarks present reasoning tasks involving intricate object motion and complex interactions, imposing significant challenges on existing models as they demand an understanding of the underlying physical dynamics to perform well.
However, the majority of complexity in the motion trajectories facilitated by these environments arises from alterations or interventions in the initial conditions of the physical experiments.
The impacts of objects' intrinsic physical properties, along with the distinct challenges they present, hold significant importance for further research.

However, it is non-trivial to construct a benchmark for compositional physical reasoning.
A straightforward approach might involve adhering to the settings established in previous benchmarks~\cite{yi2019clevrer,ates2020craft}, wherein a model is required to observe a video and subsequently respond to questions regarding physical properties. 
Nevertheless, physical properties are intricate and often cannot be comprehensively elucidated within the confines of a single video.
Another approach is to establish correlations between object appearance and physical properties, such as designating all \textit{red spheres} as \textit{heavy}, and subsequently posing questions regarding their dynamics.
Nonetheless, this design may lead models to employ shortcuts by merely memorizing appearances rather than comprehending the interconnected physical properties.

\revise{In this paper, we present an extended version of the ComPhy benchmark~\cite{chen2021grounding}, with significant additions, including more diverse simulated scenes, real-world videos, and new experimental settings. It centers on the comprehension of object-centric and relational physics properties not readily discernible from visual appearances.}
Initially, \dataset presents a limited number of video examples featuring dynamic interactions among objects. Models are tasked with identifying the physical properties of objects and subsequently answering questions pertaining to these properties and their associated dynamics.

\begin{figure*}[t]
\centering
\includegraphics[width =0.9\textwidth]{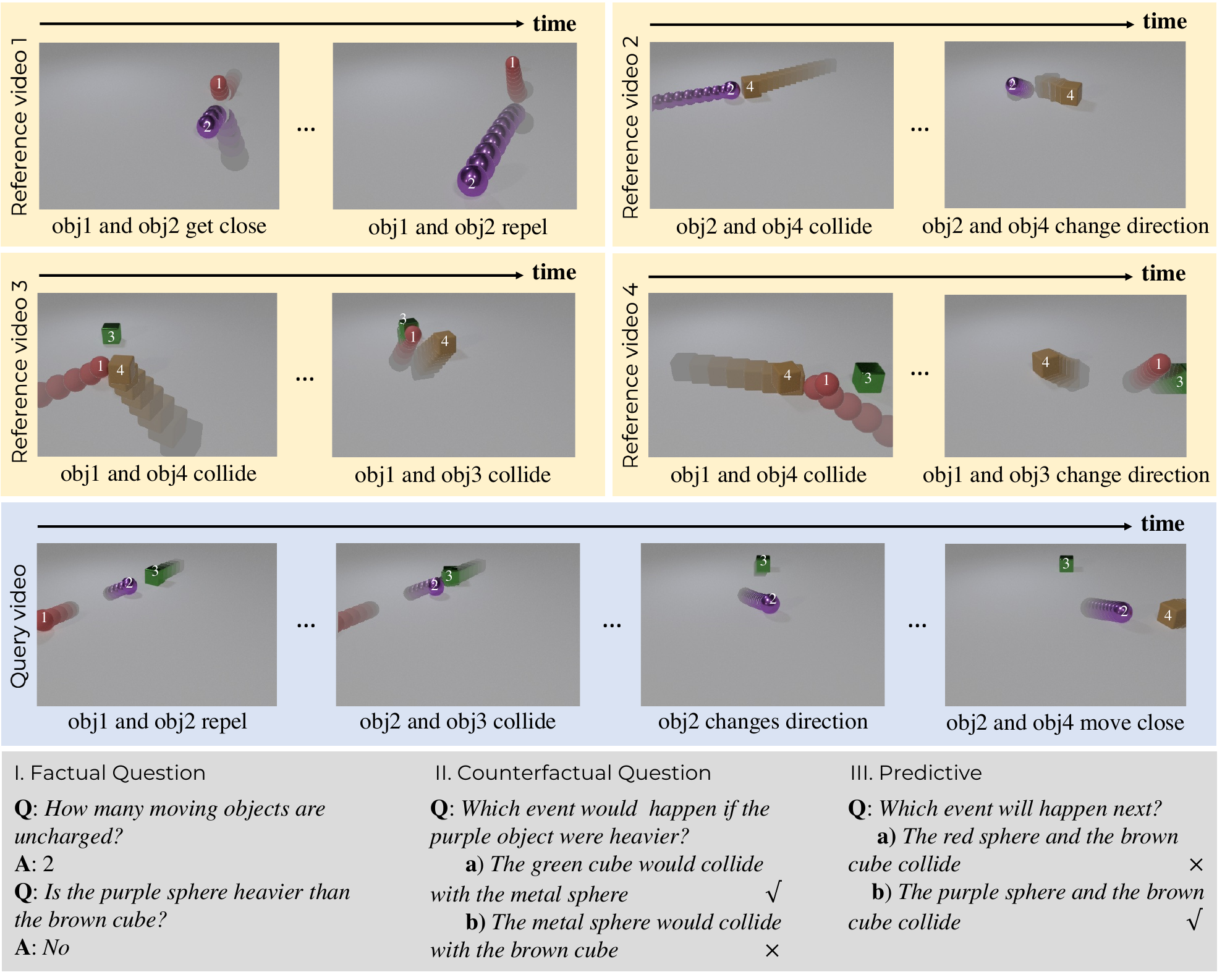}
\vspace{-0.5em}
\caption{Sample target video, reference videos and question-answer pairs from \dataset. \label{fig:2}
}
\vspace{-1em}
\end{figure*}

As depicted in Figure~\ref{fig:2}, the \dataset is composed of meta-train and meta-test sets, with each data point comprising four reference videos and one target video.
In each set, the objects consistently possess the same intrinsic physical properties across all videos.
To facilitate the task, we systematically ensure that each object in the query video appears in at least one of the reference videos.
Reasoning on the \dataset is challenging. First, models must infer both the intrinsic and compositional physical properties of the object set using only a limited number of video samples. Moreover, they must predict video dynamics based on the predicted physical properties.

To overcome the challenges in \dataset, we introduce \modelNewFull (\modelNew). Inspired by recent work on neural-symbolic reasoning on images and videos~\cite{yi2018neural,yi2019clevrer,Mao2019NeuroSymbolic}, our model is modularized with four disentangled components: perception, physical property learning, physical dynamics prediction, and symbolic reasoning.
Our \modelNew model can learn to infer objects' compositional and intrinsic physical properties, predict their future dynamics, and make counterfactual imaginations by only watching videos and reading question-answer pairs.

To summarize, this paper makes the following contributions.
\revise{First, we extend the original \dataset benchmark~\cite{chen2021grounding} by introducing new diverse simulated scenes and real-world video data. It is based on a few-shot reasoning setting that integrates physical properties (mass and charge), physical events (attraction and repulsion), and their compositions.
}
Second, we introduce a new neural-symbolic framework \modelNew, a modularized model that can infer objects' physical properties and predict the objects' movements from watching videos and reading question-answer pairs. 
Additionally, we collect a real-video dataset to better assess the physical reasoning capabilities of current models in real-world scenarios.

\revise{
Some preliminary results were presented in our earlier ICLR 2022 paper~\cite{chen2021grounding}. In this manuscript, we significantly extend that work in three aspects.
}
First, we introduce a \modelNewFull, \modelNew, to learn hidden physical properties like \textit{mass} and \textit{charge} from video and language efficiently without dense property supervision signals during training and perform reasoning in counterfactual and predictive scenes.
Second, besides the experiments in the original data~\cite{chencomphy}, we \revise{also simulate more diverse physical scenes} and collect \textbf{real} videos for physical reasoning. We perform experiments in both synthetic and \textbf{real} videos and analyze how the new proposed \modelNew works and fails, while there are only experiments for synthetic data in the original conference version. Third, we also evaluate recent state-of-the-art large vision-language models (LVLMs)~\cite{achiam2023gpt,team2023gemini} on ComPhy, providing a more thorough analysis. Our code, datasets, and models can be found at \hyperlink{https://physicalconceptreasoner.github.io}{https://physicalconceptreasoner.github.io}.

The rest of the paper is organized as follows. Section~\ref{sec:related} reviews the related datasets and models based on physical reasoning, video question answering, and few-shot learning. Section~\ref{sec:dataset} introduces how we construct the dataset and reduce its biases. Section~\ref{sec:baselines} analyze how representative baselines and the recent state-of-the-art models perform on the \dataset benchmark.
Section~\ref{sec:models} introduces the new \modelNew model and its optimization mechanism. Section~\ref{sec:conclusion} summarizes the paper's contribution, discusses its limitations, and suggests potential extension directions.

\section{Related Work}\label{sec:related}
\vspace{-0.5em}
\begin{table*}[t]
    \centering
    \setlength{\tabcolsep}{3.5pt}
	\begin{tabular}{lcccccccc}
	\toprule
         \multirow{2}{*}{Dataset} & \multirow{2}{*}{Video} & Question & Diagnostic & \multirow{2}{*}{Composition} &   Few-shot  & Physical & {Counterfactual} & Evaluated \\
         &  & Answering  & Annotation &  & Reasoning   & Property & Property Dynamics & on LVLM  \\
         \midrule
         CLEVR~\cite{johnson2017clevr}  & - & \checkmark &  \checkmark &  \checkmark & - & - & - & - \\
         MovieQA~\cite{tapaswi2016movieqa} & \checkmark & \checkmark & - & \checkmark & - & - & - & -  \\
         TGIF-QA~\cite{jang2017tgif}  & \checkmark & \checkmark & - & - & - & - & - & - \\
         TVQA/ TVQA+~\cite{lei2019tvqa} & \checkmark & \checkmark & - & \checkmark & - & - & - & -  \\
         AGQA~\cite{grunde2021agqa} & \checkmark & \checkmark & - & - & - & - & - & - \\
         \midrule
         IntPhys~\cite{riochet2018intphys} & \checkmark & - & \checkmark & - & - & \checkmark & - & -  \\
         PHYRE/ ESPRIT~\cite{rajani2020esprit} & \checkmark & - & \checkmark & \checkmark & - & \checkmark & - & -  \\
         Cater~\cite{riochet2018intphys} & \checkmark & \checkmark  & \checkmark & \checkmark & - & - & - & -  \\
         CoPhy\cite{Baradel2020CoPhy} & \checkmark & - & \checkmark & - & - & \checkmark & - & - \\
         CRAFT~\cite{ates2020craft} & \checkmark & \checkmark & \checkmark & \checkmark & - & - & - & -  \\
         CLEVRER~\cite{yi2019clevrer} & \checkmark & \checkmark & \checkmark & \checkmark & - & - & - & -  \\ 
         Physion~\cite{bear2021physion} & \checkmark & - & \checkmark & - & - & - & - & - \\ 
         Physion++~\cite{tung2023physion++} & \checkmark & - & \checkmark & - & - & \checkmark & - & -  \\ 
         \midrule
         \textbf{\dataset (ours)} & \checkmark & \checkmark & \checkmark & \checkmark & \checkmark & \checkmark & \checkmark & \checkmark \\ 
    \bottomrule
	\end{tabular}
	\caption{Comparison between \dataset and other visual reasoning benchmarks. \dataset is a physical reasoning dataset with a wide range of reasoning tasks for physical property learning and corresponding dynamic prediction.}
	\label{tab:dataset_comparison}
        \vspace{-1em}
\end{table*}
\noindent{\textbf{Physical Reasoning.}}
Our research prominently aligns with contemporary advancements in the domain of physical reasoning benchmarks, as delineated by recent studies~\cite{riochet2018intphys,girdhar2019cater,ates2020craft,tung2023physion++,bear2021physion,zheng2024contphy}. PHYRE~\cite{bakhtin2019phyre} and its variant, ESPRIT~\cite{rajani2020esprit}, establish an environment where objects maneuver within a vertical 2D plane, influenced by gravitational forces. Each task within this framework is tethered to a distinct goal state, and the model seeks resolution by delineating initial conditions conducive to achieving said state. Conversely, CLEVRER~\cite{yi2019clevrer} incorporates videos featuring multiple objects in motion, colliding on a planar surface, and poses natural language questions pertaining to the description, explanation, prediction, and counterfactual reasoning of the resultant collision events. CoPhy~\cite{Baradel2020CoPhy} encompasses experimental trials involving objects moving in 3D space under the influence of gravity, with a focal point on predicting object trajectories following counterfactual interventions upon initial conditions.
~\revise{
CRIPP-VQA~\cite{patel2022cripp} introduces a challenge that emphasizes reasoning over physical properties such as mass and friction from a single video with simple primitive shapes, material and colors. 
}
\revise{
Our work builds upon the original \dataset dataset introduced in our prior work~\cite{chen2021grounding}, extending it with more diverse physical scenes and real-world videos, which requires models to infer physical properties from a few physical interactions in reference videos.
Compared to other previous datasets, \dataset requires models to infer intrinsic properties from a limited array of video examples and draw dynamic predictions based on the identified properties.
}

\noindent{\textbf{Dynamics Modeling.}}
Modeling the dynamics of physical systems has long been a focal point of research. This issue has been explored by some researchers through physical simulations, drawing inferences regarding crucial system- and object-level properties via statistical methodologies such as MCMC~\cite{battaglia2013simulation,hamrick2016inferring,wu2015galileo}. In contrast, others have proposed to directly ascertain the forward dynamics employing neural networks~\cite{lerer2016learning}. Owing to their object- and relation-centric inductive biases and efficacy, Graph Neural Networks (GNNs)\cite{kipf2017semi} have been broadly applied in predicting forward dynamics across a diverse array of systems\cite{battaglia2016interaction,chang2016compositional,sanchez2020learning, li2018learning}. Our research combines the strengths of both approaches: initially inferring the object-centric intrinsic physical properties and subsequently predicting their dynamics predicated on these intrinsic properties.

\noindent{\textbf{Video Question Answering.}}
Our research also pertains to the domain of video question answering, which responds to queries about visual content. Several benchmarks~\cite{wang2024sok,mun2017marioqa,lei2018tvqa} have been posited to address the task of video question answering, such as MarioQA~\cite{mun2017marioqa}, TVQA~\cite{lei2018tvqa}, and AGQA~\cite{grunde2021agqa}. Nevertheless, these datasets primarily concentrate on comprehending human actions and activities rather than acquiring knowledge of physical events and properties, a competency crucial for robotic planning and control.

\revise{
We summarize the differences between our extended \dataset benchmark and other prior physical reasoning datasets in Table~\ref{tab:dataset_comparison}. Compared to our previous version~\cite{chen2021grounding}, this work introduces more diverse simulated scenes and real-world videos. Notably, \dataset remains the only dataset requiring models to infer physical properties from a sparse set of video examples, perform dynamics prediction, and answer compositional reasoning questions.
}

\noindent{\textbf{Few-shot Learning.}}
Our research bears relevance to few-shot learning, which learns to classify images utilizing merely a few examples~\cite{vinyals2016matching,snell2017prototypical,sung2018learning,Han2019Visual}. \dataset mandates that models identify object property labels from a limited selection of video examples. Contrasting with the aforementioned works, reference videos in our approach do not furnish labels for objects' physical properties but exhibit more interactions among objects, thereby providing models with information to discern objects' physical properties.

\vspace{-0.7em}
\section{Dataset}
\vspace{-0.3em}
\label{sec:dataset}
\revise{This section describes the dataset used in our benchmark. We build upon our prior work~\dataset~\cite{chen2021grounding}, originally introduced in ICLR 2022, and present a significantly extended version. In addition to the synthetic split described in~\cite{chen2021grounding}, we enrich the synthetic dataset with more diverse physical scenes and include a new real-world video dataset.
}
First, we introduce video details and the task setup in Section~\ref{sec:video}.
Subsequently, Section~\ref{sec:question} delves into the different categories of questions, while Section~\ref{sec:stat} explores the underlying statistics and ensures balance. Finally, in Section~\ref{sec:real}, we introduce how we build the real-world data set.

\vspace{-0.1em}
\subsection{Videos}
\vspace{-0.1em}
\label{sec:video}
\noindent{\textbf{Objects and Events.}}
Following~\cite{johnson2017clevr}, objects in \dataset are characterized by compositional appearance attributes, including color, shape, and material. For ease of identification, each object in the videos is uniquely distinguishable based on these three characteristics. The dataset incorporates events such as \textit{in}, \textit{out}, \textit{collision}, \textit{attraction}, and \textit{repulsion}. The basic concepts in \dataset are derived from these object appearance attributes, events, and their compositionality.

\noindent{\textbf{Physical Properties.}}
Previous benchmarks~\cite{riochet2018intphys,yi2019clevrer} predominantly focused on visually perceptible appearance concepts like color and collision, discernible in a single frame.  
In contrast, our dataset, \dataset, additionally explores the \textit{intrinsic} physical properties of \textit{mass} and \textit{charge}, which are not directly discernible from an object's static appearance (Figure~\ref{fig:teaser}(a,b)). These properties are independent of visual features and can interact, resulting in more intricate and diverse dynamic scenarios. For simplicity, the dataset categorizes objects into discrete mass groups (\textit{heavy}/ \textit{light}) and charge categories (\textit{positively} / \textit{negatively charged}/ \textit{uncharged}). While introducing additional continuous parameters like bounciness and friction is possible, the complexity could render the dataset overly intricate and hinder intuitive property inference from people.

\noindent{\textbf{Video Generation.}}
Each target video designated for question-answering encompasses 3 to 5 objects, integrating a random compositionality of appearance attributes and physical properties. The videos are standardized to a duration of 5 seconds, with an extended simulation of the 6-th and 7-th seconds specifically for annotating questions for future prediction.

~\revise{
As in our prior work~\cite{chen2021grounding}, the synthetic videos are generated in a two-step process using the Bullet physics engine and rendered via Blender. 
}
In the first step, we employ the Bullet physical engine~\cite{coumans2021} to simulate the movements of objects and their interactions with one another.
Since Bullet does not officially support the effect of electronic charges, we add external forces between charged objects, whose values are inversely proportional to the square of the objects' distance, to simulated Coulomb forces.
We assign a mass value of 1 to the \textit{light} objects and a mass value of 5 to the \textit{heavy} objects.
We manually ensure that every reference video includes at least one interaction, such as collision, attraction, or repulsion, among objects, to provide sufficient information for inferring physical properties.
Every object in the target video must appear in the reference videos at least once.
The simulated object movements are then transmitted to Blender\cite{blender} for high-quality image sequences.

\noindent{\textbf{Task Setup.}}
It presents a non-trivial challenge to design an evaluative framework that accurately assesses a model's capacity for physical reasoning because physical properties are not discernible within a static frame. 
A simplistic approach would involve associating physical attributes directly with object appearances like ``\textit{The red object is heavy}", ``\textit{The yellow object is light}" and then asking \textit{``What would happen if they collide?}"
However, this setting is flawed, as it fails to ascertain whether the model genuinely comprehends the physical properties or merely relies on memorizing visual cues. 
An ideal setup would demand a model to demonstrate human-like discernment of objects' properties from their motion and mutual interactions within dynamic scenes, and subsequently formulate relevant dynamic predictions.

To achieve this goal, We introduce a meta-framework for physical reasoning that pairs a target video with a limited set of reference videos, enabling models to infer physical properties.
Questions are then formulated regarding these properties and underlying dynamics, as illustrated in Figure~\ref{fig:2}. 
Thus, each collection includes a target video, four reference videos, and numerous inquiries related to the target video.
Notably, all objects within each collection maintain consistent visual attributes, including color, shape, and material, as well as intrinsic physical properties, specifically mass and charge.

\noindent{\textbf{Reference Videos.}}
To enrich the visual content for physical property inferring, we supplement each target video with four reference videos.
From the target video, we select 2 to 3 objects, assign them different initial velocities and positions, and orchestrate interactions such as attraction, repulsion, or collision. 
 The reference videos, though lasting 2 seconds each for scalability, follow the same generation criteria as the target videos. 
 These supplementary interactions help models deduce physical properties; for example, observing repulsion in Reference Video 1 of Figure~\ref{fig:2} indicates that \textit{object 1} and \textit{object 2} possess the same electrical charges.

\subsection{Questions}
\label{sec:question}

Inspired by the previous datasets~\cite{johnson2017clevr,yi2019clevrer}, we propose a question engine capable of generating questions that test \textit{factual}, \textit{predictive}, and \textit{counterfactual} reasoning abilities. 

\noindent\textbf{Queries.} 
\textit{Factual questions} are open-ended, requiring concise answers in the form of a single word or short phrase, and assess a model's understanding and reasoning about objects' physical properties, visual attributes, events, and relationships. Building upon existing benchmarks~\cite{yi2019clevrer,ates2020craft}, our dataset (\dataset) introduces novel and challenging factual questions focused specifically on the physical properties of charge and mass (See Figure~\ref{fig:2}~(I)).
Predictive and counterfactual questions, conversely, adopt a multiple-choice format that critically evaluates the plausibility of each provided answer option. 
\textit{Predictive questions} require models to analyze objects' physical properties and dynamics to forecast events in future video frames.
\textit{Counterfactual questions} investigate hypothetical scenarios where an object’s physical properties (\eg, charge or mass) are altered, focusing on their impact on object dynamics (See Figure~\ref{fig:2}~(II)).
This methodology contrasts with prior research~\cite{yi2019clevrer,riochet2018intphys} that centered on object removal, emphasizing the divergent implications of changing physical properties for predicting motion instead.

\noindent\textbf{Templates.}
We present typical question templates in Table~\ref{tb:tp}.
Examining the table reveals that these novel question templates incorporate diverse symbolic operators associated with physical properties.
For example, phrases such as ``\textit{heavy moving spheres}" and ``\textit{charged cubes}" demand that models deduce the values of objects' physical properties.
For counterfactual questions, we introduce novel conditions, such as ``\textit{If the cyan object were uncharged}" and ``\textit{If the sphere were lighter}".
These conditions are designed to enable reasoning about the dynamics when a particular object possesses an alternative physical property.
\begin{table}[t]
\centering
\setlength{\tabcolsep}{1pt}
\resizebox{\linewidth}{!}{
\begin{tabular}{cl}
\toprule
Type & Template and Example \\
\midrule
\multirow{2}{*}{CUN1}                                   &  If the \textit{SA} were \textit{MP}, \textit{Q}? \\
                                       & If the sphere were lighter, which event would not happen? \\
\midrule
\multirow{2}{*}{CUN2}                               &  If the \textit{SA} were \textit{CP}, \textit{Q}? \\
                                       &  If the cube were uncharged, which event would happen? \\
\midrule
\multirow{2}{*}{Mass1} & Is the \textit{DA1} \textit{SA1} heavier than the \textit{DA2} \textit{SA2}? \\
&  Is the blue sphere heavier than the gray cube? \\
\midrule
\multirow{2}{*}{Mass2} & Is the \textit{DA1} \textit{SA1} lighter than the \textit{DA2} \textit{SA2}? \\
&  Is the blue sphere lighter than the gray cube? \\
\midrule
\multirow{2}{*}{CHR1} & Are the \textit{DA1} \textit{SA1} and the \textit{DA2} \textit{SA2} oppositely charged? \\
& Are the blue sphere and the purple sphere oppositely charged? \\
\midrule
\multirow{2}{*}{CHR2} & Are the \textit{DA1} \textit{SA1} and the \textit{DA2} \textit{SA2} with the same type of charge? \\
& Are the cube and the cylinder with the same type of charge?\\
\midrule
\multirow{2}{*}{CHR3} & What are the \textit{Hs} of the two objects that are charged? \\
&  What are the colors of the two objects that are charged? \\
\midrule
\multirow{2}{*}{Query}                                   &  What is the \textit{H} of the \textit{DA} \textit{SA} that is \textit{PA}? \\
                                       &  What is the color of the moving cylinder that is heavy? \\
\midrule
\multirow{2}{*}{Exist}                                   & Are there any \textit{PA}  \textit{DA} \textit{SA} \textit{TI}? \\
                                       & Are there any charged moving cube when the video ends? \\
\midrule
\multirow{2}{*}{Count}                                   & How many \textit{PA} \textit{DA} \textit{SA} are there \textit{TI}? \\
                                       & How many heavy stationary spheres are there?
                                       \\
\bottomrule
\end{tabular}
}
\caption{{{Typical question templates and examples in \dataset. \textit{SA} denotes static attributes like ``\texttt{red}'';  \textit{DA} denotes dynamic attributes, ``\texttt{moving}''; \textit{MP} denotes mass attributes like ``\texttt{heavier}''; \textit{Q} denotes question phrases like ``\texttt{which of the following would happen}''; \textit{CP} denotes charge attributes like ``\texttt{uncharged}''; \textit{H} denotes visible concepts like ``\texttt{material}''; \textit{PA} denotes physical attributes like \texttt{heavy} and \texttt{charged}; \textit{TI} denotes time indicators like ``\texttt{when the video ends}''.}} 
\label{tb:tp}}
\vspace{-1em}
\end{table}

\subsection{Balancing and Statistics}
\label{sec:stat}

In total, \dataset features 8,000 training sets, 2,000 for validation, and 2,000 for testing, with a total of 41,933 factual, 50,405 counterfactual, and 7,506 predictive questions constituting 42\%, 50\%, and 8\% of the dataset, respectively. 
For simplicity, video sets will include a pair of charged objects only if charged objects are already present, and similarly, a video will contain a heavy object or none at all. 
We ensure that these few video examples are sufficiently informative to answer questions based on the questions' programs and the properties and interaction annotations in the videos. 
Specifically, for questions comparing mass or establishing charge relations, we meticulously confirm at least one interaction exhibited between the relevant objects. 

\subsection{Real-World Datasets}
\label{sec:real}
\begin{figure}[t]
    \centering
    \includegraphics[width=\linewidth]{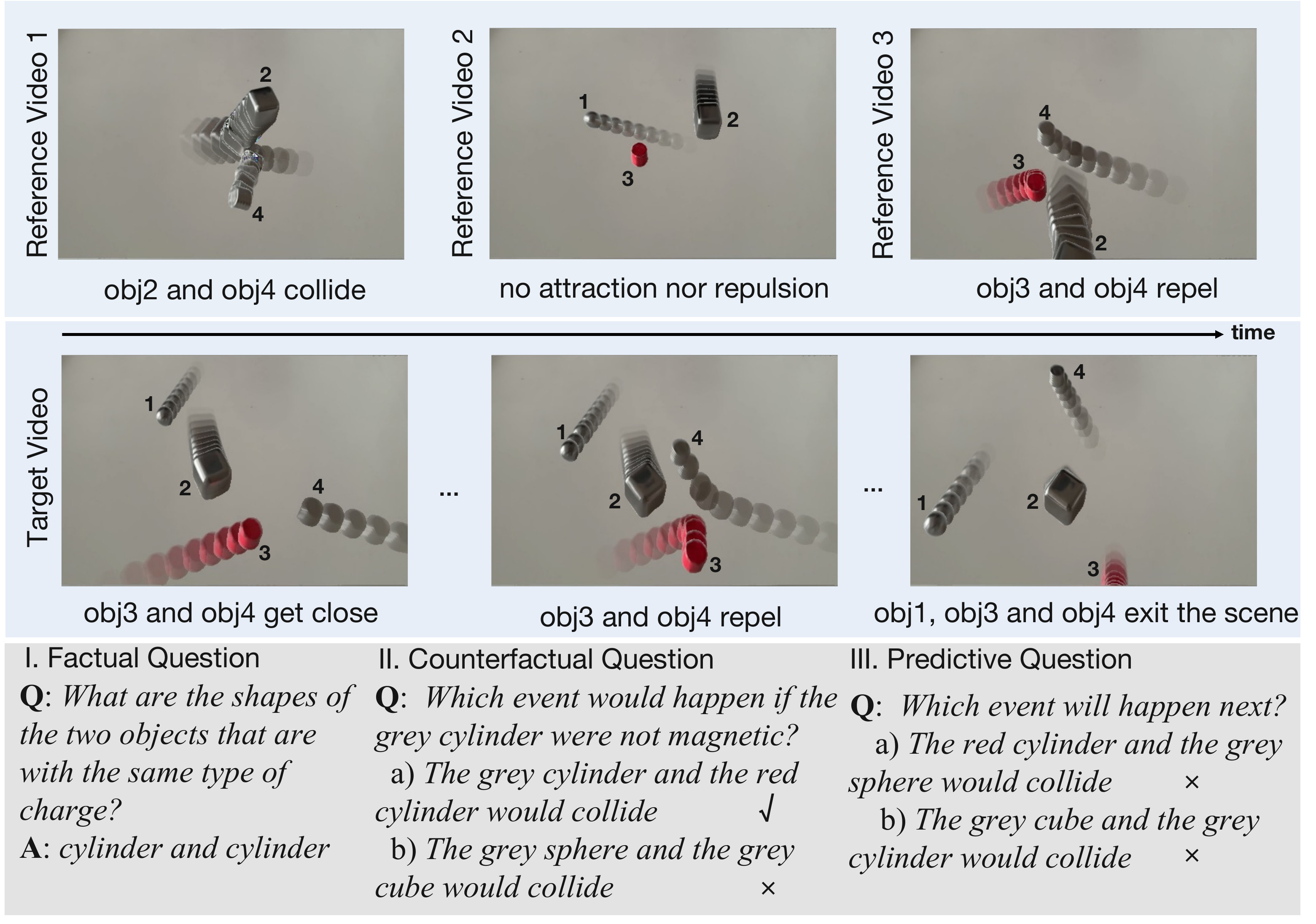}
    \caption{Samples of real data. We collect real objects of different mass values and magnetism for extensive experiments, which have a significant effect on objects' motion and interaction.
    } 
    \label{fig:real}
\end{figure}

\revise{As shown in Fig.~\ref{fig:real}, we collect a new real-world video dataset to further estimate the capabilities of physical reasoning models}. The construction of this dataset involves two key stages: real video collection and question annotation.

\noindent{\textbf{Real Video Collection.}}
We capture a dataset consisting of \revise{492} real-world videos using the iPhone's SLO-MO feature, which records high-definition slow-motion footage at 240 frames per second. 
These videos are organized into \revise{123} sets, with 60 sets designated for training and 20 sets for validation. Each set comprises one target video featuring 3-4 objects interacting and 3 associated reference videos containing 2-3 objects each. This design mirrors the simulated video split, focusing on object interactions characterized by attributes like color (red, brown, grey), shape (cylinder, cube, sphere), magnetism (neutral, attractive, repulsive), and mass (heavy and light) in physical world environment.

\noindent{\textbf{Question Annotations.}}
The static attributes, physical properties, and events in each video are initially annotated by an annotator and subsequently checked by another to ensure their correctness. 
We utilize a question engine similar to the one used in the simulated split to generate diverse questions, including counterfactual, predictive, and various property-based inquiries.
The engine randomly selects from predefined templates and incorporates video annotations to create questions that explore various aspects of physical interaction, such as magnetism's effect on dynamics, the influence of mass, and the objects' static attributes. 
We collect 1,068 questions in total, including 776 for physical properties, 134 for counterfactual reasoning, and 158 for predictive future events.~\revise{We provide more details on real-world videos in the supplementary material.}
\section{Experiments}
\label{sec:baselines}
In this section, we assess baseline models and conduct an in-depth analysis to comprehensively study \dataset.

\subsection{Baselines}
We assess multiple baseline models on \dataset, as displayed in Table~\ref{tb:qa}. These baselines fall into four categories: bias-analysis models~\cite{hochreiter1997long}, video question answering models~\cite{antol2015vqa,le2020hierarchical}, compositional reasoning models~\cite{hudson2018compositional,ding2020object}, and large vision-language foundation models~\cite{li2022align,team2023gemini,achiam2023gpt}. For a comprehensive comparison, we additionally introduce variant models that leverage both the target video and reference videos.

\noindent{\textbf{Biased Analysis Models.}}
The first category of models is bias analysis models.
These models predict answers without relying on visual input and aim to scrutinize the language bias present in \dataset.
In particular, the \textbf{Random} model randomly selects answers based on the question type \revise{and requires no training}.
The \textbf{Frequent} model selects the most frequently occurring answer \revise{in the training set} for each question type, \revise{which requires no training phase.}
\textbf{Blind-LSTM} employs an LSTM~\cite{hochreiter1997long} to encode the question and predict the answers without visual input; \revise{it is trained solely on the question-answer pairs from the dataset's training split to isolate language bias.}

\noindent{\textbf{Visual Question-Answering Models.}}
The second category of models encompasses visual question-answering models.
These models answer questions based on input videos and questions.
The \textbf{CNN-LSTM} model~\cite{VQA} is a simple question-answering model. It employs a ResNet-50~\cite{He_2016_CVPR} to extract frame-level features, averaging them across the time dimension.
We encode questions using the final hidden state from an LSTM~\cite{hochreiter1997long}. The visual features and question embedding are concatenated to make answer predictions with two fully-connected layers.
\textbf{HCRN}~\cite{le2020hierarchical} is a widely adopted model that hierarchically models visual and textual relationships. 
\revise{Both \textbf{CNN-LSTM} and \textbf{HCRN} were trained (or fine-tuned, if using pre-trained components like ResNet) on the training split until convergence on the validation set.
}

\noindent{\textbf{Visual Reasoning Models.}}
The third category, visual reasoning models, includes \textbf{MAC}\cite{hudson2018compositional}, which decomposes visual question answering into several attention-focused reasoning steps, making predictions based on the hidden output of the final step. In contrast, \textbf{ALOE}\cite{ding2020object} capitalizes on transformers\cite{vaswani2017attention} and object-centric representation to deliver cutting-edge results on CLEVRER. We use MONet\cite{burgess2019monet} to extract visual representation for \textbf{ALOE}. 
\revise{
Similar to the VQA models, both \textbf{MAC} and \textbf{ALOE} were trained (or fine-tuned from general pre-trained weights where applicable) on the dataset's training split.
}

\noindent{\textbf{Large Vision Language Models.}}
The final model category is large vision language models~\cite{li2022align,achiam2023gpt,team2023gemini}, which have been trained on massive vision-language data and shown excellent performance on both language understanding and visual question answering. For \textbf{ALPRO}, we fine-tune the model with \dataset's training set until they achieve satisfactory results on the validation set. 
For \textbf{GPT-4V} and \textbf{Gemini}, we evenly sample a fixed number of frames from each target video as visual input, pairing them with corresponding questions and a carefully crafted text prompt to guide the model in generating formatted answers. 

\noindent{\textbf{Baselines with Reference Videos.}}
We also introduce variations of existing baseline models that utilize both the target video and reference videos as input. We enhance \textbf{CNN-LSTM}, \textbf{MAC}, and \textbf{ALOE} to create \textbf{CNN-LSTM (Ref)}, \textbf{MAC (Ref)}, and \textbf{ALOE (Ref)} by incorporating the features of both reference videos and the target video as visual input. We uniformly sample 25 frames from each target video and 10 frames from each reference video.

\noindent{\textbf{Training and Evaluation Fairness.}}
\revise{
To ensure fair comparison, all models that underwent training or fine-tuning (\textbf{Blind-LSTM}, \textbf{CNN-LSTM}, \textbf{HCRN}, \textbf{MAC}, \textbf{ALOE}, \textbf{ALPRO}, and the `Ref' variants) were trained on the same training split. We employed consistent hyperparameter tuning strategies (where applicable) and evaluated all models under identical conditions on the validation/test splits using the specified metrics. The zero-shot evaluation of \textbf{GPT-4V} and \textbf{Gemini} is reported separately and interpreted in light of their lack of dataset-specific fine-tuning.
}

We employ the conventional accuracy metric to assess the performance of various methods. In the case of multiple-choice questions, we provide both per-option accuracy and per-question accuracy. A question is deemed correct if the model answers all of its options correctly.

\begin{table}[t]
\begin{center}
\setlength{\tabcolsep}{2pt}
\begin{tabular}{lccccc}
\toprule
\multirow{2}{*}{Methods}  & \multirow{2}{*}{Factual} & \multicolumn{2}{c}{Predictive} &  \multicolumn{2}{c}{Counterfactual} \\
  &               & per opt.       & per ques.      & per opt.      & per ques.     \\
\midrule
Random                   &  29.7         & 51.9      & 22.6               &    49.7           &    9.1                      \\
Frequent                 &  30.9         &  56.2     &  25.7 &    50.3       &  8.7                         \\
Blind-LSTM               & 39.0          & 57.9          & 28.7      &  55.7           &    12.5  \\
\midrule
CNN-LSTM~\cite{antol2015vqa}                 & 46.6          & 59.5          &  29.8        & 58.6               &  14.6     \\
HCRN~\cite{le2020hierarchical}              & 47.3          &  62.7         &   32.7       & 58.6               &  14.2   \\
\midrule
MAC~\cite{hudson2018compositional}          &   \Frst{68.6}        & 60.2      &  32.2    &    60.2           &    16.0   \\
ALOE~\cite{ding2020object}                  &   54.3        &   \Scnd{65.9}    &    35.2  &     \Scnd{65.4}  & \Scnd{20.8}      \\
\midrule
CNN-LSTM (Ref)~\cite{antol2015vqa}          &  {41.9}   &      {59.6}     &  {29.4}  &  {57.2}     &      {12.8}  \\
MAC (Ref)~\cite{hudson2018compositional}    & \Scnd{65.8}  & {60.2} &  {30.7} &   {60.3} & {14.3} \\
ALOE (Ref)~\cite{ding2020object}            & 57.7 & \Frst{67.9} & \Scnd{37.1} & \Frst{67.9} & \Frst{22.2} \\
\midrule
ALPRO~\cite{li2022align}                    & {45.2} & {56.9} & {27.2} & {53.7} & {14.4} \\
GPT-4V~\cite{achiam2023gpt}                 & {42.2} & {60.7} & \Frst{47.1} & {51.1} & {8.9} \\
Gemini~\cite{team2023gemini}                & {37.7} & {46.5} & {22.7} & {49.2} & {6.3} \\
\midrule
Human Performance                           & {90.6} & {88.0} & {75.9} & {80.0} & {52.9} \\
\bottomrule
\end{tabular}
\end{center}
\vspace{-1em}
\caption{Evaluation of physical reasoning on \dataset. {Human performance is based on sampled questions. See Section~\ref{sec:eval} for more details. \Frst{Red} text and \Scnd{blue} text indicate the first and the second best results.}}
\label{tb:qa}
\end{table}

\subsection{Evaluation on physical reasoning} 
\label{sec:eval}
The question-answering results of various baseline models are shown in Table~\ref{tb:qa}. Notably, there exist discrepancies in the relative performances of models across different kinds of questions, which suggests that diverse reasoning skills are necessitated by the questions in \dataset.

\noindent{\textbf{Factual Reasoning.}}
To address factual questions in \dataset, models must identify visual attributes, analyze motion trajectories, and infer physical properties of objects.
The results indicate that the ``blind" models, namely \textbf{Random}, \textbf{Frequent}, and \textbf{Blind-LSTM}, perform significantly poorly on \dataset compared to other models integrating visual context and linguistic information.
Additionally, we observe that video question-answering models and pre-trained large vision language models exhibit lower performance compared to visual reasoning models like \textbf{MAC} and \textbf{ALOE}.
We attribute this discrepancy to the fact that they are typically tailored for tasks such as object classification, action recognition, and activity understanding rather than understanding physical events in ComPhy.
Among these, \textbf{MAC} outperforms the rest baselines when answering factual questions, underscoring the effectiveness of its compositional attention mechanism and iterative reasoning processes. 

\noindent{\textbf{Dynamcis Reasoning.}}
A notable feature of \dataset is its demand for models to generate counterfactual and future dynamic predictions by leveraging their identified physical properties to address posed questions.
Among all the baseline models, we have observed that \textbf{ALOE (Ref)} consistently attains the highest performance levels in tasks involving counterfactual and future reasoning.
We posit that this superior performance is attributable to the utilization of self-attention mechanisms and self-supervised object masking techniques, enabling the model to effectively capture spatio-temporal visual context and imagine counterfactual scenarios for answering questions.

\noindent{\textbf{Reasoning with Large Vision-Language Models.}}
We also evaluate the performance of the recent large vision-language models, \textbf{ALPRO}, \textbf{Gemini} and \textbf{GPT-4V} on \dataset, which were pre-trained on massive image/video-text pairs from the internet. 
Despite their strong performance on traditional visual question-answering benchmarks such as GQA~\cite{hudson2019gqa}, VQAv2~\cite{VQA}, MSRVTT-QA~\cite{xu2017video} and MSVD-QA~\cite{xu2017video}, all of them underperform on \dataset.
The inferior performance of large vision-language models (LVLMs) is attributed to a gap in their training. These models are pretrained on internet data, which primarily focuses on object categories and semantic relations, lacking emphasis on physical commonsense. 
\dataset underscores its value in addressing the gap and complementing the missing physical commonsense in existing visual question-answering benchmarks.

\noindent{\textbf{Reasoning with Reference Videos.}}
The results reveal that \textbf{CNN-LSTM (Ref)} and \textbf{MAC (Ref)} perform comparably or slightly worse than their original counterparts, \textbf{CNN-LSTM} and \textbf{MAC}.
While \textbf{ALOE (Ref)} shows a modest improvement over \textbf{ALOE}, the variant models do not exhibit substantial improvements when incorporating the reference videos as supplementary visual input.
This phenomenon is likely due to these models' primary training on extensive datasets comprising videos and question-answer pairs, hindering their adaptability to \dataset's novel context, which necessitates discerning new compositional visible and hidden physical properties from a limited number of examples.

\noindent{\textbf{Human Performance.}} 
To evaluate human performance in \dataset, 14 participants with a basic understanding of physics and proficiency in English were tasked.
After an initial warm-up through a series of demonstration videos and questions to confirm their comprehension of events and physical properties, they were assigned to answer 25 diverse question samples from \dataset.
Their accuracy rates are as follows: 90.6\% for factual questions, 88.0\% for predictive questions per option, 80.0\% for counterfactual questions per option, 75.9\% for predictive questions per question, and 52.9\% for counterfactual questions per question.

\begin{table}[t]
\begin{center}
\setlength{\tabcolsep}{2pt}
\begin{tabular}{lccccc}
\toprule
\multirow{2}{*}{Methods}  & \multirow{2}{*}{Factual} & \multicolumn{2}{c}{Predictive} &  \multicolumn{2}{c}{Counterfactual} \\
             &               & per opt.       & per ques.      & per opt.      & per ques.     \\
\midrule
Random      &7.6 &50.0 &25.0 &50.9 &20.8               \\
Frequent    &41.7 &53.6 &28.7 &50.0 &23.9                                  \\
Blind-LSTM  &50.6  &61.5 &46.0 &51.9 &32.2            \\

\midrule
CNN-LSTM~\cite{antol2015vqa}    &55.6  &64.2 &47.3 &50.9 &33.3           \\
HCRN~\cite{le2020hierarchical}  &51.9  &62.5 &53.5 &50.9 &32.1               \\
\midrule
MAC~\cite{hudson2018compositional}      &58.9  &60.9 & {57.1} & {52.8} & {35.8}        \\
ALOE~\cite{ding2020object}         &60.8  &60.6 &42.4 &47.1 &28.7             \\
\midrule
{CNN-LSTM (Ref)}~\cite{antol2015vqa}      &49.0  & {64.3} &41.3 &50.0 &26.3           \\
{MAC (Ref)}~\cite{hudson2018compositional}    &56.4  &56.2 &46.4 &51.4 &34.9         \\
ALOE (Ref)~\cite{ding2020object}      & {61.6}  &61.4 &42.8 &51.6 &32.1          \\
\midrule
ALPRO~\cite{li2022align}  &50.9  &55.3 &39.2 &49.7 &29.2  \\
GPT-4o-mini~\cite{achiam2023gpt} &42.6  &49.6 &23.2 &47.5 &26.0        \\
Gemini~\cite{team2023gemini} &32.5  &57.7 &23.1 &52.1 &29.8   \\
\midrule
Human Performance   &90.0  &95.0 &90.0 &94.4 &88.9     \\
\bottomrule
\end{tabular}
\end{center}
\vspace{-1em}
\caption{\revise{Evaluation of physical reasoning on the real video. {Human performance is based on sampled questions.}}}
\vspace{-1em}
\label{tb:real}
\vspace{-1em}
\end{table}

\noindent{\textbf{Reasoning in the Real World.}}
\revise{
We evaluated the performance of various models on our collected real-world dataset by fine-tuning each model on the dataset's training split and evaluating on the validation split (see Table~\ref{tb:real}). Results indicate that ALOE achieves the highest accuracy on factual questions (61.6\%), consistent with its strong performance observed in simulated scenarios. Notably, MAC shows a balanced performance across all question types, particularly excelling in predictive questions (57.1\% per question accuracy). Interestingly, state-of-the-art general-purpose vision-language models such as GPT-4o-mini and Gemini significantly underperform compared to specialized models, reflecting substantial limitations in their ability to reason about physical interactions in real-world contexts. The substantial gap between human performance (exceeding 88\% across all categories) and the evaluated models underscores the complexity and challenge of physical reasoning tasks. 
}
\section{Models}\label{sec:models}
\begin{figure*}[t]
    \includegraphics[width=\linewidth]{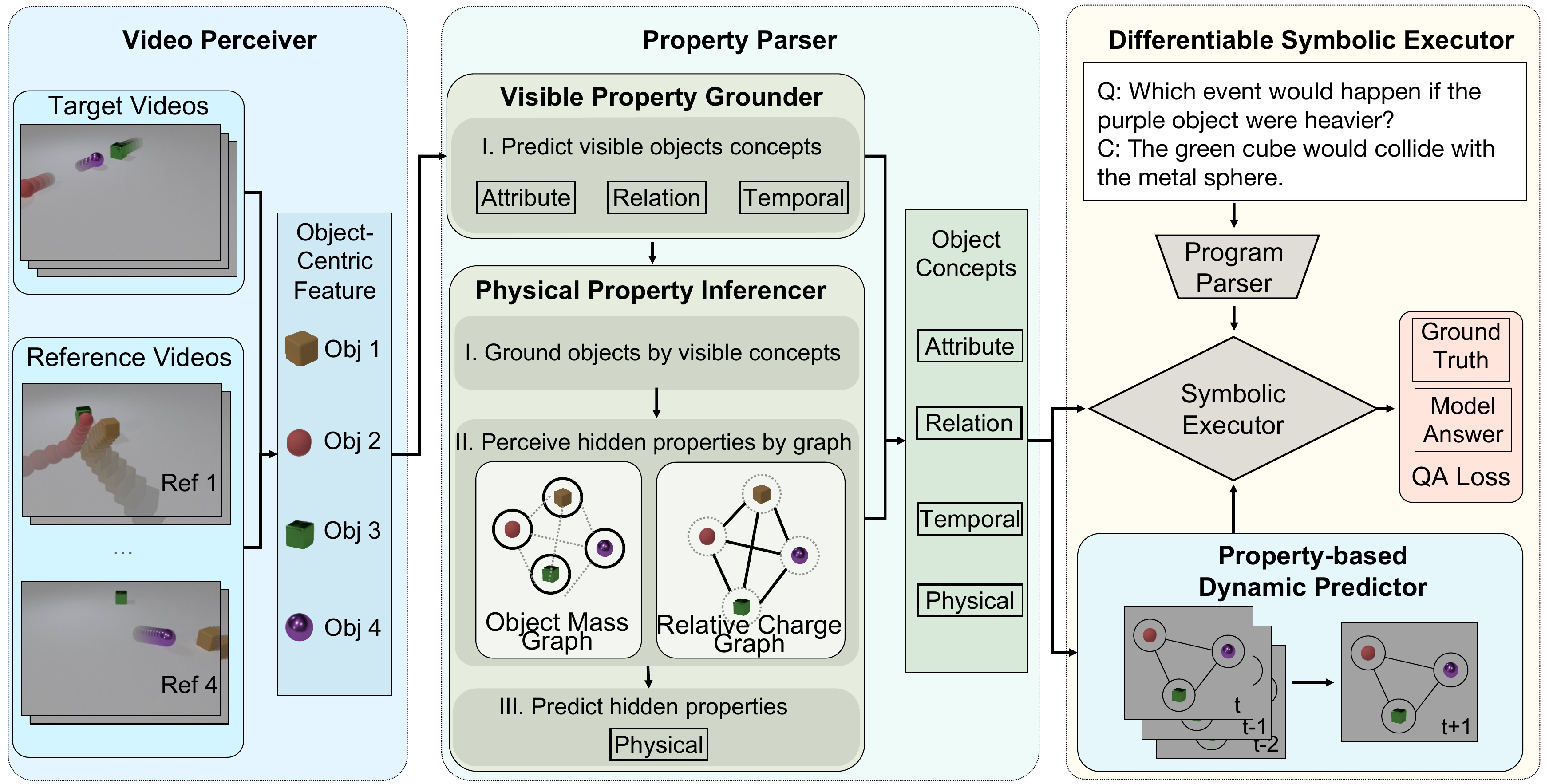}
    \caption{The perception module detects objects' location and visual appearance attributes. The physical property learner learns objects' properties based on detected object trajectories. The dynamic predictor predicts objects' dynamics in the counterfactual scene based on objects' properties and locations. Finally, an execution engine runs the program parsed by the language parser on the predicted dynamic scene to answer the question.} 
    \vspace{-1em}
    \label{fig:model}
\end{figure*}

\subsection{Model}
\label{subsec:models}
In this section, we present \modelNewFull (\modelNew), a new physical reasoning model.
It aims to comprehend objects' visible properties, infer hidden physical properties and events, and image corresponding physical dynamics by observing the videos and responding to the associated questions. Compared with our preliminary models ~\cite{chen2021grounding,chencomphy}, it is able to infer hidden physical properties and predict corresponding property-based dynamics without explicit dense property annotations.

\modelNew can be factorized into different functional modules for physical reasoning in videos. 
As shown in Fig.~\ref{fig:model}, the model consists of five major modules: (1) video perceiver, (2) visible property grounder, (3) physical property inferencer, (4) property-based dynamic predictor, and (5) differentiable symbolic executor.
When provided with a target video alongside four reference videos and a query, \modelNew employs a video perceiver to detect objects' spatial locations across frames and all videos. Subsequently, their trajectories are processed by the physical property inferencer to deduce their properties. Leveraging these data, the dynamic predictor forecasts object movements based on their physical traits. Lastly, a differentiable executor executes the program generated by a language parser~\cite{yi2018neural,vaswani2017attention}, utilizing the predicted object motions to answer the query. Note that the object-centric representation and outputs of various modules are maintained in a differentiable manner, enabling direct optimization of each module through backpropagation when answering video-related questions.

\subsubsection{Video Perceiver}
\noindent{\textbf{Object Tracking and Alignment.}}
Given a target video and 4 reference videos, the video perceiver in \modelNew is responsible to track objects in every video and align them across different videos. The first step is to track objects in the videos. At the $t$-th frame, our model first applies a regional proposal network~\cite{he2017mask,ren2015faster} to detect all objects $\{b_i^t\}_{i=1}^{N_t}$, where $N_t$ denote the object proposal number. The video perceiver then get a set of object trajectories $\{o_n\}_{n=1}^N$, where $N$ is the number of object trajectories, $o_n=\{b^t\}_{t=1}^T$ and $T$ is the number of frames. Similar to~\cite{chen2021grounding,chen19acl}, we first define the connection score $s_{cnn}(b^t_i, b^{t+1}_j)$ between two proposals $b_i^t$ and $b_j^t$ in connective frames as
\begin{equation}
    s_{cnn}(b^t_i, b^{t+1}_j) = s_{c}(b^t_i) + s_{c}(b^{t+1}_j) + \text{IoU}(b^t_i, b^{t+1}_j),
\end{equation}
where $s_c(b^t_i)$ is the confidence score predicted by the region proposal network and $\text{IoU}$ denotes the intersection over union between two proposals. We define the connection score of a candidate object trajectory $o_n$$=$$\{b^t_n\}_{t=1}^T$ as $E(o_n)=\sum_{t=1}^{T-1}s_{cnn}(b^t_n, b^{t+1}_n)$. We select the set of object trajectories $\{o_n\}_{n=1}^N$ with the highest connection scores and solve the problem with a linear sum assignment~\cite{munkres1957algorithms}. 
We then align objects in reference videos to the target videos with the predicted static visual attributes, color, shape, and material. Objects in reference videos are assigned to objects in the target video that have the most similar predicted labels.

\noindent{\textbf{Object-Centric Representation.}}
We use a set of object-centric features to represent the videos for physical reasoning. Specifically, we compute the \textbf{averaged visual regional features} ($\textbf{f}^v_n \in \mathbf{R^{D_v}}$) from the faster-RCNN~\cite{ren2015faster} for static visual appearance attributes like \textit{shape}, \textit{color} and \textit{material}, where $D_v$ equals to $512$ and is the regional feature's dimension from ResNet-34. We adopt the \textbf{temporal trajectory features} ($\textbf{f}^{t}_n \in \mathbf{R^{\times D_t}}$) for predicting temporal concepts like \textit{in} and \textit{out}, where $D_s$ $=$ $T \times 4$ is the concatenation of the object location $b_n^t$ across all $T$ frames.
Since we can only infer objects' physical property values from their movement and interaction, we use a set of \textbf{aligned trajectory features} for physical property inference. For the $n$-th object in the target video, we represent it with $\mathbf{p}_n$ and $\{\mathbf{p}_{n,r}\}_{r=1}^R$, where $\mathbf{p}_{n,r}^t$ and $\textbf{p}_{n,r}$ are the concatenation of the object coordinates $(x_n^t, y_n^t)$ along all $T$ frames in the target video and the $r$-th reference video. $R$ equals to $4$ and is the number of the reference videos. We add all the objects without appearance in the specific reference videos with zero vectors. 

We use the \textbf{interaction feature} ($f^{int}_{i,j,t} \in \mathbf{R}^{D_{int}}$) for prediction the the collision event between the $i$-th and the $j$-th objects at the $t$-th frame.
 we define $f^{int}_{i, j, t}=f^u_{i, j, t} || f_{i, j, t}^{loc}$, where $f^u_{i, j, t}$ is the ResNet feature of the union region of the $i$-th and $j$-th objects at the $t$-th frame and $f^{sp}_{i, j, t}$ is a spatial embedding for correlations between bounding box trajectories.
We define $f^{sp}_{i, j, t}=\text{IoU}(s_{i,t}, s_{j,t})||(s_{i,t}-s_{j,t})||(s_{i,t}\times s_{j,t})$, where 
$s_{i,t}= ||_{k=t-2}^{t+2}{b_i^k}$ is the concatenated segment of the $i$-th object centering at the $t$-th frame.
It concatenates the intersection over union ($\text{IoU}$), difference ($-$), and multiplication ($\times$) of the normalized trajectory coordinates for the $i$-th and $j$-th objects centering at the $t$-th frame. For the collision event in the future and counterfactual scenes, we predict the collision event based on the $f_{i,j,t}$ only since there is no RGB image for extracting the $f^u_{i,j,t}$ feature. 

\vspace{-0.5em}
\subsubsection{Visible property grounder}
\label{subsub:vpg}
The visible property grounder grounds objects' visible properties like \textit{color}, \textit{shape}, and \textit{collision} onto the objects extracted by the video perceiver. \modelNew accomplishes this by aligning the representations of objects and events with learned concept embeddings in \modelNew. For example, to predict whether the $n$-th object is \textit{red} or not, we use a confidence score $s^{red}_n$. We define $s^{red}_n =\left[\cos(c^\text{red}, m_{sa}(f^v_n)) -\delta\right]/\lambda$, 
where $c^{red}$ is a vector, representing the concept \textit{red}, $m_{sa}$ is a fully-connected layer, mapping the object feature $f_n^v$ to the color space. $cos$ calculates the cosine similarity between the two vectors. $\delta$ and $\lambda_{sa}$ are constant scalars, controlling the value range of $s_{red}$. Similarly, we predict two objects collide at the $t$-th frame with $s_{i,j,t}^{cl}$, where $s_{i,j,t}^{cl}$ equals to $s^{cl}_{i,j,t} =\left[\cos(c^\text{cl}, m_{cl}(f^{int}_{i,j,t})) -\delta\right]/\lambda$. $c^\text{cl}$ represents the concept vector for the \textit{collision} event and $m_{cl}$ is a fully-connected layer transforming $f^{int}_{i,j,t}$ into the desired space.

\subsubsection{Physical Property Inferencer}
\label{subsub:ppi}
At the heart of our model, the Physical Property Inferencer (PPI) handles intricate and
composite physical interactions by analyzing object motion trajectories extracted from both reference and target videos. 
The PPI utilizes a graph neural network~\cite{kipf2018neural} to predict mass and relative charge for each object pair, where node features capture object-centric properties (such as mass), and edge features encode pairwise properties (such as relative charge).
The PPI employs a series of message-passing operations on the input trajectories of N objects within the video. The process is described by:

\begin{equation}
    \small
    \begin{aligned}
        \mathbf{v}_{n}^0 = &f_{emb}(\mathbf{p}_n^t),~~~~ \mathbf{e}_{n_1,n_2}^l = f_{rel}^l(\mathbf{v}_{n_1}^l, \mathbf{v}_{n_2}^l), \\
        \mathbf{v}_{n_1}^{l+1} &= f_{enc}^l(\sum_{n_1 \neq n_2} \mathbf{e}_{n_1,n_2}^l),
    \end{aligned}
\end{equation}

Here, $f_{(...)}$ are functions implemented by fully-connected layers. 
We then use two fully-connected layers to predict the output mass label $f_{v}^{pred}(\mathbf{v}_i^2)$ and edge charge label $f_{e}^{pred}(\mathbf{e}_{i,j}^1)$, respectively. 
Notably, the PPI is not trained in a fully-supervised manner but is optimized via leveraging the gradients from differentiable question answering.

The complete physical property of a set of videos can be represented as a fully connected property graph, where each node corresponds to an object that appears in at least one video within the set. Meanwhile, each edge indicates whether the connected nodes possess the same, opposite, or no relative charge ( \ie it signifies whether one or both objects are charge-neutral). 
In Figure~\ref{fig:model}, we illustrate that the physical property inferencer (PPI) independently predicts the objects' properties in each reference video, covering only part of the property graph. To align predictions across different nodes and edges, we utilize the static attributes of objects identified by the video perceiver. 
By aggregating the sub-graphs generated from each video in the set through max-pooling over nodes and edge predictions, we obtain the complete object properties graph. 

\vspace{-0.5em}
\subsubsection{{Property-based Dynamic Predictor}}
To predict objects' positions at the t + 1 frame, based on their full trajectories and properties (mass and charge) at the $t$-th frame, we employ a dynamic predictor 
For the $n$-th object at the $t$-th frame, we represent it with $\mathbf{o}_n^{t,0}=||_{t-3}^t(x_n^t, y_n^t, w_n^t,, h_n^t, m_n)$, using a concatenation of its object location ($x_n^t, y_n^t$), size ($w_n^t, h_n^t$) and the mass prediction ($m_n$) by the Physical Property inferener over a history window of $3$. 
By incorporating a history of object locations rather than solely relying on the location at the $t$-th frame, we encode object velocity and accommodate for perception errors.
Specifically, we have
\begin{equation}
    \small
    \begin{aligned}
        \mathbf{h}_{n_1,n_2}^{t,0} & = \sum_k z_{n_1,n_2,k}g_{emb}^k(\textbf{o}_{n_1}^{t,0}, \textbf{o}_{n_2}^{t,0}), \\[-5pt]
        \mathbf{o}_{n_2}^{t, l+1} & = \mathbf{o}_{n_2}^{t,l} + g_{rel}^l\left(\sum_{n_1}^{n_1 \neq n_2}(\mathbf{h}^{t,l}_{n_1,n_2})\right), \\[-5pt]
        \mathbf{h}_{n_1,n_2}^{t,l+1} & = \sum_k z_{n_1,n_2,k}g_{enc}^{k,l}([\textbf{o}_{n_1}^{t,l+1},\textbf{o}_{n_1}^{t,0}], [\textbf{o}_{n_2}^{t,l+1}, \textbf{o}_{n_2}^{t,0}]),
    \end{aligned}
\end{equation}
where the variable $k \in {0, 1, 2}$ represents whether the two connected nodes carry the same, opposite, or no relative charge.
The $k$-th element of the one-hot indication vector $\mathbf{z}{n_1,n_2}$ is denoted as $\mathbf{z}_{n_1,n_2,k}$.
The message-passing steps are indicated by $l \in [0,1]$ and functions $g_{(...)}$ are implemented through fully-connected layers.
For predicting object location and size in the $(t+1)$-th frame, we employ a function comprising a single fully-connected layer, $g_{pred}(\mathbf{o}_{n_2}^{t, 2})$.

To forecast future frames for predictive questions, we initialize the dynamic predictor with the last three frames of the target video and iteratively predict subsequent frames by feeding the generated predictions back into the model. For counterfactual queries, we use the first three frames of the target video as input, updating the predicted objects' mass labels ($m_i$) and the corresponding one-hot indicator vector $\mathbf{z}_{i,j}$ accordingly, to obtain physical predictions with counterfactual properties labels.

\subsubsection{Differentiable Symbolic Executor}
The differentiable symbolic executor first adopts a program parser~\cite{yi2018neural,bahdanau2014neural} to transform the input question into a series of program operations. The program parser is trained in a fully-supervised manner as in~\cite{chen2021grounding,yi2018neural}. The executor then executes the symbolic operations on the latent object-centric representation derived from the other modules and the output of the final operator serves as the solution to the question.
We adopt a probabilistic approach, similar to the methodology proposed in~\cite{Mao2019NeuroSymbolic}, to represent the object states, events, and results of all operators during the training phase. This probabilistic representation allows for a differentiable execution process, considering the latent representations derived from both the observed and predicted scenes. As shown in the dotted lines of Figure~\ref{fig:model}, it becomes feasible to optimize the video perceiver,  visible property grounder, physical property inferencer, and property-based dynamic predictor within the symbolic execution procedure.

\vspace{-1em}
\subsection{Training Mechanisms}
The proposed \modelNew features multiple functional modules, and optimizing these modules presents great challenges due to several factors: 1) the lack of dense property annotations for both visible concepts and hidden physical properties, 2) the complexity of physical properties and their interaction with other visible properties, and 3) fewer training examples compared to the previous physical reasoning dataset CLEVRER.
To address these challenges, we propose two novel training mechanisms for model optimization: 1) Curriculum Learning for Physical Reasoning in Section~\ref{training:curr}, and 2) Learning by Imagination in Section~\ref{training:imagine}.

\vspace{-0.5em}
\subsubsection{Curriculum Learning for Physical Reasoning}
\label{training:curr}
We design a novel curriculum learning mechanism to optimize the \modelNew introduced in Section~\ref{subsec:models}. 
We first train a program parser to parse the question and answers into executable programs with a sequence-to-sequence model~\cite{bahdanau2014neural}. 
In lesson 1, we filter out and select the factual questions without physical property description to learn an initial model to ground visible properties like \textit{colors}, \textit{shapes}, and \textit{collisions}. 
In lesson 2, we include all the factual questions to teach the model to infer objects' physical properties with the physical property inferencer. During this lesson, we align the objects' dynamics in different videos and property predictions with the static visible property label prediction from lesson 1. 

In lesson 3, we utilize the property prediction results from the last lesson as pseudo labels to train a property-based dynamic predictor, which predicts objects' dynamics in the counterfactual and predictive scenes. 
Finally, we fine-tune all components in an end-to-end manner with all question-answer pairs from the training set.       

\vspace{-0.5em}
\subsubsection{Learning by Imagination}
\label{training:imagine}
One key challenge for the \modelNew on the \dataset dataset is the complexity of its video scenarios compared to previous datasets like CLEVRER~\cite{yi2019clevrer}, which has 152,572 question-answer pairs, while \dataset has only 55,764 pairs but with more variance in physical property variances. 
To improve the training efficiency, we introduce a new training mechanism, named \textit{Learning by Imagination}. 
Specifically, when a counterfactual question states ``\textit{Which event would happen if the purple object were heavier?}", it implicitly indicates that "\textit{there is a purple object}" and ``\textit{the purple object is \textbf{not} heavy}". 
These implicit statements can be transformed into executable programs to enhance the learning of both the visible property grounder and the physical property inference introduced in Section~\ref{subsub:vpg}. Note that the ability to learn and reason in counterfactual situations is a hallmark of human thought~\cite{van2015cognitive,buchsbaum2012power}.

\begin{table*}[t]
\small\setlength{\tabcolsep}{0.2em}
\parbox{.48\linewidth}{
 \centering
    \setlength{\tabcolsep}{2pt}
    \begin{tabular}{lccccccc}
    \toprule
    \multirow{2}{*}{Methods} & \multirow{2}{*}{Factual} & \multicolumn{2}{c}{Predictive} &  \multicolumn{2}{c}{Counterfactual} \\
    \cmidrule(lr){3-4}\cmidrule(lr){5-6}
     &      & per opt.       & per ques.      & per opt.      & per ques. &         \\
    \midrule
    {\model-DPI}~\cite{li2018propagation}   &  -           & {73.3}           &  {50.8}   &   {61.1}         & {16.6}      \\
    \model~\cite{chencomphy}           &   \textbf{80.5}        & 75.3           &  56.4    &   68.3        & \textbf{29.1}       \\
   \midrule
    {\modelNew}  &  76.0    &    \textbf{80.0}        &  \textbf{62.0}      &   \textbf{70.0}         &  29.0     \\    
    \bottomrule
    \end{tabular}
     \vspace{-1em}
    \caption{Evaluation of \modelNew on the test set of \dataset. The best performance is in boldface. \label{tb:nsdr}}
}
\parbox{.48\linewidth}{
    \centering
    \setlength{\tabcolsep}{2pt}
    \begin{tabular}{lccccccc}
    \toprule
    \multirow{2}{*}{Methods}  & \multirow{2}{*}{Factual} & \multicolumn{2}{c}{Predictive} &  \multicolumn{2}{c}{Counterfactual} \\
    \cmidrule(lr){3-4}\cmidrule(lr){5-6}
     &        & per opt.       & per ques.      & per opt.      & per ques. &         \\   
    \midrule
    \modelNew w/o R &  68.7   &   52.0         &  28.1 &    54.9        &  28.0       \\
    \modelNew w/o CI   &  70.3    &    51.2        &  24.4      &   54.0         &  28.0  \\
    \midrule
    {\modelNew}  &   \textbf{78.3}        &  75.0        &  56.5   &     70.5       &  50.2     \\
    \bottomrule
    \end{tabular}
    \vspace{-1em}
    \caption{Ablation study of \modelNew on the validation set of \dataset. The best performance is in boldface. \label{tb:abs}}
}
\end{table*}

\begin{figure*}[t]
    \centering
    \includegraphics[width =1\textwidth]{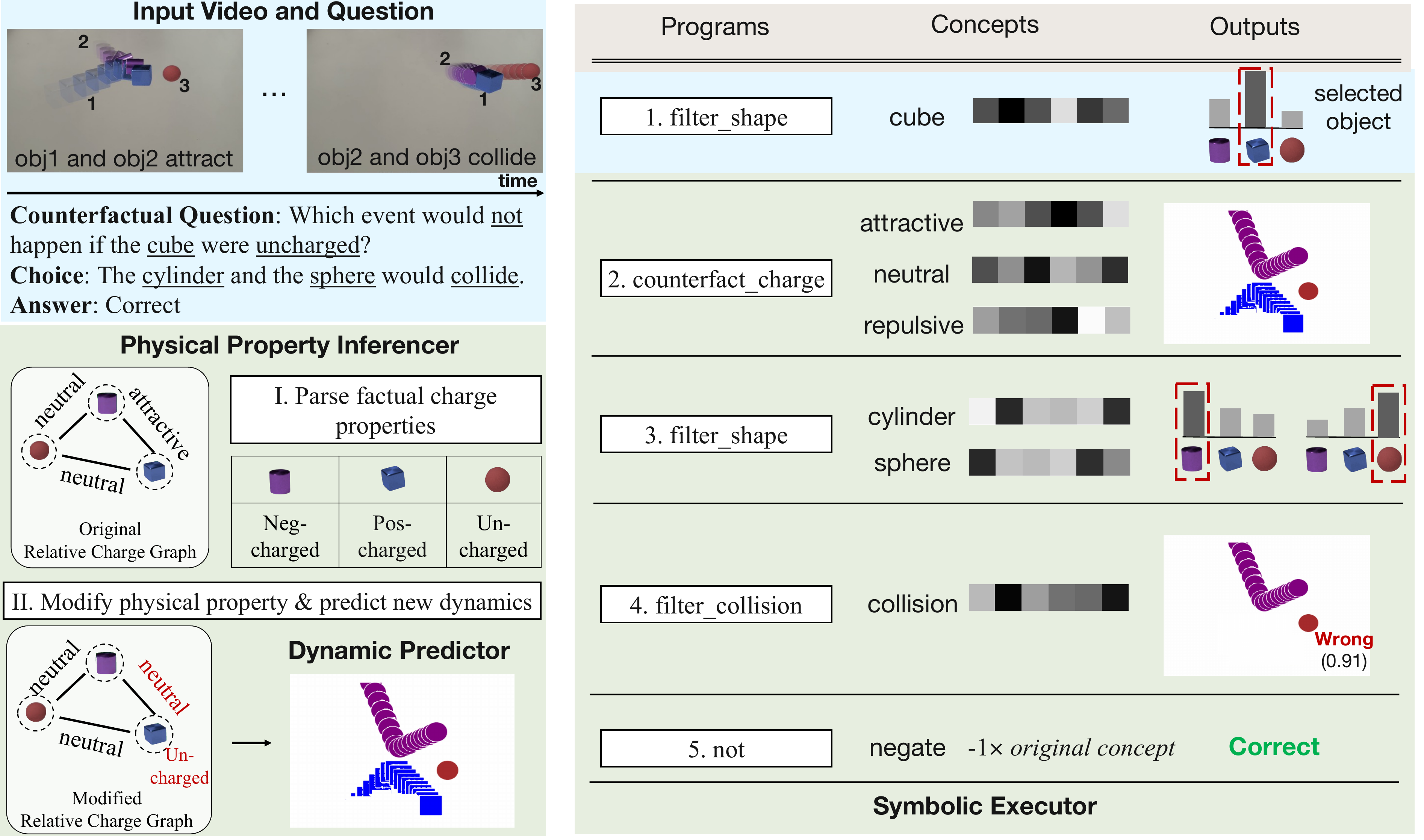}
    \caption{A qualitative example of \modelNew on \dataset. 
    The left-up blue box shows the original video and a counterfactual question to answer. The right half table shows the executable program sequence parsed from the question with concepts related to it and outputs after execution. Specifically, the left-down chart illustrates the execution process of \modelNew for the program ``counterfact charge": 1. \modelNew utilizes a PPI to parse factual charge properties of objects in the scene; 2. \modelNew modifies their properties according to the counterfactual concept and predicts new dynamics using a dynamic predictor.
    }
    \label{fig:qualitative}
    \vspace{-1em}
\end{figure*}

\begin{table*}[t]
\centering
\begin{tabular}{lcccccccccc}
\toprule
\multirow{2}{*}{Methods}& \multicolumn{3}{c}{Static Attributes} & \multicolumn{2}{c}{Dynamic Attributes} & \multicolumn{3}{c}{Events} & \multicolumn{2}{c}{Physical Properties}  \\
\cmidrule(lr){2-4}\cmidrule(lr){5-6}\cmidrule(lr){7-9}\cmidrule(lr){10-11}
                       & Color & Shape & Material   & Moving       & Stationary      & In      & Out              & Collision   &  Mass & Charge       \\
\midrule
\modelNew w/o R        & 91.0 & 91.8 & 92.8 & 83.3 & 85.2 & 85.6 & 81.8  & 86.8  & 79.5  & 45.0  \\
\modelNew w/o CI    & 91.9  & 89.1 & 94.0 & 82.6 & 84.9 & 86.3 & 81.4  & 89.0  & 80.8  & 44.8  \\
\modelNew              & 96.3 & 96.8 & 97.1 & 81.5 & 86.0 & 85.5 & 70.3 & 88.0  & 86.8  & 68.1  \\
\bottomrule
\end{tabular}
\vspace{-1em}
\caption{Evaluation of video concept learning on the validation set.}
\label{tb:concept}
\vspace{-1em}
\end{table*}

\vspace{-1em}
\subsection{Performance Analysis}
{
\noindent{\textbf{Effectiveness of Physical Property Inference.}}
}
We compare the proposed \modelNew with the previous neuro-symbolic method \model~\cite{chen2021grounding} in table~\ref{tb:nsdr}. 
We can see that the \modelNew performs better on all kinds of questions compared to the baseline methods in Table~\ref{tb:qa}. This shows the effectiveness of neuro-symbolic models for physical reasoning. Second, although our \modelNew has no reliance on physical property labels and visual attribute labels during training, it can achieve comparable performance to the previous model \model that requires dense annotation for videos on factual questions. 

One distinguished advantage of \modelNew over end-to-end models~\cite{ding2020object,le2020hierarchical} is it enables step-by-step investigations and thorough analysis for physical concept learning in videos. We compare the model prediction from the \modelNew with the ground-truth labels and calculate the accuracy. Table~\ref{tb:concept} lists the result. We found that our model could effectively grasp visible concepts like ``\textit{colors}", ``\textit{moving}" and ``\textit{collisions}". We also notice that the physical property inferencer in \modelNew can achieve reasonable  
accuracy on physical concepts like ``\textit{mass}" and ``\textit{charge}", which shows \modelNew is able to learn physical properties from objects' trajectories and interactions. However, we also notice the performance gap between the hidden physical properties and the visible properties, which indicates that the bottleneck of the performance on factual questions lies in the hidden physical property inference.

{
\noindent{\textbf{Effectiveness of Dynamics Reasoning.}} We further compare our \modelNew with \model and its variant \model-DPI for dynamic reasoning in table~\ref{tb:nsdr}. \model-DPI follows the previous model NS-DR~\cite{yi2018neural} to adopt dynamic particle interaction networks~\cite{li2018propagation} (DPI)  for dynamic prediction. Note that DPI adopts graph neural networks for dynamic prediction without considering the variance of physical properties. Compared \model and \modelNew with \model-DPI, we can see the importance of modeling mass and charges on nodes and edges of the graph neural networks for dynamic prediction.
Moreover, compared with \model, \modelNew achieves better performance on predictive questions and performs competitively on counterfactual questions, which shows the effectiveness of the differentiable executor for the optimization of property-based dynamic predictor, physical property inference, and visible property grounder.  
The performance of our approach surpasses that of the baselines listed in table~\ref{tb:qa}, particularly in counterfactual and predictive questions.
This achievement demonstrates the model's capability to predict the movements of objects in counterfactual and future scenarios, based on the identification of their underlying physical properties. 
}

{
Furthermore, our evaluation in Section~\ref{sec:eval} highlights a noticeable disparity between the performance of our model, \modelNew, and human performance, particularly in the domain of counterfactual reasoning. We observed that \modelNew's dynamic predictor still exhibits limitations when it comes to long-term dynamic prediction. This indicates that further enhancements to the dynamic predictor could potentially yield even higher performance improvements for \modelNew.
}

\noindent{\textbf{Ablation Study.}}
We conduct a series of ablation studies to prove the effectiveness of the \modelNew in table~\ref{tb:abs} and table~\ref{tb:concept}. \textbf{\modelNew w/o R} denotes learning the property model without using reference videos. \textbf{\modelNew w/o CI} denotes the model without counterfactual imaging. We want to answer the following questions. We report the question-answering accuracy in table~\ref{tb:abs} and the concept classification accuracy in table~\ref{tb:concept}. Comparing with \textbf{\modelNew w/o R} and \textbf{\modelNew}, we can see that reference videos provide important information for concept identification especially for the physical properties, \textit{mass} and \textit{charge} in table~\ref{tb:concept} and constantly improve question-answering performance in different kinds of questions. Comparing \textbf{\modelNew w/o CI} and \textbf{\modelNew}, we can see that the counterfactual imaging mechanism in Section~\ref{training:imagine} can improve the models' abilities in physical property identification in table~\ref{tb:concept}, showing its effectiveness to learn physical reasoning.

\noindent\textbf{{More Diverse Physical Simulated Scenes.}}
\revise{
To better evaluate the model's performance on diverse physical scenes, we have simulated a diverse set of \dataset dataset.
The diverse set introduces 13 distinct object categories—including items such as mugs, pots, chairs, and more—in contrast to the primitive shapes used in the original benchmark. In addition, we incorporate 9 varied backgrounds with realistic textures and lighting conditions, and increase the total number of possible question-answer pairs to 175. The new objects span a wider range of shapes and material properties. These enhancements allow for a richer set of physical interactions, enabling the simulation of complex, compositional events.
}

\revise{
We have also conducted new experiments on these new scenes, and the performance results can be seen in Table~\ref{tab:diverse}. From the table, we have the following observations. First, we can see that our model (\modelNew) still constantly outperforms the other baselines, showing the effectiveness of using neuro-symbolic models for physical reasoning. 
Second, we also observe that the average model performance is worse than their accuracy on the original data in Table~\ref{tb:qa} and Table~\ref{tb:nsdr}. We believe that the reason is that the new physical scenes have provided more diverse physical interaction among the objects, making it more challenging for the AI models.
We have also conducted a human study similar to the original \dataset paper. The accuracy for different kinds of questions is 88.6 for factual questions, 73.7 for predictive questions, and 78.9 for counterfactual questions, much better than existing models in Table~\ref{tab:diverse}. This shows that although the scenes become more diverse, people can still handle these questions well.
}\revise{We provide more details on diverse simulated videos in the supplementary material.}
\begin{table}[t]
\begin{center}
\setlength{\tabcolsep}{2pt}
\begin{tabular}{lccccc}
\toprule
\multirow{2}{*}{Methods}  & \multirow{2}{*}{Factual} & \multicolumn{2}{c}{Predictive} &  \multicolumn{2}{c}{Counterfactual} \\
             &               & per opt.       & per ques.      & per opt.      & per ques.     \\
\midrule
Random    &1.8 &50.1 &22.9 &48.1 &24.0      \\
Frequent  &15.7 &50.0 &0.0 &50.0 &0.0          \\
Blind-LSTM    &43.2  &50.3  &25.0  &49.2  &23.2            \\
\midrule
CNN-LSTM~\cite{antol2015vqa}    &49.6  &52.8  &29.9  &55.7  &29.7             \\
HCRN~\cite{le2020hierarchical}  &51.5  &\Scnd{56.3}  & \Scnd{34.1}  &51.9  &30.1             \\
\midrule
MAC~\cite{hudson2018compositional}     & \Scnd{51.7}  &50.4  &28.9  &51.9  &26.3      \\
ALOE~\cite{ding2020object}         &46.9  &52.4  &29.0  &51.5  &28.6              \\
\midrule
{CNN-LSTM (Ref)}~\cite{antol2015vqa}      &49.7  &51.4  &23.3  &55.6  &30.5      \\
{MAC (Ref)}~\cite{hudson2018compositional}    &50.6  &51.9  &33.3  &50.8  &25.2  \\
ALOE (Ref)~\cite{ding2020object}      &48.6  &51.2  &26.1  &52.9  &27.2       \\
\midrule
ALPRO~\cite{li2022align}  &47.1  &51.8  &28.9  &52.6  &28.4   \\
GPT-4o-mini~\cite{achiam2023gpt} &42.5  &50.0  &29.2  & \Scnd{58.8}  & \Scnd{30.7}       \\
Gemini~\cite{team2023gemini} &34.2  &50.3  &25.7  &49.4  &30.6     \\
\midrule
\textbf{PCR (ours)}  & \Frst{68.4}  & \Frst{58.3}  & \Frst{34.9}  & \Frst{60.3}  & \Frst{32.8}        \\
\midrule
Human Performance  &88.6  &82.9  &73.7  &88.2  &78.9      \\
\bottomrule
\end{tabular}
\end{center}
\vspace{-1em}
\caption{\revise{Evaluation of physical reasoning on \datasetNew. {Human performance is based on sampled questions. See the text for more details. \Frst{Red} text and \Scnd{blue} text indicate the first and second best results other than human performance.}}}
\vspace{-2em}
\label{tab:diverse}
\end{table}

\noindent{\textbf{Generalization to Real-World Scenes.}}
\revise{
We evaluated the performance of our new model, \modelNew, on the real-world dataset. It achieved 63.5\% accuracy on factual questions, 70.4\% on predictive questions (per option), 62.7\% on predictive questions (per question), 54.6\% on counterfactual questions (per option), and 36.5\% on counterfactual questions (per question). From Table~\ref{tb:real}, \modelNew consistently outperforms the MAC model across all question types, demonstrating its enhanced effectiveness in physical reasoning.
}

\noindent{\textbf{Qualitative Case Study.}}
As shown in Figure~\ref{fig:qualitative}, \modelNew can transfer the question query into a series of executable operators, perceive objects' visible properties, infer objects' physical properties, and predict their corresponding dynamics to correctly answer the question. Note that such step-by-step investigation is not possible in previous end-to-end models like \textbf{MAC} and \textbf{ALOE}, showing the transparency and interpretability of our \modelNew.

\vspace{-1em}
\revise{
\subsection{Discussion on Intergrate \modelNew with LVLMs}
Combining \modelNew with LVLMs offers a powerful paradigm for enhancing both robustness and flexibility. First, LVLMs can replace or augment the program parser in \modelNew via in‑context learning, improving program synthesis for diverse linguistic formulations. Second, LVLMs’ broad world knowledge can be invoked through a dedicated large language model-based module to handle commonsense reasoning tasks that lie outside \modelNew’s original training distribution. Finally, LVLMs can act as high‑level controllers, orchestrating \modelNew’s neural modules alongside external modules to seamlessly tackle novel tasks. This integration leverages the precise, learned functionality of \modelNew and the generalist capabilities of LVLMs, yielding a more versatile and powerful system.
}\revise{We provide more experiments and analysis of integration of \modelNew and LVLMs in the supplementary material.}
\vspace{-1em}
\section{Conclusions}\label{sec:conclusion}
In this paper, we introduce the Compositional Physical Reasoning benchmarks, which challenge models to infer hidden physical properties such as mass and charge from limited video observations and leverage this information to predict dynamics and answer structured questions. Our evaluation of state-of-the-art models on \dataset reveals substantial limitations in their ability to reason about these hidden attributes. We also propose a neuro-symbolic framework, \modelNew, that integrates object-centric representations with modular reasoning to jointly learn and infer both visible and hidden physical properties. We further present a real-world dataset to evaluate the generalization of physical reasoning models beyond simulation. Our findings highlight the critical role of hidden physical properties in dynamic scene understanding and expose the gap between current model capabilities and human-level reasoning, paving the way for more robust and generalizable physical reasoning in AI systems.

\bibliographystyle{IEEEtran}
\bibliography{IEEEabrv,egbib}

\begin{thebibliography}{10}
\providecommand{\url}[1]{#1}
\csname url@samestyle\endcsname
\providecommand{\newblock}{\relax}
\providecommand{\bibinfo}[2]{#2}
\providecommand{\BIBentrySTDinterwordspacing}{\spaceskip=0pt\relax}
\providecommand{\BIBentryALTinterwordstretchfactor}{4}
\providecommand{\BIBentryALTinterwordspacing}{\spaceskip=\fontdimen2\font plus
\BIBentryALTinterwordstretchfactor\fontdimen3\font minus \fontdimen4\font\relax}
\providecommand{\BIBforeignlanguage}[2]{{%
\expandafter\ifx\csname l@#1\endcsname\relax
\typeout{** WARNING: IEEEtran.bst: No hyphenation pattern has been}%
\typeout{** loaded for the language `#1'. Using the pattern for}%
\typeout{** the default language instead.}%
\else
\language=\csname l@#1\endcsname
\fi
#2}}
\providecommand{\BIBdecl}{\relax}
\BIBdecl

\bibitem{bakhtin2019phyre}
A.~Bakhtin, L.~van~der Maaten, J.~Johnson, L.~Gustafson, and R.~Girshick, ``Phyre: A new benchmark for physical reasoning,'' in \emph{Advances in Neural Information Processing Systems}, vol.~32, 2019.

\bibitem{yi2019clevrer}
K.~Yi, C.~Gan, Y.~Li, P.~Kohli, J.~Wu, A.~Torralba, and J.~B. Tenenbaum, ``Clevrer: Collision events for video representation and reasoning,'' in \emph{International Conference on Learning Representations}, 2020.

\bibitem{Baradel2020CoPhy}
F.~Baradel, N.~Neverova, J.~Mille, G.~Mori, and C.~Wolf, ``Cophy: Counterfactual learning of physical dynamics,'' in \emph{International Conference on Learning Representations}, 2020.

\bibitem{ates2020craft}
T.~Ates, M.~S. Atesoglu, C.~Yigit, I.~Kesen, M.~Kobas, E.~Erdem, A.~Erdem, T.~Goksun, and D.~Yuret, ``Craft: A benchmark for causal reasoning about forces and interactions,'' \emph{arXiv preprint arXiv:2012.04293}, 2020.

\bibitem{chen2021grounding}
Z.~Chen, J.~Mao, J.~Wu, K.-Y.~K. Wong, J.~B. Tenenbaum, and C.~Gan, ``Grounding physical concepts of objects and events through dynamic visual reasoning,'' in \emph{International Conference on Learning Representations}, 2021.

\bibitem{yi2018neural}
K.~Yi, J.~Wu, C.~Gan, A.~Torralba, P.~Kohli, and J.~B. Tenenbaum, ``{Neural-Symbolic VQA: Disentangling Reasoning from Vision and Language Understanding},'' in \emph{Advances in Neural Information Processing Systems (NIPS)}, 2018.

\bibitem{Mao2019NeuroSymbolic}
J.~Mao, C.~Gan, P.~Kohli, J.~B. Tenenbaum, and J.~Wu, ``{The Neuro-Symbolic Concept Learner: Interpreting Scenes, Words, and Sentences From Natural Supervision},'' in \emph{International Conference on Learning Representations}, 2019.

\bibitem{chencomphy}
Z.~Chen, K.~Yi, Y.~Li, M.~Ding, A.~Torralba, J.~B. Tenenbaum, and C.~Gan, ``Comphy: Compositional physical reasoning of objects and events from videos,'' in \emph{International Conference on Learning Representations}.

\bibitem{achiam2023gpt}
J.~Achiam, S.~Adler, S.~Agarwal, L.~Ahmad, I.~Akkaya, F.~L. Aleman, D.~Almeida, J.~Altenschmidt, S.~Altman, S.~Anadkat \emph{et~al.}, ``Gpt-4 technical report,'' \emph{arXiv}, 2023.

\bibitem{team2023gemini}
G.~Team, R.~Anil, S.~Borgeaud, Y.~Wu, J.-B. Alayrac, J.~Yu, R.~Soricut, J.~Schalkwyk, A.~M. Dai, A.~Hauth \emph{et~al.}, ``Gemini: a family of highly capable multimodal models,'' \emph{arXiv}, 2023.

\bibitem{johnson2017clevr}
J.~Johnson, B.~Hariharan, L.~Van Der~Maaten, L.~Fei-Fei, C.~Lawrence~Zitnick, and R.~Girshick, ``Clevr: A diagnostic dataset for compositional language and elementary visual reasoning,'' in \emph{CVPR}, 2017.

\bibitem{tapaswi2016movieqa}
M.~Tapaswi, Y.~Zhu, R.~Stiefelhagen, A.~Torralba, R.~Urtasun, and S.~Fidler, ``Movieqa: Understanding stories in movies through question-answering,'' in \emph{Proceedings of the IEEE conference on computer vision and pattern recognition}, 2016.

\bibitem{jang2017tgif}
Y.~Jang, Y.~Song, Y.~Yu, Y.~Kim, and G.~Kim, ``Tgif-qa: Toward spatio-temporal reasoning in visual question answering,'' in \emph{Proceedings of the IEEE conference on computer vision and pattern recognition}, 2017.

\bibitem{lei2019tvqa}
J.~Lei, L.~Yu, T.~L. Berg, and M.~Bansal, ``Tvqa+: Spatio-temporal grounding for video question answering,'' in \emph{Tech Report, arXiv}, 2019.

\bibitem{grunde2021agqa}
M.~Grunde-McLaughlin, R.~Krishna, and M.~Agrawala, ``Agqa: A benchmark for compositional spatio-temporal reasoning,'' in \emph{CVPR}, 2021.

\bibitem{riochet2018intphys}
R.~Riochet, M.~Y. Castro, M.~Bernard, A.~Lerer, R.~Fergus, V.~Izard, and E.~Dupoux, ``Intphys: A framework and benchmark for visual intuitive physics reasoning,'' \emph{arXiv preprint arXiv:1803.07616}, 2018.

\bibitem{rajani2020esprit}
N.~F. Rajani, R.~Zhang, Y.~C. Tan, S.~Zheng, J.~Weiss, A.~Vyas, A.~Gupta, C.~Xiong, R.~Socher, and D.~Radev, ``Esprit: explaining solutions to physical reasoning tasks,'' in \emph{ACL}, 2020.

\bibitem{bear2021physion}
D.~M. Bear, E.~Wang, D.~Mrowca, F.~J. Binder, H.-Y.~F. Tung, R.~Pramod, C.~Holdaway, S.~Tao, K.~Smith, F.-Y. Sun \emph{et~al.}, ``Physion: Evaluating physical prediction from vision in humans and machines,'' \emph{arXiv}, 2021.

\bibitem{tung2023physion++}
H.-Y. Tung, M.~Ding, Z.~Chen, D.~Bear, C.~Gan, J.~B. Tenenbaum, D.~L. Yamins, J.~E. Fan, and K.~A. Smith, ``Physion++: Evaluating physical scene understanding that requires online inference of different physical properties,'' \emph{arXiv}, 2023.

\bibitem{girdhar2019cater}
R.~Girdhar and D.~Ramanan, ``Cater: A diagnostic dataset for compositional actions and temporal reasoning,'' in \emph{ICLR}, 2020.

\bibitem{zheng2024contphy}
Z.~Zheng, X.~Yan, Z.~Chen, J.~Wang, Q.~Z.~E. Lim, J.~B. Tenenbaum, and C.~Gan, ``Contphy: continuum physical concept learning and reasoning from videos,'' in \emph{ICML}, 2024.

\bibitem{patel2022cripp}
M.~Patel and T.~Gokhale, ``Cripp-vqa: Counterfactual reasoning about implicit physical properties via video question answering,'' in \emph{EMNLP}, 2022.

\bibitem{battaglia2013simulation}
P.~W. Battaglia, J.~B. Hamrick, and J.~B. Tenenbaum, ``Simulation as an engine of physical scene understanding,'' \emph{Proceedings of the National Academy of Sciences}, vol. 110, no.~45, pp. 18\,327--18\,332, 2013.

\bibitem{hamrick2016inferring}
J.~B. Hamrick, P.~W. Battaglia, T.~L. Griffiths, and J.~B. Tenenbaum, ``Inferring mass in complex scenes by mental simulation,'' \emph{Cognition}, vol. 157, pp. 61--76, 2016.

\bibitem{wu2015galileo}
J.~Wu, I.~Yildirim, J.~J. Lim, B.~Freeman, and J.~Tenenbaum, ``Galileo: Perceiving physical object properties by integrating a physics engine with deep learning,'' \emph{Advances in neural information processing systems}, vol.~28, pp. 127--135, 2015.

\bibitem{lerer2016learning}
A.~Lerer, S.~Gross, and R.~Fergus, ``Learning physical intuition of block towers by example,'' in \emph{International conference on machine learning}.\hskip 1em plus 0.5em minus 0.4em\relax PMLR, 2016, pp. 430--438.

\bibitem{kipf2017semi}
T.~N. Kipf and M.~Welling, ``Semi-supervised classification with graph convolutional networks,'' in \emph{International Conference on Learning Representations (ICLR)}, 2017.

\bibitem{battaglia2016interaction}
P.~Battaglia, R.~Pascanu, M.~Lai, D.~Jimenez~Rezende, and k.~kavukcuoglu, ``Interaction networks for learning about objects, relations and physics,'' in \emph{Advances in Neural Information Processing Systems}, vol.~29, 2016.

\bibitem{chang2016compositional}
M.~B. Chang, T.~Ullman, A.~Torralba, and J.~B. Tenenbaum, ``A compositional object-based approach to learning physical dynamics,'' \emph{arXiv preprint arXiv:1612.00341}, 2016.

\bibitem{sanchez2020learning}
A.~Sanchez-Gonzalez, J.~Godwin, T.~Pfaff, R.~Ying, J.~Leskovec, and P.~Battaglia, ``Learning to simulate complex physics with graph networks,'' in \emph{International Conference on Machine Learning}.\hskip 1em plus 0.5em minus 0.4em\relax PMLR, 2020, pp. 8459--8468.

\bibitem{li2018learning}
Y.~Li, J.~Wu, R.~Tedrake, J.~B. Tenenbaum, and A.~Torralba, ``Learning particle dynamics for manipulating rigid bodies, deformable objects, and fluids,'' in \emph{ICLR}, 2019.

\bibitem{wang2024sok}
A.~Wang, B.~Wu, S.~Chen, Z.~Chen, H.~Guan, W.-N. Lee, L.~E. Li, and C.~Gan, ``Sok-bench: A situated video reasoning benchmark with aligned open-world knowledge,'' in \emph{CVPR}, 2024.

\bibitem{mun2017marioqa}
J.~Mun, P.~Hongsuck~Seo, I.~Jung, and B.~Han, ``Marioqa: Answering questions by watching gameplay videos,'' in \emph{Proceedings of the IEEE International Conference on Computer Vision}, 2017.

\bibitem{lei2018tvqa}
J.~Lei, L.~Yu, M.~Bansal, and T.~L. Berg, ``Tvqa: Localized, compositional video question answering,'' in \emph{EMNLP}, 2018.

\bibitem{vinyals2016matching}
O.~Vinyals, C.~Blundell, T.~Lillicrap, D.~Wierstra \emph{et~al.}, ``Matching networks for one shot learning,'' in \emph{NeurIPS}, 2016.

\bibitem{snell2017prototypical}
J.~Snell, K.~Swersky, and R.~Zemel, ``Prototypical networks for few-shot learning,'' in \emph{NeurIPS}, 2017.

\bibitem{sung2018learning}
F.~Sung, Y.~Yang, L.~Zhang, T.~Xiang, P.~H. Torr, and T.~M. Hospedales, ``Learning to compare: Relation network for few-shot learning,'' in \emph{CVPR}, 2018.

\bibitem{Han2019Visual}
C.~Han, J.~Mao, C.~Gan, J.~B. Tenenbaum, and J.~Wu, ``{Visual Concept Metaconcept Learning},'' in \emph{NeurIPS}, 2019.

\bibitem{coumans2021}
E.~Coumans and Y.~Bai, ``Pybullet, a python module for physics simulation for games, robotics and machine learning,'' \url{http://pybullet.org}, 2016--2021.

\bibitem{blender}
\BIBentryALTinterwordspacing
B.~O. Community, ``Blender - a 3d modelling and rendering package,'' Blender Foundation, Stichting Blender Foundation, Amsterdam, 2018. [Online]. Available: \url{http://www.blender.org}
\BIBentrySTDinterwordspacing

\bibitem{hochreiter1997long}
S.~Hochreiter and J.~Schmidhuber, ``Long short-term memory,'' \emph{Neural computation}, vol.~9, no.~8, pp. 1735--1780, 1997.

\bibitem{antol2015vqa}
S.~Antol, A.~Agrawal, J.~Lu, M.~Mitchell, D.~Batra, C.~L. Zitnick, and D.~Parikh, ``Vqa: Visual question answering,'' in \emph{ICCV}, 2015.

\bibitem{le2020hierarchical}
T.~M. Le, V.~Le, S.~Venkatesh, and T.~Tran, ``Hierarchical conditional relation networks for video question answering,'' in \emph{CVPR}, 2020.

\bibitem{hudson2018compositional}
D.~A. Hudson and C.~D. Manning, ``Compositional attention networks for machine reasoning,'' in \emph{ICLR}, 2018.

\bibitem{ding2020object}
D.~Ding, F.~Hill, A.~Santoro, and M.~Botvinick, ``Attention over learned object embeddings enables complex visual reasoning,'' \emph{arXiv}, 2020.

\bibitem{li2022align}
D.~Li, J.~Li, H.~Li, J.~C. Niebles, and S.~C. Hoi, ``Align and prompt: Video-and-language pre-training with entity prompts,'' in \emph{Proceedings of the IEEE/CVF Conference on Computer Vision and Pattern Recognition}, 2022, pp. 4953--4963.

\bibitem{VQA}
S.~Antol, A.~Agrawal, J.~Lu, M.~Mitchell, D.~Batra, C.~L. Zitnick, and D.~Parikh, ``{VQA}: {V}isual {Q}uestion {A}nswering,'' in \emph{International Conference on Computer Vision (ICCV)}, 2015.

\bibitem{He_2016_CVPR}
K.~He, X.~Zhang, S.~Ren, and J.~Sun, ``Deep residual learning for image recognition,'' in \emph{CVPR}, 2016.

\bibitem{vaswani2017attention}
A.~Vaswani, N.~Shazeer, N.~Parmar, J.~Uszkoreit, L.~Jones, A.~N. Gomez, {\L}.~Kaiser, and I.~Polosukhin, ``Attention is all you need,'' in \emph{NeurIPS}, 2017.

\bibitem{burgess2019monet}
C.~P. Burgess, L.~Matthey, N.~Watters, R.~Kabra, I.~Higgins, M.~Botvinick, and A.~Lerchner, ``Monet: Unsupervised scene decomposition and representation,'' \emph{arXiv preprint arXiv:1901.11390}, 2019.

\bibitem{hudson2019gqa}
D.~A. Hudson and C.~D. Manning, ``Gqa: A new dataset for real-world visual reasoning and compositional question answering,'' in \emph{CVPR}, 2019.

\bibitem{xu2017video}
D.~Xu, Z.~Zhao, J.~Xiao, F.~Wu, H.~Zhang, X.~He, and Y.~Zhuang, ``Video question answering via gradually refined attention over appearance and motion,'' in \emph{Proceedings of the 25th ACM international conference on Multimedia}, 2017, pp. 1645--1653.

\bibitem{he2017mask}
K.~He, G.~Gkioxari, P.~Doll{\'a}r, and R.~Girshick, ``Mask r-cnn,'' in \emph{CVPR}, 2017.

\bibitem{ren2015faster}
S.~Ren, K.~He, R.~Girshick, and J.~Sun, ``Faster r-cnn: Towards real-time object detection with region proposal networks,'' \emph{NeurIPS}, 2015.

\bibitem{chen19acl}
Z.~Chen, L.~Ma, W.~Luo, and K.-Y.~K. Wong, ``Weakly-supervised spatio-temporally grounding natural sentence in video,'' in \emph{ACL}, 2019.

\bibitem{munkres1957algorithms}
J.~Munkres, ``Algorithms for the assignment and transportation problems,'' \emph{Journal of the society for industrial and applied mathematics}, 1957.

\bibitem{kipf2018neural}
T.~Kipf, E.~Fetaya, K.-C. Wang, M.~Welling, and R.~Zemel, ``Neural relational inference for interacting systems,'' in \emph{International Conference on Machine Learning}.\hskip 1em plus 0.5em minus 0.4em\relax PMLR, 2018, pp. 2688--2697.

\bibitem{bahdanau2014neural}
D.~Bahdanau, K.~Cho, and Y.~Bengio, ``Neural machine translation by jointly learning to align and translate,'' in \emph{ICLR}, 2015.

\bibitem{van2015cognitive}
N.~Van~Hoeck, P.~D. Watson, and A.~K. Barbey, ``Cognitive neuroscience of human counterfactual reasoning,'' \emph{Frontiers in human neuroscience}, 2015.

\bibitem{buchsbaum2012power}
D.~Buchsbaum, S.~Bridgers, D.~Skolnick~Weisberg, and A.~Gopnik, ``The power of possibility: Causal learning, counterfactual reasoning, and pretend play,'' \emph{Philosophical Transactions of the Royal Society B: Biological Sciences}, 2012.

\bibitem{li2018propagation}
Y.~Li, J.~Wu, J.-Y. Zhu, J.~B. Tenenbaum, A.~Torralba, and R.~Tedrake, ``Propagation networks for model-based control under partial observation,'' in \emph{ICRA}, 2019.

\bibitem{Xie_2017_CVPR}
S.~Xie, R.~Girshick, P.~Dollar, Z.~Tu, and K.~He, ``Aggregated residual transformations for deep neural networks,'' in \emph{CVPR}, 2017.

\bibitem{hara2018can}
K.~Hara, H.~Kataoka, and Y.~Satoh, ``Can spatiotemporal 3d cnns retrace the history of 2d cnns and imagenet?'' in \emph{CVPR}, 2018.

\bibitem{qwen2.5}
A.~Yang, B.~Yang, B.~Zhang, B.~Hui, B.~Zheng, B.~Yu, C.~Li, D.~Liu, F.~Huang, H.~Wei, H.~Lin, J.~Yang, J.~Tu, J.~Zhang, J.~Yang, J.~Yang, J.~Zhou, J.~Lin, K.~Dang, K.~Lu, K.~Bao, K.~Yang, L.~Yu, M.~Li, M.~Xue, P.~Zhang, Q.~Zhu, R.~Men, R.~Lin, T.~Li, T.~Xia, X.~Ren, X.~Ren, Y.~Fan, Y.~Su, Y.~Zhang, Y.~Wan, Y.~Liu, Z.~Cui, Z.~Zhang, and Z.~Qiu, ``Qwen2.5 technical report,'' \emph{arXiv}, 2024.

\bibitem{Rombach_2022_CVPR}
R.~Rombach, A.~Blattmann, D.~Lorenz, P.~Esser, and B.~Ommer, ``High-resolution image synthesis with latent diffusion models,'' in \emph{CVPR}, 2022.

\end{thebibliography}
\newpage
\section{appendix}\label{sec:appendix}

In this section, we first provide more examples of the datasets in Section~\ref{sec:data_syn} and Section~\ref{sec:data_real}. We then provide more details on video generation in Section~\ref{sec:data_gen}. We provide more details about symbolic program we learn in Section~\ref{sec:model}. We provide more details about the baselines in Section~\ref{sec:baseline1} and Section~\ref{sec:baseline2}.
\revise{
We discuss how to evaluate our models on more diverse simulated scenes in Section~\ref{sec:appendsim}.
We study how to evaluate models on more diverse real scenes in Section~\ref{sec:appendreal}. Finally, we discuss how to integrate the proposed \modelNew and large vision-language models in Section~\ref{sec:appenlvlm}.
}

\subsection{Examples from \dataset}
\label{sec:data_syn}
\begin{figure*}[t]
    \centering
    \includegraphics[width =0.8\textwidth]{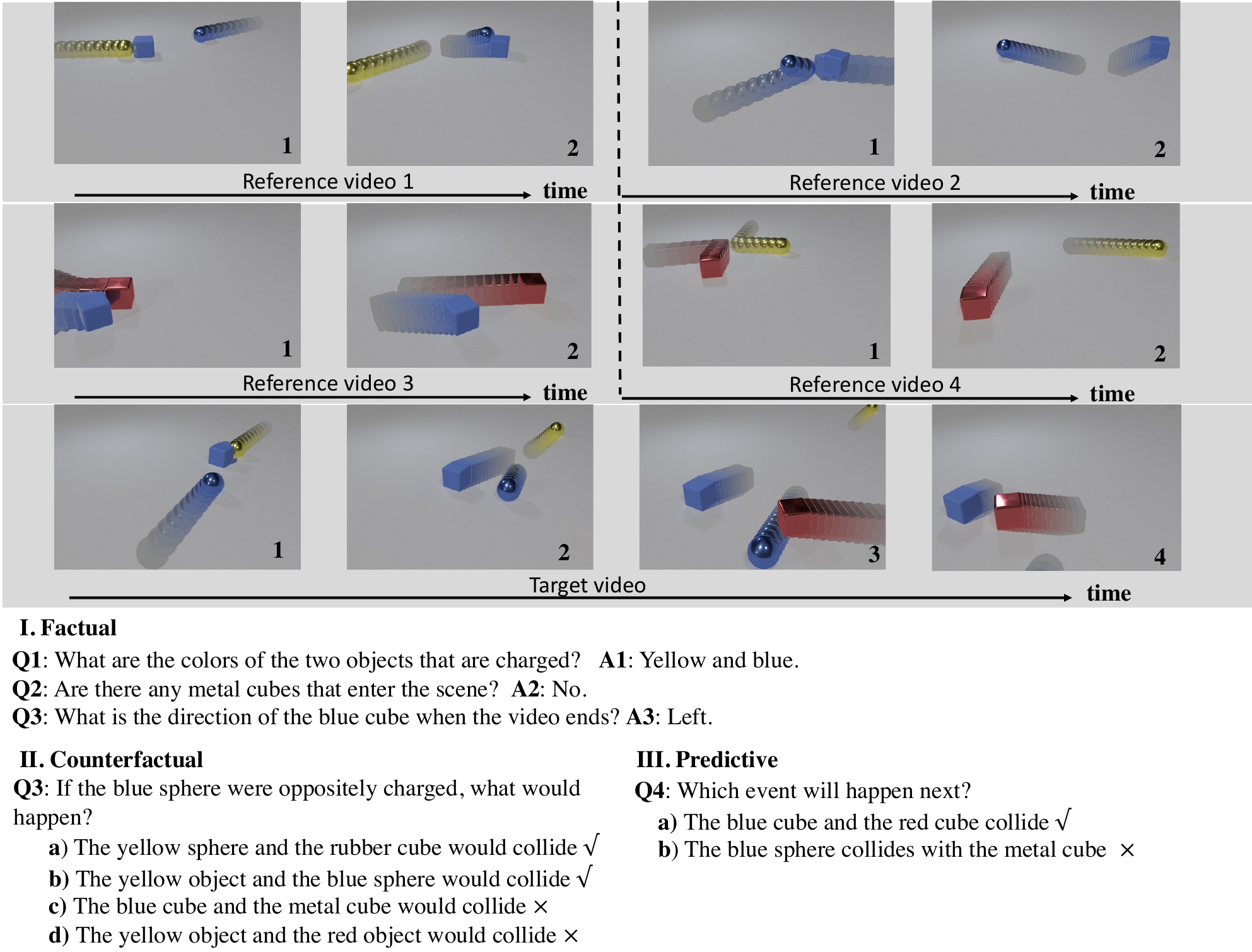}
    \caption{Sample target video, reference videos and question-answer pairs from \dataset.}
    \label{fig:data2}
\end{figure*}

\begin{figure*}[t]
    \centering
    \includegraphics[width =1\textwidth]{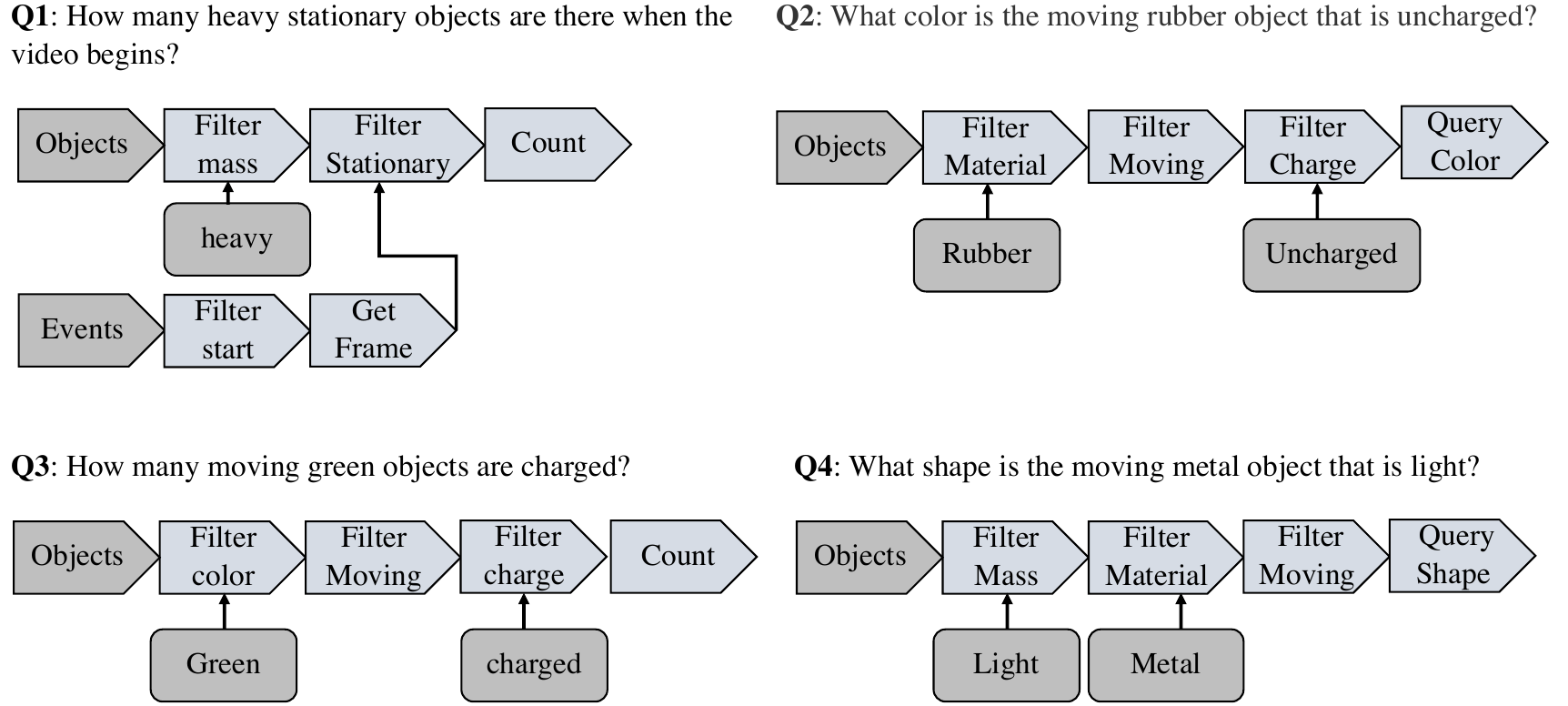}
    \caption{Sample of factual questions and their underlying functional programs in \dataset.}
    \label{fig:ques1}
\end{figure*}

\begin{figure*}[t]
    \centering
    \includegraphics[width =1\textwidth]{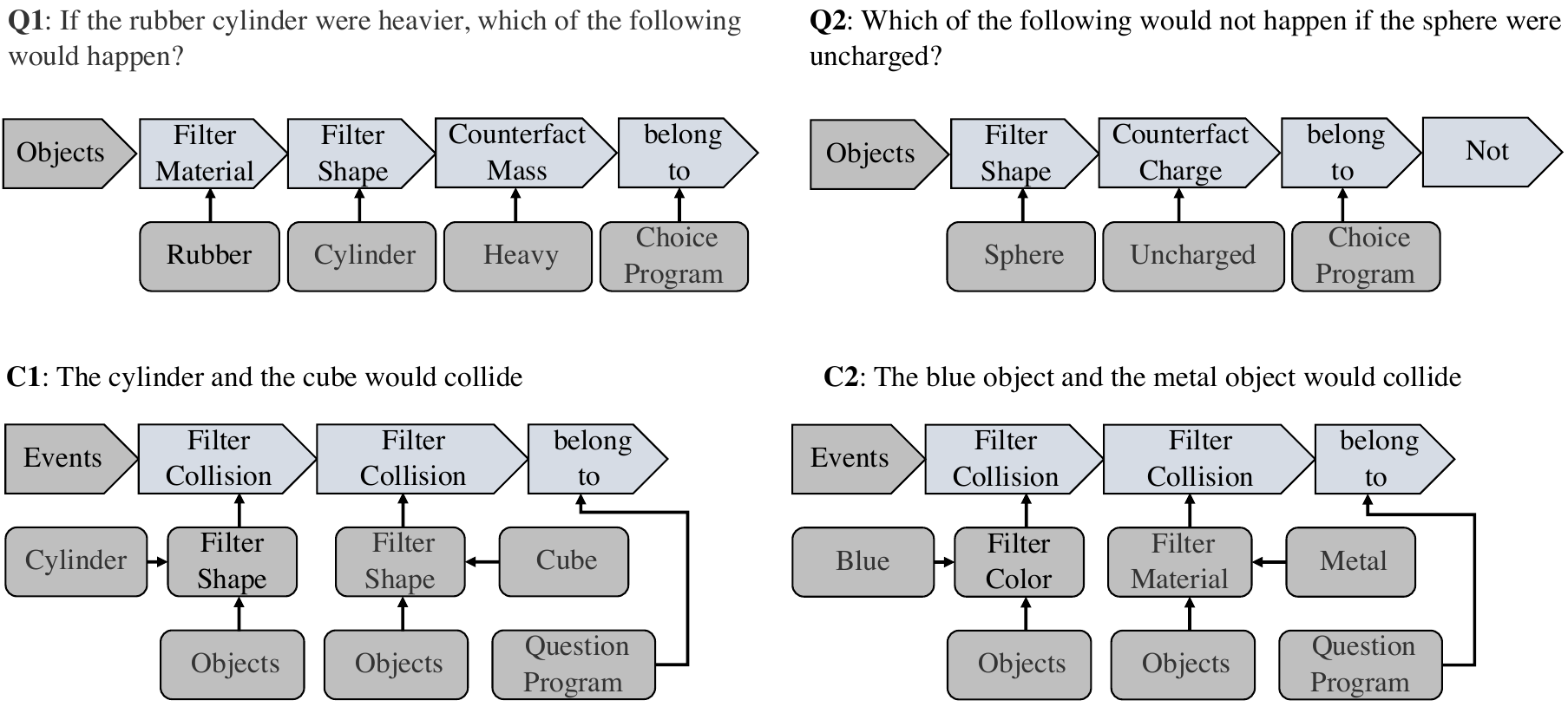}
    \caption{Sample of counterfactual questions, choice options and their underlying functional programs in \dataset.}
    \label{fig:ques2}
\end{figure*}

Here we provide more examples from \dataset in Fig.~\ref{fig:data2}. From these examples, we can see the following features of \dataset. First, to answer the factual questions, models not only need to recognize objects' visual appearance attributes and events in the video but also identify their intrinsic physical properties from the given video set. Second, to answer counterfactual and predictive questions, models need to predict objects' dynamics in counterfactual or future scenes, which can be severely affected by intrinsic physical properties. We also show some typical questions and choice samples as well as their underlying reasoning program logic in Fig.~\ref{fig:ques1} and Fig.~\ref{fig:ques2}.

\subsection{Examples from Real-World Scenario}
\label{sec:data_real}
\begin{figure*}[t]
    \centering
    \includegraphics[width =0.8\textwidth]{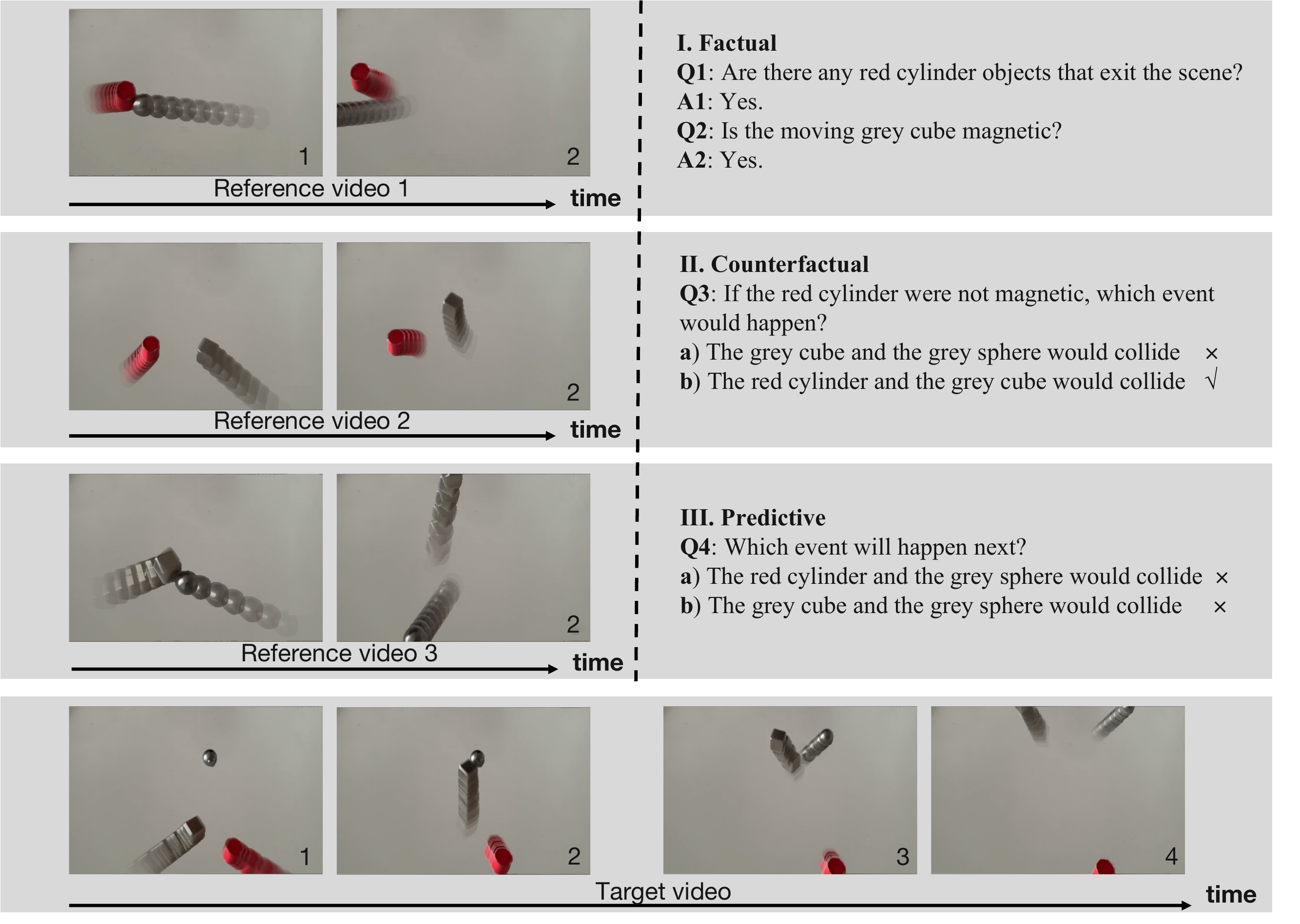}
    \caption{Sample target video, reference videos and question-answer pairs from real-world dataset.}
    \label{fig:real2}
\end{figure*}

We also provide some examples captured from real-world scenarios in Fig.~\ref{fig:real2}. Similarly to the procedure of answering questions for the synthetic data in \dataset, the model needs first to answer the factual questions based on objects' visual attributes and intrinsic physical properties and then answer the counterfactual and predictive questions by predicting the related dynamics. In comparison to \dataset, the real-world dataset exhibits two distinct characteristics. First, unlike objects with a single charge, magnetic monopoles do not exist in the natural world, which results in each magnetized object within a scene lacking a consistent magnetic label across different videos. This necessitates that models rigorously infer magnetic properties through interactions between objects, avoiding shortcuts based on strong coupling between objects and physical attributes. Second, the real-world dataset is manually collected, so it tends to be noisier, especially in more pronounced interaction instances, such as collisions, attraction, and repulsion between objects. These dynamic behaviors may even cause objects to temporarily leave the ground plane. As a result, robustness becomes a critical requirement for models trained on such datasets.
In summary, the real-world dataset serves as a valuable complement to \dataset, offering diverse challenges and enhancing model performance in handling complex and noisy scenarios.

\subsection{Video Generation}
\label{sec:data_gen}
We provide more details for video generation. The generation of the videos in \dataset can be decomposed into two steps. First, we adopt a physical engine Bullet~\cite{coumans2021} to simulate objects' motions and their interactions with each other. Since Bullet does not officially support the effect of electronic charges, we add external forces between charged objects, whose values are inversely proportional to the square of the objects' distance, to simulated Coulomb forces. We assign the \textit{light} object a mass value of 1 and assign the \textit{heavy} object a mass value of 5. We manually make sure that each reference video contains at least an interaction (collision, charge, and mass) among objects to provide enough information for physical property inference. Each object should appear at least once in the reference videos. The simulated objects' motions are sent to Blender\cite{blender} to render high-quality image sequences.

\subsection{Symbolic Program Details}
\label{sec:model}
The symbolic execution component first adopts a program parser to parse the query question into a functional program, containing a series of neural operations. The program parser is an attention-based seq2seq model~\cite{bahdanau2014neural}, whose input is the word sequence in the question/choice and output is the sequence of neural operations. The symbolic executor then executes the operations on the predicted dynamic scene to get the answer to the question.  We summarize all the symbolic operations in \model in table~\ref{tb:operation}. Compared with the previous benchmarks~\cite{yi2019clevrer,ates2020craft}, \dataset has more operations on physical property identification, comparison and corresponding dynamic prediction. We show each symbolic operator in table~\ref{tb:operation}.
\begin{table*}[htbp!]
\centering
\begin{tabular}{cll}
\toprule
Type    & Operation & Signature \\ 
\midrule
\multirow{8}{1.4cm}{\centering{Counterfact Operation}}
& \texttt{Counterfactual\_mass\_heavy} & $ (\textit{object}) \rightarrow \textit{events} $  \\
&  Return all events after making the object heavy  & \\
& \texttt{Counterfactual\_mass\_light} & $ (\textit{object}) \rightarrow \textit{events} $  \\
&  Return all events after making the object light  & \\
& \texttt{Counterfactual\_uncharged} & $ (\textit{object}) \rightarrow \textit{events} $  \\
&  Return all events after making the object uncharged  & \\
& \texttt{Counterfactual\_opposite\_charged} & $ (\textit{object}) \rightarrow \textit{events} $  \\
&  Return all events after making the object oppositely charged  & \\
\midrule
\multirow{8}{1.4cm}{\centering{Object Property Operations}}
& \texttt{filter\_heavy} & $ (\textit{objects}) \rightarrow \textit{objects} $  \\
& select all the heavy objects & \\
& \texttt{filter\_light} & $ (\textit{objects}) \rightarrow \textit{objects} $  \\
& select all the light objects & \\
& \texttt{filter\_charged} & $ (\textit{objects}) \rightarrow \textit{objects} $  \\
& select all the charged objects & \\
& \texttt{filter\_uncharged} & $ (\textit{objects}) \rightarrow \textit{objects} $  \\
& select all the uncharged objects & \\
\midrule
\multirow{4}{1.4cm}{\centering{Object Appearance Operations}} 
& \texttt{Filter\_static\_attr} & $ (\textit{objects}, \textit{attr}) \rightarrow \textit{objects} $  \\
&  Select objects from the input list with the input static attribute & \\ 
& \texttt{Filter\_dynamic\_attr} & $ (\textit{objects}, \textit{attr}, \textit{frame}) \rightarrow \textit{objects} $ \\
&  Selects objects in the input frame with the dynamic attribute & \\ 
\midrule
\multirow{12}{1.4cm}{\centering{Event Operations}} 
& \texttt{Filter\_event} & $ (\textit{events}, \textit{objects}) \rightarrow \textit{events} $  \\
&  Select all events that involve the input objects & \\
& \texttt{Get\_col\_partner} & $ (\textit{event}, \textit{object}) \rightarrow \textit{object} $  \\
& Return the collision partner of the input object & \\
& \texttt{Filter\_before} & $ (\textit{events}, \textit{events}) \rightarrow \textit{events} $  \\
&  Select all events before the target event & \\
& \texttt{Filter\_after} & $ (\textit{events}, \textit{events}) \rightarrow \textit{events} $  \\
&  Select all events after the target event & \\
& \texttt{Filter\_order} & $ (\textit{events}, \textit{order}) \rightarrow \textit{event} $  \\
&  Select the event at the specific time order & \\
& \texttt{Get\_frame} & $ (\textit{event}) \rightarrow \textit{frame} $  \\
&  Return the frame of the input event in the video & \\
\midrule
\multirow{2}{1.4cm}{\centering{Others}} & \texttt{Unique} & $ (\textit{events/objects}) \rightarrow \textit{event/object} $  \\
&  Return the only event/object in the input list  \\
\midrule
\multirow{10}{1.4cm}{\centering{Input Operations}} 
               & \texttt{Start}  & $()\rightarrow \textit{event}$  \\
               &  Returns the special ``start'' event & \\
               & \texttt{end}  & $()\rightarrow \textit{event}$  \\
               &  Returns the special ``end'' event & \\
               & \texttt{Objects}   & $()\rightarrow \textit{objects}$ \\    
               &  Returns all objects in the video &   \\
               & \texttt{Events}   & $()\rightarrow \textit{events}$\\
               & Returns all events happening in the video & \\
               & \texttt{UnseenEvents}  & $()\rightarrow \textit{events}$ \\
               & Returns all future events happening in the video & \\
\midrule
\multirow{18}{1.4cm}{\centering{Output Operations}}
& \texttt{Query\_both\_attribute} & $ (\textit{object},\textit{object}) \rightarrow \textit{attr} $  \\
& Returns the attributes of the input two objects & \\ 
& \texttt{Query\_direction} & $ (\textit{object}, \textit{frame}) \rightarrow \textit{attr} $  \\
& Returns the direction of the object at the input frame & \\ 
& \texttt{Is\_heavier} & $ (\textit{obj1}, \textit{obj2}) \rightarrow \textit{bool} $  \\
& Returns ``yes'' if \textit{obj1} is heavier than \textit{obj2}  & \\ 
& \texttt{Is\_lighter} & $ (\textit{obj1}, \textit{obj2}) \rightarrow \textit{bool} $  \\
& Returns ``yes'' if \textit{obj1} is lighter than \textit{obj2}  & \\ 
& \texttt{Query\_attribute} & $ (\textit{object}) \rightarrow \textit{attr} $  \\
& Returns the attribute of the input objects like color & \\ 
& \texttt{Count} & $ (\textit{objects}) \rightarrow \textit{int} $  \\
&  Returns the number of the input objects/ events & $ (\textit{events}) \rightarrow \textit{int} $ \\
& \texttt{Exist} & $ (\textit{objects}) \rightarrow \textit{bool} $  \\
&  Returns ``yes'' if the input objects is not empty & \\
& \texttt{Belong\_to} & $ (\textit{event}, \textit{events}) \rightarrow \textit{bool} $  \\
&  Returns ``yes'' if the input event belongs to the input event sets & \\
& \texttt{Negate} & $ (\textit{bool}) \rightarrow \textit{bool} $  \\
&  Returns the negation of the input boolean & \\
\bottomrule
\end{tabular}
\caption{Symbolic operations of \modelNew on \dataset.
In this table, ``order'' denotes the chronological order of an event, \eg ``First'' and ``Last''; ``static attribute'' denotes object static concepts like ``Red'' and ``Rubber'' and ``dynamic attribute'' represents object dynamic concepts like ``Moving''.
}
\label{tb:operation}
\end{table*}

\subsection{Baseline Implementation Details}
\label{sec:baseline1}
In this section, we provide more details for baselines in the experimental section. We implement baselines based on the publicly available source code. For multiple-choice questions, we independently concatenate the words of each option and the question as a binary classification question.
Similar to CLEVRER~\cite{yi2019clevrer}, we use ResNet-50~\cite{He_2016_CVPR} to extract visual feature sequences for \textbf{CNN+LSTM} and \textbf{MAC} and variants with reference videos. We evenly sample 25 frames for each target video and 10 frames for each reference video. 
For \textbf{HCRN}, we use the appearance feature from ResNet-101~\cite{He_2016_CVPR} and the motion feature from ResNetXt-101~\cite{Xie_2017_CVPR,hara2018can} following the official implementation. For \textbf{ALOE}, we use MONet\cite{burgess2019monet} to extract visual representation and sample 25 frames for each target video. For \textbf{ALOE (Ref)}, we sample 10 frames for each reference video and concatenate the reference frames and the target frames as visual representations. We train all the models until they are fully converged, select the best checkpoint on the validation set and finally test on the testing set.

\subsection{Large Vision Language Models Details}
\label{sec:baseline2}
In this section, we provide more details on how we utilize Large Vision Language Models, such as \cite{li2022align}, \cite{achiam2023gpt}, \cite{team2023gemini}, to test their physical reasoning ability on \dataset. For \textbf{ALPRO}\cite{li2022align}, we fine-tune the model with both factual, counterfactual, and predictive questions in \dataset's training set until they achieve satisfactory results on the validation set. We convert both open-ended and multiple-choice question formats to align with the input of the model. For open-ended questions, we simply collect the answers to build the vocabulary dictionary. For multiple-choice questions, we assemble each choice with its question to form a new question and utilize the original True/False judgment as the answer. Due to the large variance between open-ended and multiple-choice questions' answer domains, we fine-tune the model separately on the two different types of questions. For \textbf{GPT-4V}\cite{achiam2023gpt} and \textbf{Gemini}\cite{team2023gemini}, we leverage a zero-shot method to test their performance. We evenly sample 16 frames from each target video to form a sequence of frames to represent the original video in the test set and pair the sequence with related questions from the dataset. Then, we add an instructive prompt to guide the model in understanding the physical events that happened in the scenarios and answer the questions in a predefined format.

\subsection{Evaluate Models on More Diverse Simulated Scenes}
\label{sec:appendsim}

\begin{figure*}[t]
    \centering
    \includegraphics[width=\textwidth]{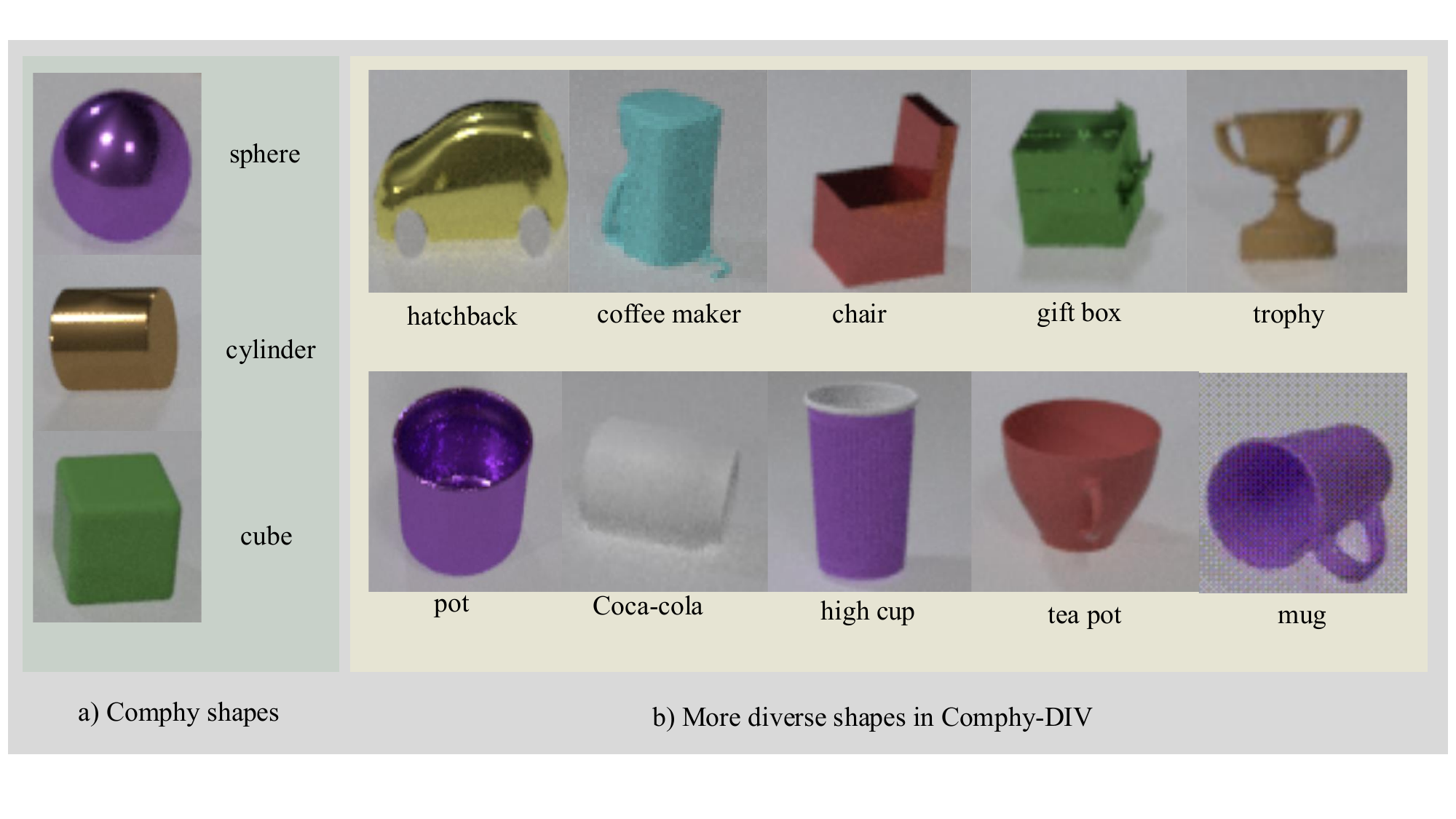}
    \vspace{-2em}
    \caption{\revise{Comparison of shape diversity between \dataset and \datasetNew. As shown in a), the three objects belong to the \dataset dataset, whereas b) illustrates the ten newly added objects in \datasetNew.}}
    \label{fig:visual_div}
\end{figure*}

\noindent\textbf{{Goal of Our benchmark.}}
\revise{
We would like to clarify that the original goal of \dataset is not to mimic complex real-world scenes, but rather to \textbf{provide a diagnostic testbed that isolates and evaluates the physical reasoning capabilities of AI models.} Simplicity in object design and scene setup allows for controlled physical interactions, making it easier to attribute model behavior to underlying reasoning mechanisms. However, we also agree that greater diversity can improve robustness evaluation and broaden the benchmark's applicability.
}

\begin{figure*}[t]
    \centering
    \includegraphics[width=\textwidth]{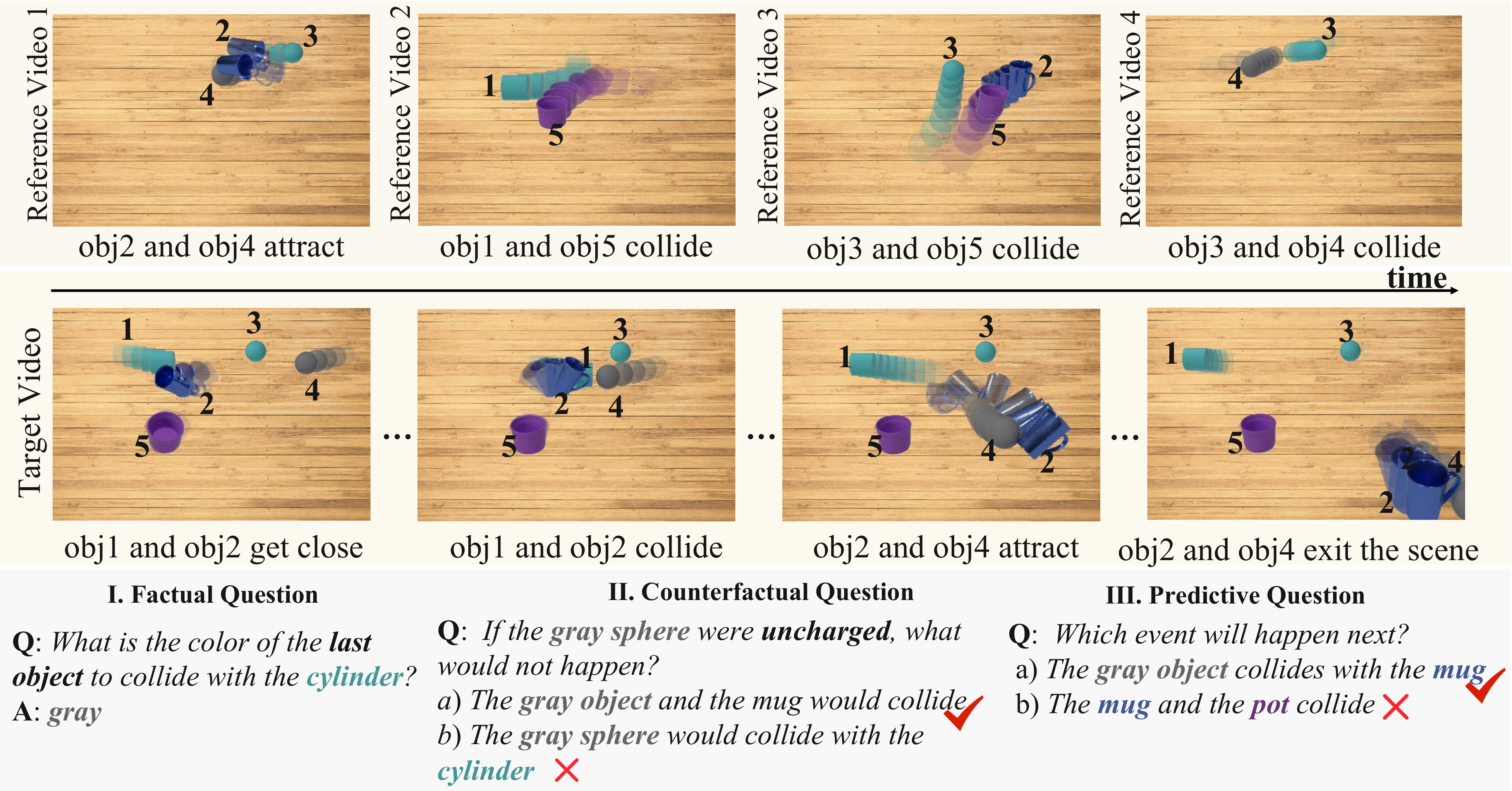}
    \caption{\revise{Qualitative examples of more diverse scenes in the \datasetNew. As shown in the figure, we have more diverse physical interactions between the blue mug and the sphere in the video. The image background is also more diverse with different textures and colors in contrast to the original \dataset in~\cite{chencomphy}.}}
    \label{fig:qual_div}
\end{figure*}

\begin{figure*}[t]
    \centering
    \includegraphics[width=\textwidth]{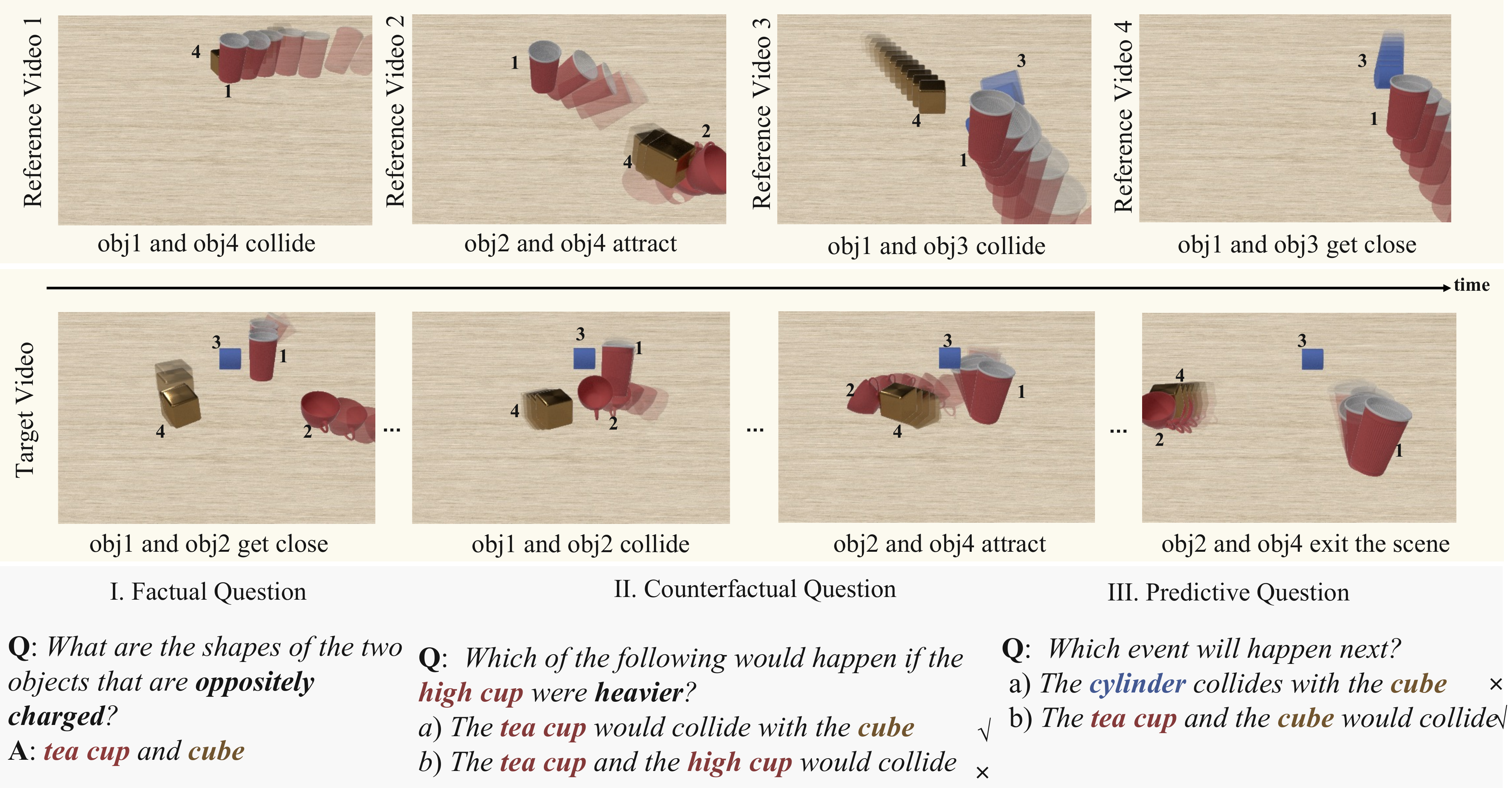}
    \caption{\revise{More qualitative examples of more diverse scenes in the \datasetNew.}}
    \label{fig:qual_div2}
\end{figure*}

\begin{figure*}[t]
    \centering
    \includegraphics[width=\textwidth]{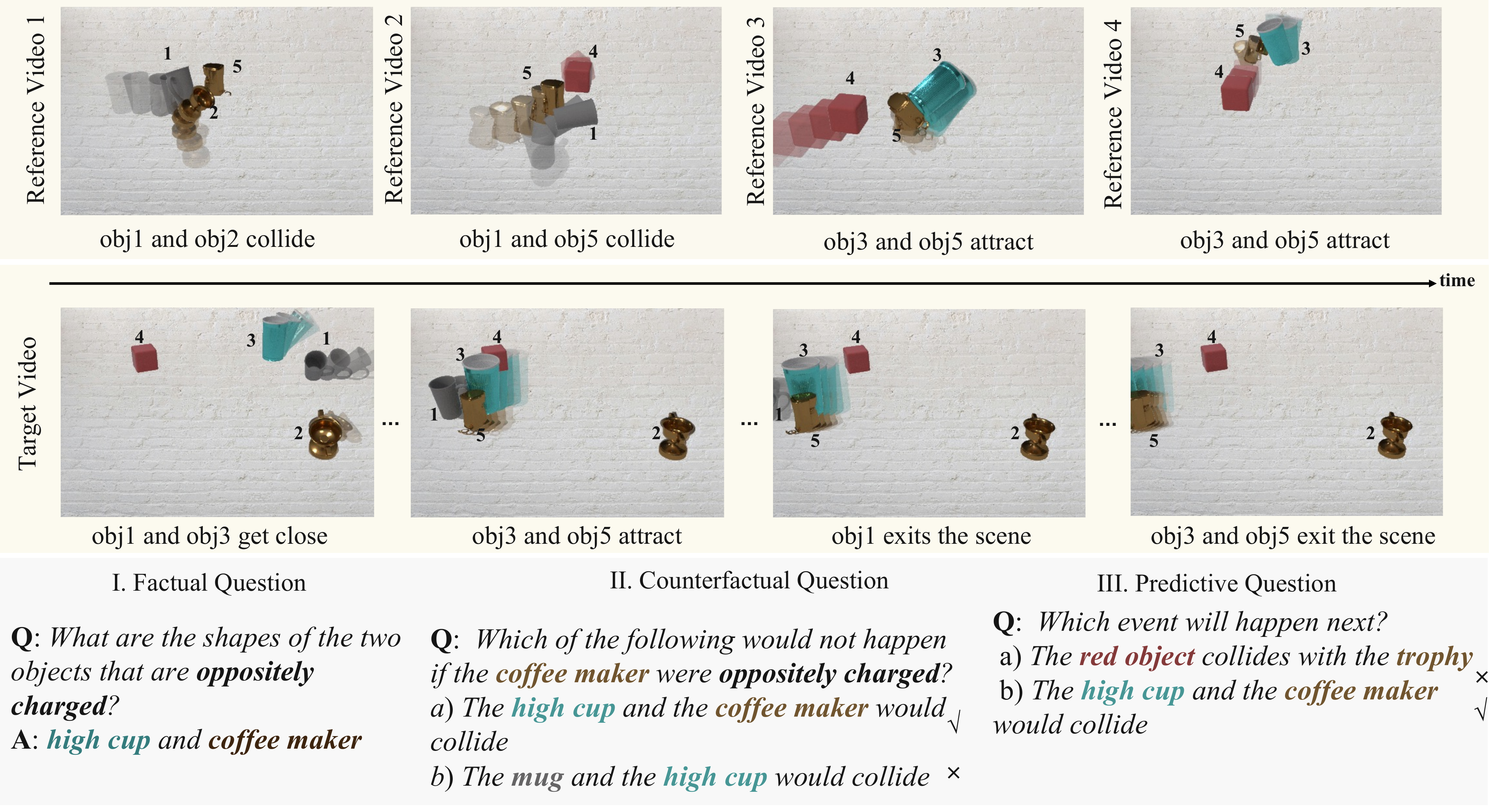}
    \caption{\revise{More qualitative examples of more diverse scenes in the \datasetNew.}}
    \label{fig:qual_div3}
\end{figure*}

\begin{figure*}[t]
    \centering
    \includegraphics[width=\textwidth]{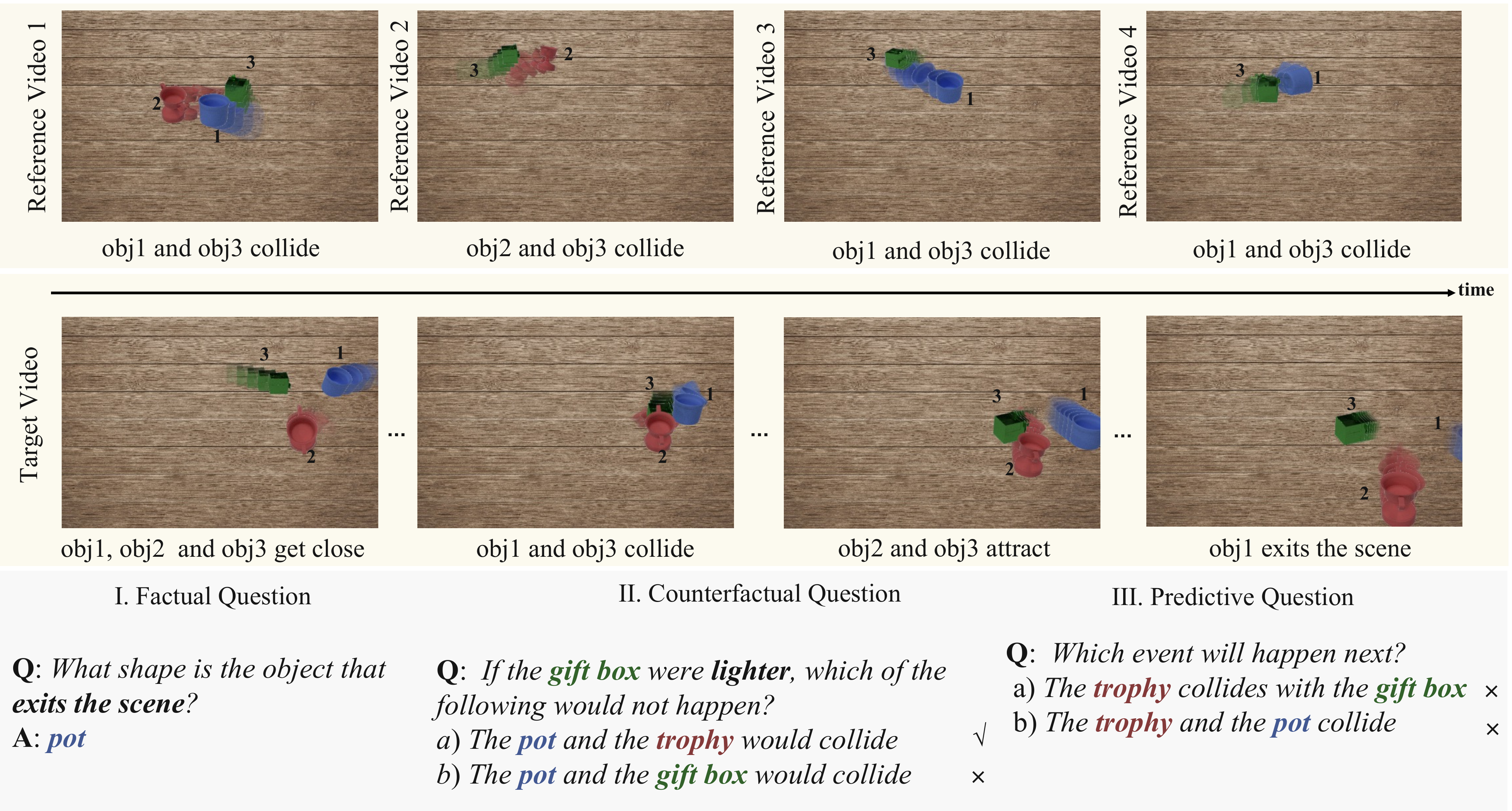}
    \caption{\revise{More qualitative examples of more diverse scenes in the \datasetNew.}}
    \label{fig:qual_div4}
\end{figure*}

\noindent\textbf{{More Diverse Physical Simulated Scenes.}} 
\revise{
To provide more diverse physical reasoning, we have significantly expanded the dataset to create a new version, \datasetNew. This version introduces 13 distinct object categories—including items such as mugs, pots, chairs, and more—in contrast to the primitive shapes used in the original benchmark. In addition, we incorporate 9 varied backgrounds with realistic textures and lighting conditions, and increase the total number of possible question-answer pairs to 175. Note that there are only 3 primitive shapes in the same background in the original dataset.  As shown in Figure~\ref{fig:visual_div}, the new objects span a wider range of shapes and material properties. These enhancements allow for a richer set of physical interactions, enabling the simulation of complex, compositional events. Qualitative examples of these new scenes are presented in Fig.~\ref{fig:qual_div} and Fig.~\ref{fig:qual_div2}--\ref{fig:qual_div4}, which demonstrate diverse object movements, interactions, and backgrounds.
}

{
\noindent\textbf{{New Experimental Results on the Simulated Scenes.}} 
To evaluate the effectiveness of \datasetNew, we conducted new experiments with both our proposed method and baseline models. Results are summarized in Table~\ref{tab:diverse}.
Our model (\modelNew) continues to outperform baseline methods, indicating its superior reasoning ability even in the presence of increased visual and physical complexity. Notably, the overall performance of all models has declined compared to results on the original dataset (see Table~\blue{3} and Table~\blue{5} in the main paper), which confirms that the added diversity makes the benchmark more challenging and discriminative. Additionally, we conducted a human study following the same protocol used in the original ComPhy paper.
Human participants achieved accuracies of 88.6\% for factual questions, 73.7\% for predictive questions, and 78.9\% for counterfactual questions—substantially higher than those of AI models—demonstrating that despite increased complexity, humans remain robust and reliable at these reasoning tasks.
}

\begin{table}[t]
\begin{center}
\setlength{\tabcolsep}{2pt}
\begin{tabular}{lccccc}
\toprule
\multirow{2}{*}{Methods}  & \multirow{2}{*}{Factual} & \multicolumn{2}{c}{Predictive} &  \multicolumn{2}{c}{Counterfactual} \\
             &               & per opt.       & per ques.      & per opt.      & per ques.     \\
\midrule
Random    &1.8 &50.1 &22.9 &48.1 &24.0      \\
Frequent  &15.7 &50.0 &0.0 &50.0 &0.0          \\
Blind-LSTM    &43.2  &50.3  &25.0  &49.2  &23.2            \\
\midrule
CNN-LSTM~\cite{antol2015vqa}    &49.6  &52.8  &29.9  &55.7  &29.7             \\
HCRN~\cite{le2020hierarchical}  &51.5  &\Scnd{56.3}  & \Scnd{34.1}  &51.9  &30.1             \\
\midrule
MAC~\cite{hudson2018compositional}     & \Scnd{51.7}  &50.4  &28.9  &51.9  &26.3      \\
ALOE~\cite{ding2020object}         &46.9  &52.4  &29.0  &51.5  &28.6              \\
\midrule
{CNN-LSTM (Ref)}~\cite{antol2015vqa}      &49.7  &51.4  &23.3  &55.6  &30.5      \\
{MAC (Ref)}~\cite{hudson2018compositional}    &50.6  &51.9  &33.3  &50.8  &25.2  \\
ALOE (Ref)~\cite{ding2020object}      &48.6  &51.2  &26.1  &52.9  &27.2       \\
\midrule
ALPRO~\cite{li2022align}  &47.1  &51.8  &28.9  &52.6  &28.4   \\
GPT-4o-mini~\cite{achiam2023gpt} &42.5  &50.0  &29.2  & \Scnd{58.8}  & \Scnd{30.7}       \\
Gemini~\cite{team2023gemini} &34.2  &50.3  &25.7  &49.4  &30.6     \\
\midrule
\textbf{PCR (ours)}  & \Frst{68.4}  & \Frst{58.3}  & \Frst{34.9}  & \Frst{60.3}  & \Frst{32.8}        \\
\midrule
Human Performance  &88.6  &82.9  &73.7  &88.2  &78.9      \\
\bottomrule
\end{tabular}
\end{center}
\vspace{-1em}
\caption{\revise{Evaluation of physical reasoning on \datasetNew. {Human performance is based on sampled questions. See the text for more details. \Frst{Red} text and \Scnd{blue} text indicate the first and second best results other than human performance.}}}
\vspace{-2em}
\label{tab:diverse}
\end{table}

\subsection{Evaluate Models on More Diverse Real-World Scenes}
\label{sec:appendreal}

\begin{figure*}[t]
    \centering
    \includegraphics[width=\textwidth,height=0.25\textwidth]{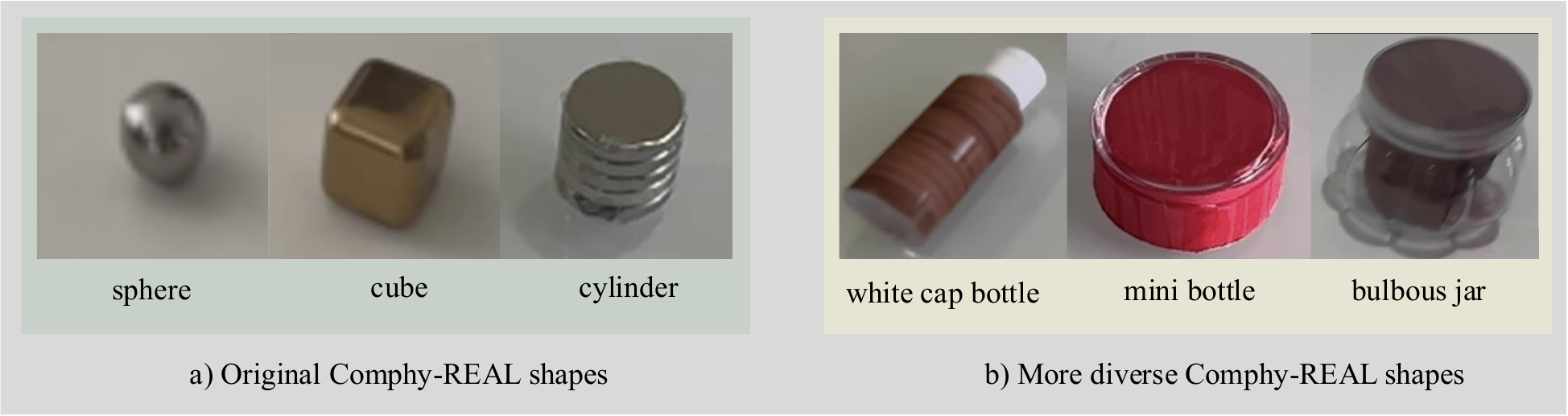}
    \vspace{-1em}
    \caption{\revise{Comparison of shape diversity between the original and extended \datasetNewReal. As shown in a), the three objects belong to the original dataset, whereas we added three more diverse ones in b) to \datasetNewReal.}}
    \label{fig:visual_real}
\end{figure*}

\noindent\textbf{{Enhanced Diversity of Real Physical Scenes.}}
\revise{
To evaluate models on more diverse real physical scenes, we significantly expanded the variety and complexity of real-world scenes in our revised dataset, \datasetNewReal. Specifically, we increased the object count from the original three to six distinct real-world objects, each varying significantly in shape and appearance, as illustrated in Figure~\ref{fig:visual_real}. Additionally, we manually altered the surface colors of these objects by applying different paint colors, thus further diversifying their visual appearances. To enrich the visual context, we applied object matting techniques to place these objects onto nine different realistic backgrounds featuring varied textures and lighting conditions.
}

\begin{figure*}[t]
    \centering
    \includegraphics[width=\textwidth]{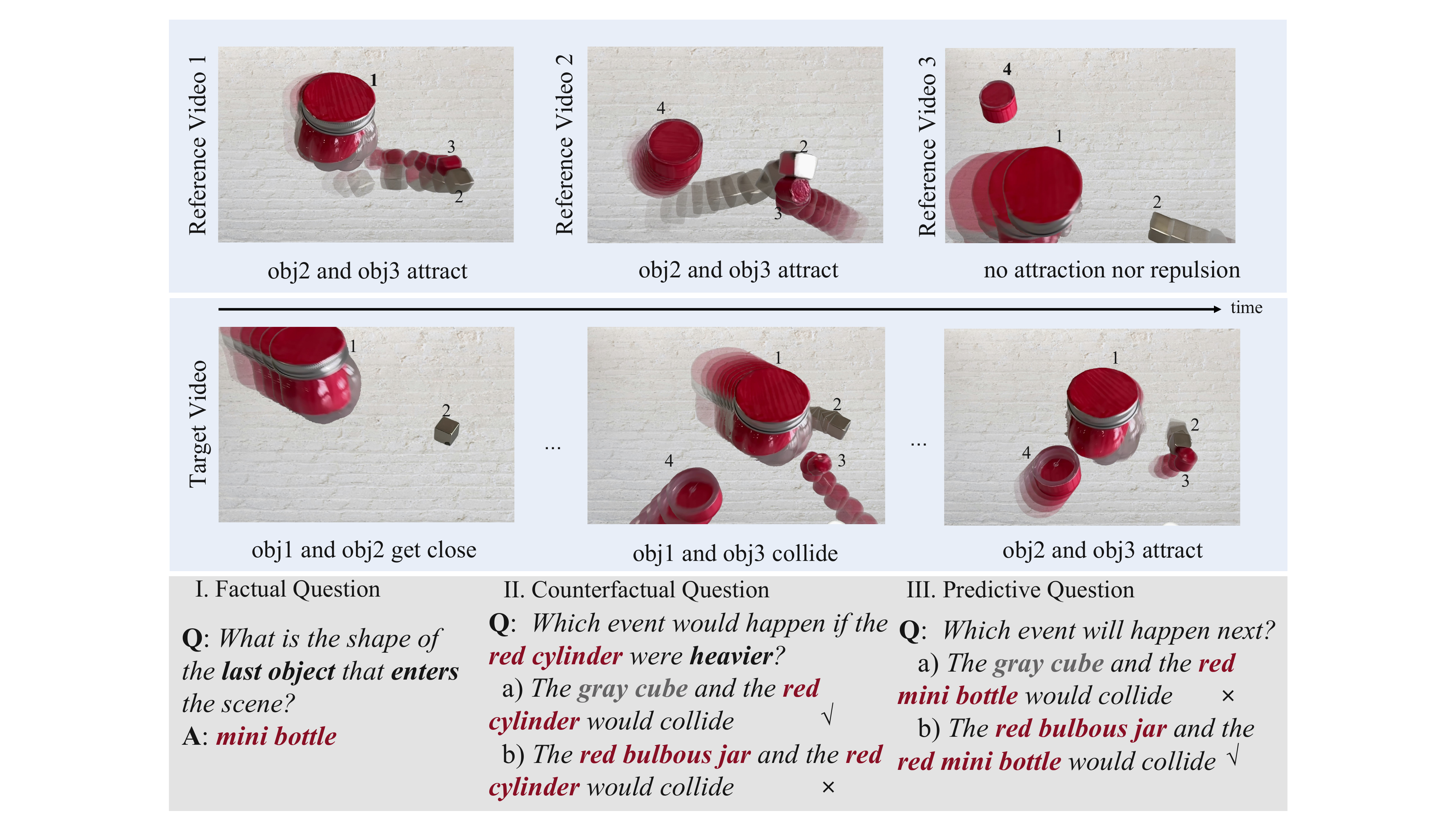}
    \caption{\revise{Qualitative examples of more diverse scenes in the \datasetNewReal. As shown in the figure, we have more diverse physical interactions between objects in the video. The image background is also more diverse with different textures and colors in contrast to the original \dataset in~\cite{chencomphy}.}}
    \label{fig:qual_real}
\end{figure*}

\begin{figure*}[t]
    \centering
    \includegraphics[width=\textwidth]{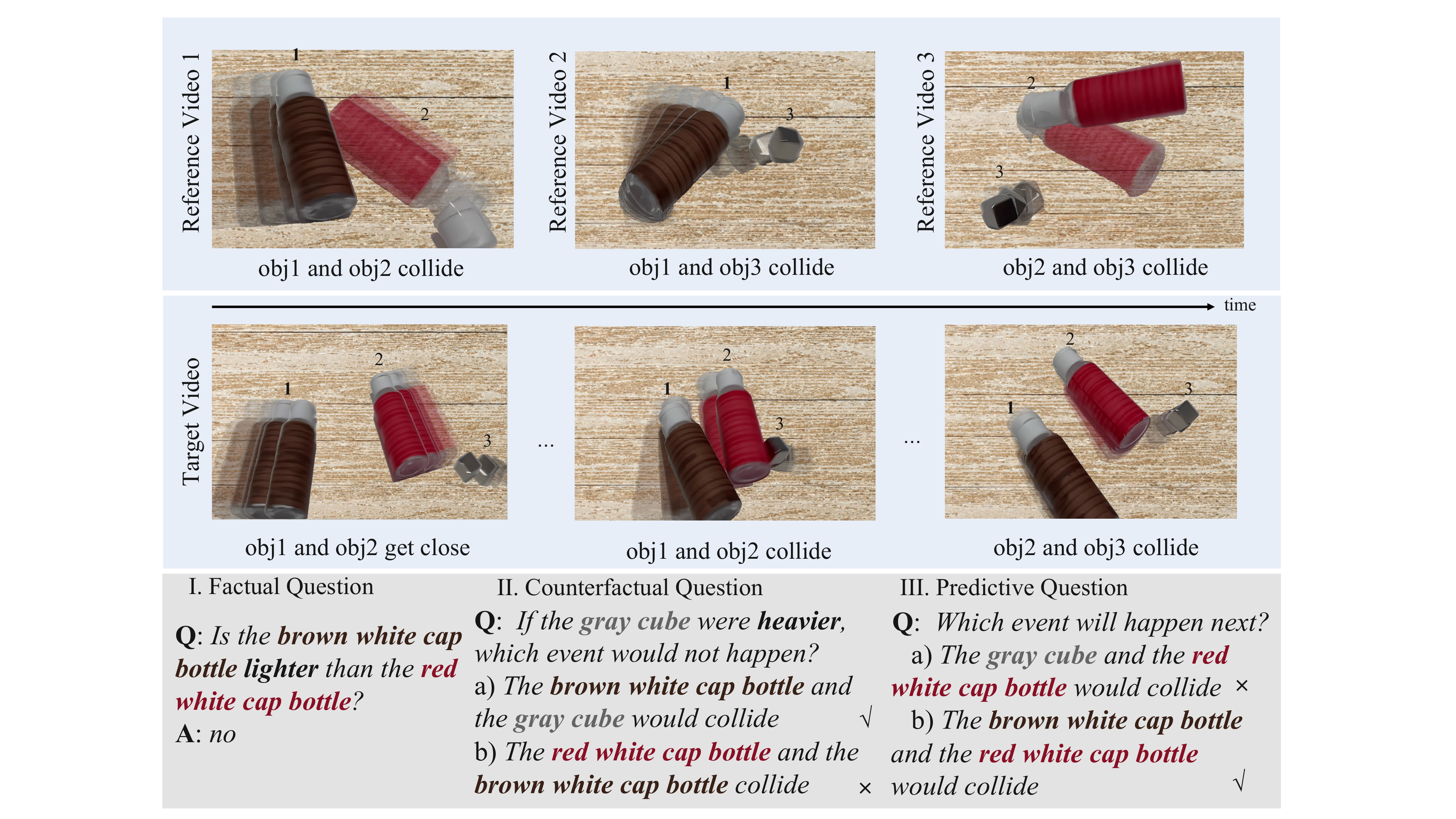}
    \caption{\revise{More qualitative examples of more diverse scenes in the \datasetNewReal.}}
    \label{fig:qual_real2}
\end{figure*}

\begin{figure*}[t]
    \centering
    \includegraphics[width=\textwidth]{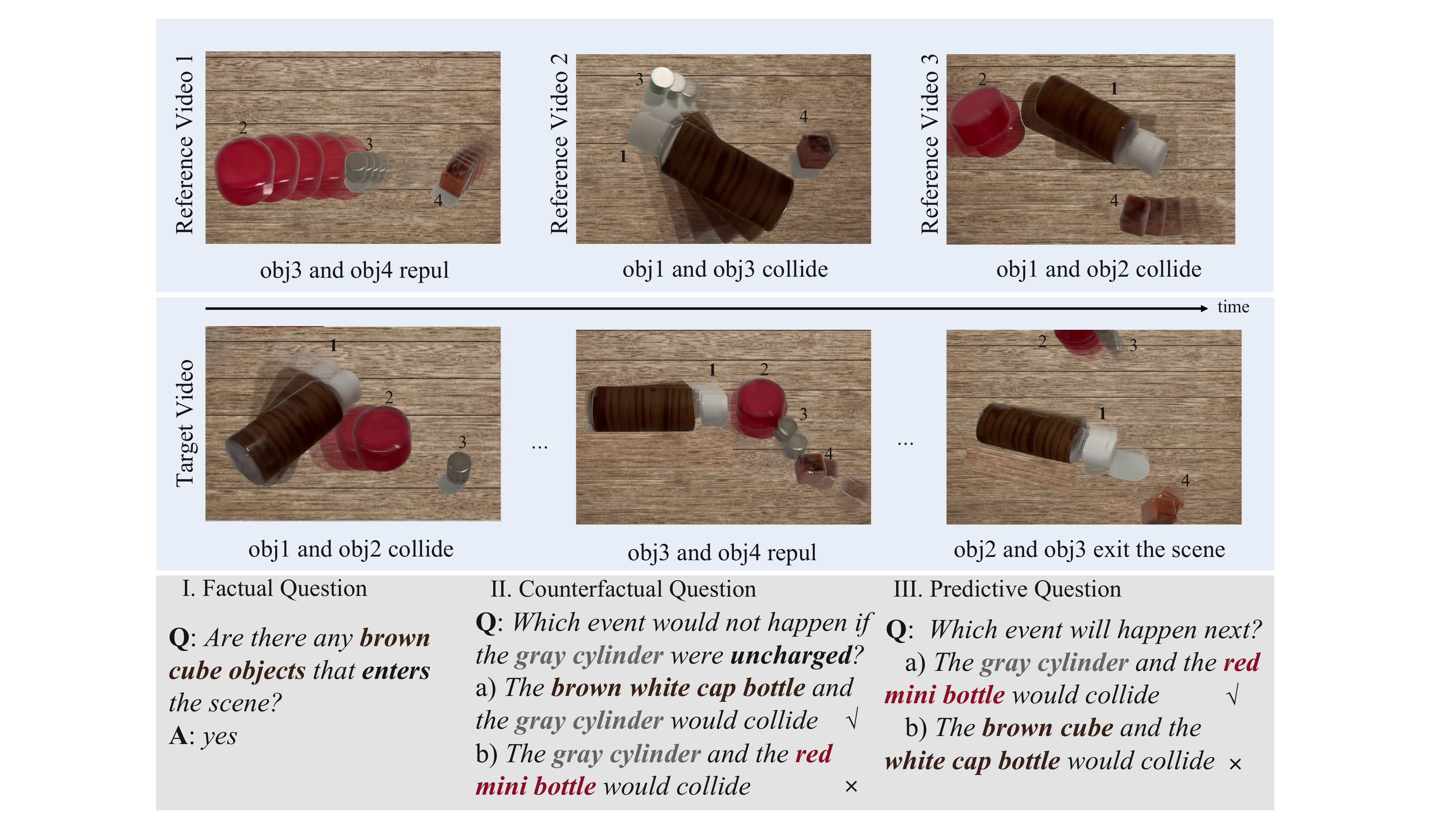}
    \caption{\revise{More qualitative examples of more diverse scenes in the \datasetNewReal.}}
    \label{fig:qual_real3}
\end{figure*}

\begin{figure*}[t]
    \centering
    \includegraphics[width=\textwidth]{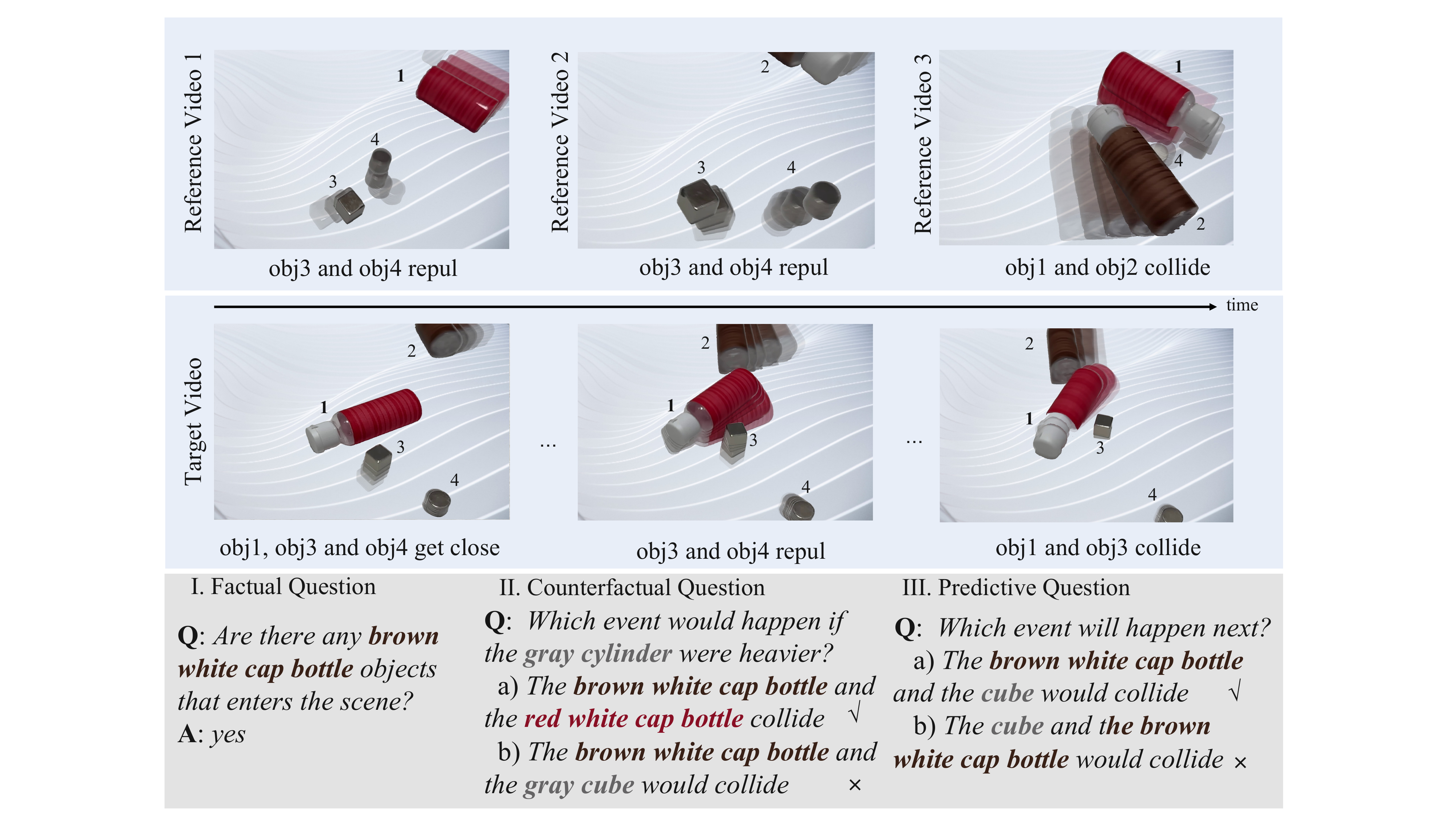}
    \caption{\revise{More qualitative examples of more diverse scenes in the \datasetNewReal.}}
    \label{fig:qual_real4}
\end{figure*}

\revise{
We acknowledge that collecting real-world data involves \textbf{substantial manual effort}, including carefully painting objects, initializing their positions and velocities, precisely segmenting objects from videos, and replacing backgrounds through matting. As a result of these efforts, our enhanced real-world dataset now comprises \textbf{123} distinct scene sets, yielding a total of \textbf{492} unique real-world videos. Figure~\ref{fig:qual_real} provides representative qualitative examples of these more diverse and realistic scenes, highlighting intricate physical interactions such as collisions and attraction events among multiple objects. Additional examples are presented in Figures~\ref{fig:qual_real2}--\ref{fig:qual_real4}.
}

\begin{table*}[t]
\begin{center}
\setlength{\tabcolsep}{8pt}
\begin{tabular}{llccccc}
\toprule
\multirow{2}{*}{Categories} & \multirow{2}{*}{Methods}  & \multirow{2}{*}{Factual} & \multicolumn{2}{c}{Predictive} &  \multicolumn{2}{c}{Counterfactual} \\
  &             &               & per opt.       & per ques.      & per opt.      & per ques.     \\
\midrule
\multirow{3}{*}{Bias analysis models} 
& Random      &7.6 &50.0 &25.0 &50.9 &20.8                 \\
& Frequent    &41.7 &53.6 &28.7 &50.0 &23.9                                  \\
& Blind-LSTM    &50.6  &61.5 &46.0 &51.9 &32.2            \\

\midrule
\multirow{2}{*}{video question answering  models} 
& CNN-LSTM~\cite{antol2015vqa}    &55.6  &64.2 &47.3 &50.9 &33.3           \\
& HCRN~\cite{le2020hierarchical}  &51.9  &62.5 &53.5 &50.9 &32.1               \\
\midrule
\multirow{2}{*}{Compositional reasoning
models} 
& MAC~\cite{hudson2018compositional}      &58.9  &60.9 & \Scnd{57.1} & \Scnd{52.8} & \Scnd{35.8}        \\
& ALOE~\cite{ding2020object}         &60.8  &60.6 &42.4 &47.1 &28.7             \\
\midrule
\multirow{3}{*}{Models with Reference Videos}
& {CNN-LSTM (Ref)}~\cite{antol2015vqa}      &49.0  & \Scnd{64.3} &41.3 &50.0 &26.3           \\
& {MAC (Ref)}~\cite{hudson2018compositional}    &56.4  &56.2 &46.4 &51.4 &34.9         \\
& ALOE (Ref)~\cite{ding2020object}      & \Scnd{61.6}  &61.4 &42.8 &51.6 &32.1          \\
\midrule

\multirow{4}{*}{Large Vision Language Models} 
& ALPRO~\cite{li2022align}  &50.9  &55.3 &39.2 &49.7 &29.2  \\
& GPT-4o-mini~\cite{achiam2023gpt} &42.6  &49.6 &23.2 &47.5 &26.0        \\
& Gemini~\cite{team2023gemini} &32.5  &57.7 &23.1 &52.1 &29.8   \\
\midrule
& \textbf{PCR(ours)}  & \Frst{63.5}  &\Frst{70.4} & \Frst{62.7} & \Frst{54.6} & \Frst{36.5}      \\
& {Human Performance}   &90.0  &95.0 &90.0 &94.4 &88.9     \\
\bottomrule
\end{tabular}
\end{center}
\vspace{-1em}
\caption{\revise{Evaluation of physical reasoning on \datasetNewReal. {Human performance is based on sampled questions. See the text for more details. \Frst{Red} text and \Scnd{blue} text indicate the first and the second best results other than human performance.}}}
\label{tb:real}
\end{table*}

\noindent\textbf{{New Experimental Results on Enhanced Real Scenes.}}
\revise{
To validate the increased complexity and diversity, we conducted extensive experiments using these newly collected real-world scenes. As reported in Table~\ref{tb:real}, our proposed model (\modelNew) consistently outperforms all baseline methods, demonstrating robustness and strong physical reasoning capabilities even when confronted with diverse and realistic data. Furthermore, we conducted an additional human evaluation study on this expanded dataset, revealing that human participants still achieve high accuracy, underscoring that although the dataset presents notable challenges for AI, it remains intuitive and manageable for humans.
}

\subsection{Discussion on integrating \modelNew with LVLMs}  
\label{sec:appenlvlm}
\revise{
We argue that it is quite promising to combine neuro-symbolic models like our \modelNew that learns neural modules for specific functions directly from the training question-answer pairs and the general capability of LVLMs. We think that LVLMs can at least help with the following aspects of the \modelNew framework, (1) improving the robustness of the language parsing capabilities; (2) enabling challenging commonsense reasoning that combines the outside knowledge from LVLMs and domain-specific knowledge; and (3) handling new tasks by cooperating with pre-trained modules and learned modules.
}

\noindent\textbf{(1). Improving Language Parsing Capabilities.}
\revise{
To improve the AI systems' capability to understand the language query, we can replace the language parser~\cite{yi2019clevrer,chen2021grounding} with shallow two-layer Seq2seq LSTMs~\cite{hochreiter1997long} with the LVLMs. One limitation for the previous shallow language parser is that it shows its limitations when transforming the language instructions with a new format into executable programs. And the capture the semantics of language is quite easy for LVLMs. Thus, we can use in-context learning to transform any language instructions into executable programs. To evaluate this capability by combining LVLMs and \modelNew, we first generate a new test set that contains much more diverse language instructions for the tasks in \dataset. Specifically, we follow a generate-verify strategy to synthesize diverse language instructions. We first use \texttt{Qwen/Qwen2.5-72B-Instruct-AWQ}~\cite{qwen2.5} to paraphrase the questions in \dataset and generate questions with diverse formats, but keep the same meaning of the original questions. We then ask the LLM to verify that the revised new question has the same semantic meaning as the original ones and abandon those questions without the same meaning. Sample questions are shown in Table~\ref{tab:revise} and the results of using LLMs to parse the question can be seen in Table~\ref{tab:parse}. From Table~\ref{tab:parse}, we can see that LVLMs can parse the language instruction into the programs better much better than the original program parser~\cite{yi2019clevrer,chen2021grounding}. To provide a quantitative evaluation, we revise the questions from the validation set of \dataset and evaluate the performance of the original \modelNew and \modelNew + LVLMs. To relieve the API cost, we use the Qwen(\texttt{Qwen/Qwen2.5-72B-Instruct-AWQ}) to serve as an alternative to LVLMs to parse the programs. The results are shown in Table~\ref{tab:revise}. We found that although both models still work on this revised set. Combining Qwen for robust program parsing, it performs much better on all types of questions from the dataset.
}

\begin{table*}[t]
  \centering
  \begin{tabular}{c|l}
    \toprule
    Original Question 1 
      & If the cyan sphere were heavier, what would not happen? \\
    \midrule
    \multirow{4}{*}{Revised Question 1} 
      & What would not occur if the cyan sphere had more weight?     \\
      & If the cyan sphere had more mass, what outcome would be impossible? \\
      & What would not occur if the cyan sphere were to have a greater weight? \\
      & What would not occur if the cyan sphere had more weight?      \\
    \midrule
    Original Question 2 
      & What will happen next?                                      \\
    \midrule
    \multirow{4}{*}{Revised Question 2} 
      & What is the next event that will take place?               \\
      & What is likely to happen next?                             \\
      & What is the next event that will occur?                    \\
      & What is expected to happen next?                           \\
    \bottomrule
  \end{tabular}
  \caption{\revise{Examples of revised questions that preserve the original semantics while exhibiting greater linguistic diversity and flexibility. These variations challenge language parsers~\cite{yi2019clevrer,chen2021grounding} by introducing textual patterns not encountered during training.}}
  \label{tab:revise}
\end{table*}

\begin{table*}[t]
  \centering
  \resizebox{\textwidth}{!}{%
  \begin{tabular}{ll}
    \toprule
    Question & What would be the outcome if the sphere had a greater \blue{mass}? \\
    \midrule
    PCR’s parser
      & all\_events, objects, sphere, filter\_shape, unique, \red{counterfact\_uncharged}, filter\_counterfact, belong\_to, \red{not} \\
    LVLMs
        & all\_events, objects, sphere, filter\_shape, unique, counterfact\_heavier, filter\_counterfact, belong\_to \\
    \midrule
    Question & What color is the metal sphere that remains stationary at the \blue{start} of the video? \\
    \midrule
    PCR’s parser
          & objects, metal, filter\_material, sphere, filter\_shape, \red{filter\_end}, query\_frame, filter\_stationary, query\_color \\
    LVLMs
          & objects, metal, filter\_material, sphere, filter\_shape, \red{filter\_start}, query\_frame, filter\_stationary, query\_color \\
    \bottomrule
  \end{tabular}
   }
  \caption{\revise{Comparison of parsing results between \modelNew’s program parser and LVLMs. \modelNew’s parser fails on the revised questions due to distribution shift from its original training set, whereas LVLMs succeed thanks to superior generalization. Key operators are highlighted in \red{red}.}}
  \label{tab:parse}
\end{table*}

\begin{table*}[thbp]
    \centering
    \begin{tabular}{lcccccc}
        \toprule
         \multirow{2}{*}{Methods}  & \multirow{2}{*}{Factual} & \multicolumn{2}{c}{Predictive} &  \multicolumn{2}{c}{Counterfactual} \\
          &               & per opt.       & per ques.      & per opt.      & per ques.     \\
        \midrule
        \modelNew & 51.9 & 60.6 & 44.2 & 57.4 & 41.5  \\
        PCR+LVLMs & 69.7 & 73.5 & 54.0 & 76.3 & 60.9  \\
        \bottomrule
    \end{tabular}
    \caption{\revise{Performance Comparison of \modelNew and PCR+LVLMs on revised questions, where \modelNew fails under a distribution shift from its training set, whereas PCR+LVLMs succeed thanks to superior generalization.}}
    \label{tab:metrics_transposed}
\end{table*}

\begin{figure*}
    \centering
    \includegraphics[width=\textwidth]{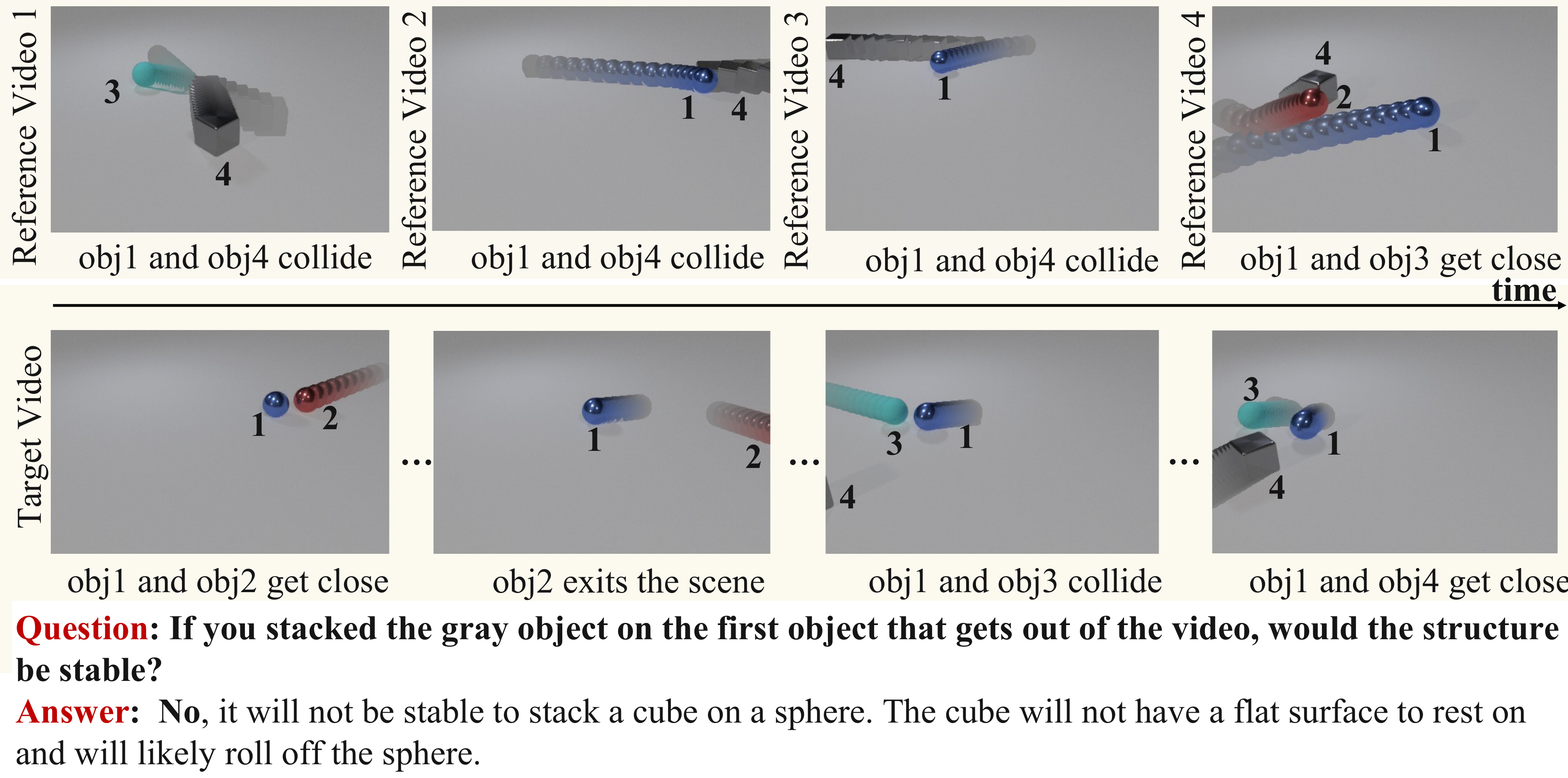}
    \vspace{-1em}
    \caption{\revise{An example of combining the strength of \modelNew and LVLMs to enable new commonsense reasoning capabilities. The LVLM is able to write a program in Python (Figure~\ref{fig:code1}) that calls the reasoning modules (\texttt{get\_color} and \texttt{filter\_out}) in \modelNew and (\texttt{llm\_query}) from LVLMs to handle the problem and provides the correct answer with explanation.}}
    \label{fig:newcap}
 \begin{minipage}[t]{1\linewidth}
        \begin{minted}[
        frame=lines,
        framesep=2mm,
        fontsize=\footnotesize,
        bgcolor=Gray,
        numbersep=5pt,
        escapeinside=||,
        linenos
        ]{python}
def execute_command(video, possible_answers, query, ImagePatch,
            VideoSegment, llm_query, bool_to_yesno, distance, best_image_match):
    video_segment = VideoSegment(video)
    num_objects = video_segment.count_objects()
    # Find the first object that gets out of the scene
    out_list = []
    for idx in range(num_objects):
        out_frm = video_segment.filter_out(idx)
        if out_frm is not None:
            out_list.append([idx, out_frm])
    if len(out_list) == 0:
        return "There is no object that exits the scene"
    out_list = sorted(out_list, key=lambda x: x[1])
    first_out_idx = out_list[0][0]
    # Find the gray object
    gray_obj_idx = None
    for idx in range(num_objects):
        color = video_segment.get_color(idx)
        if color == 'gray':
            gray_obj_idx = idx
            break
    if gray_obj_idx is None:
        return "There is no gray object in the video"
    # Get shapes of the gray object and the first object that gets out
    gray_shape = video_segment.get_shape(gray_obj_idx)
    first_out_shape = video_segment.get_shape(first_out_idx)
    # Use llm_query to determine stability
    answer = llm_query(f"Will it be stable to stack a {gray_shape} on 
                a {first_out_shape}?")
    return answer
     \end{minted}
    \centering
\end{minipage}
     \caption{\revise{The program that the LVLM generates to handle the query in Figure~\ref{fig:newcap}. The program first calls the modules (\texttt{get\_color} and \texttt{filter\_out}) in \modelNew to identify the object 4 and the object 2 in the video. The program that calls the \texttt{get\_shape} module in \modelNew to get the objects' shape and finally sends the LVLM a question based on the shape to identify the stability of the structure and gives the explanation (the \textbf{\red{answer}} in Figure~\ref{fig:newcap}).}}
    \label{fig:code1}
\end{figure*}

\noindent\textbf{(2). Enabling New Commonsense Reasoning Capabilities.}
\revise{
By cooperating with the \modelNew with LVLMs, we are able to answer questions that require commonsense knowledge that does not exist in the original \modelNew's training set. For example, as shown in Figure~\ref{fig:newcap}, when we ask the model \modelNew+LVLMs the question, \textit{``If you stacked the gray object on the first object gets out of the video, would the structure be stable?"}, the LVLM (specially, GPT4-o in this example) is able to write a program in Python (Figure~\ref{fig:code1}) that calls the reasoning modules (\texttt{get\_color} and \texttt{filter\_out}) in \modelNew and (\texttt{llm\_query}) from LVLMs to handle the problem and provides the correct answer with explanation (\textit{``No, it will not be stable to stack a cube on a sphere. The cube will not have a flat surface to rest on and will likely roll off the sphere."}). Note that either \modelNew or GPT4-o alone is not able to solve this task. \modelNew can not transform such an out-of-domain question query into an executable Python program (Figure~\ref{fig:code1}) and does not have the commonsense to know the outcome of stacking a cube on a sphere. When adopting GPT4-o alone, we can not distinguish the fine-grained details in the video and might miss the frame where the first object that gets out of the scene from only a few frames.
}

\begin{figure*}
    \centering
    \includegraphics[width=\textwidth]{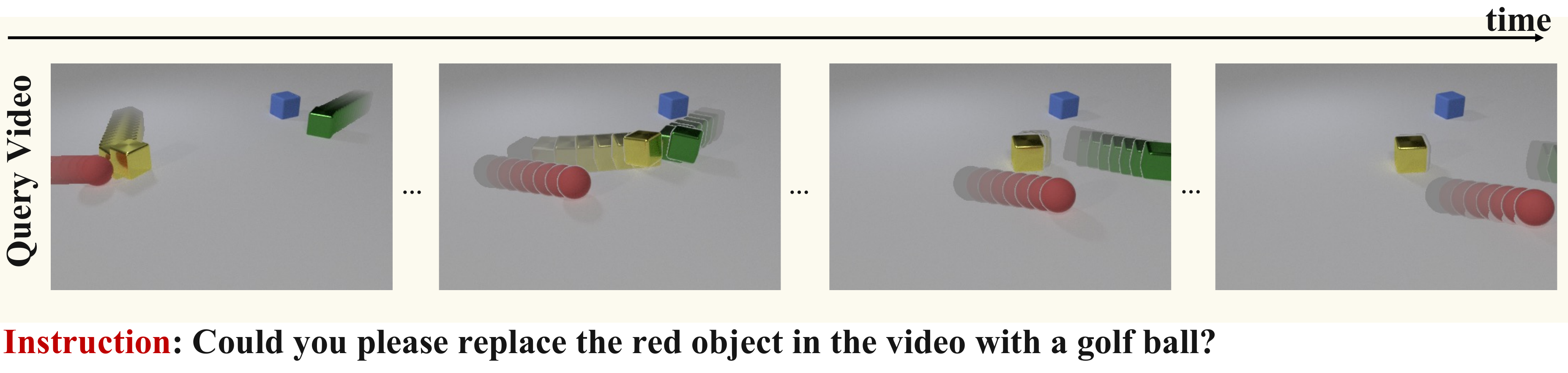}
    \vspace{-1em}
    \caption{\revise{An input video and a language instruction to ask AI models to edit the target video.}}
    \label{fig:video1}
 \begin{minipage}[t]{1\linewidth}
        \begin{minted}[
        frame=lines,
        framesep=2mm,
        fontsize=\footnotesize,
        bgcolor=Gray,
        numbersep=5pt,
        escapeinside=||,
        linenos
        ]{python}
def execute_command(video) -> list:
    video_segment = VideoSegment(video)
    num_objects = video_segment.count_objects()
    red_obj_idx = None
    for idx in range(num_objects):
        color = video_segment.get_color(idx)
        if color == 'red':
            red_obj_idx = idx
            break
    if red_obj_idx is not None:
        edited_images = video_segment.edit_objects(red_obj_idx, "golf ball")
        return edited_images
    return []
     \end{minted}
    \centering
\end{minipage}
    \vspace{-1em}
    \caption{\revise{Program generated by the LVLM to address the query in Figure~\ref{fig:video1}. It first uses the \texttt{get\_color} module to locate the red object, then applies the Stable Diffusion–based \texttt{edit\_objects} module to transform that region into a \textit{golf ball}. Integrating \modelNew with LVLMs empowers the model to incorporate new modules for new tasks such as video editing.
    \label{fig:code2}
    }}
    \centering
    \includegraphics[width=\textwidth]{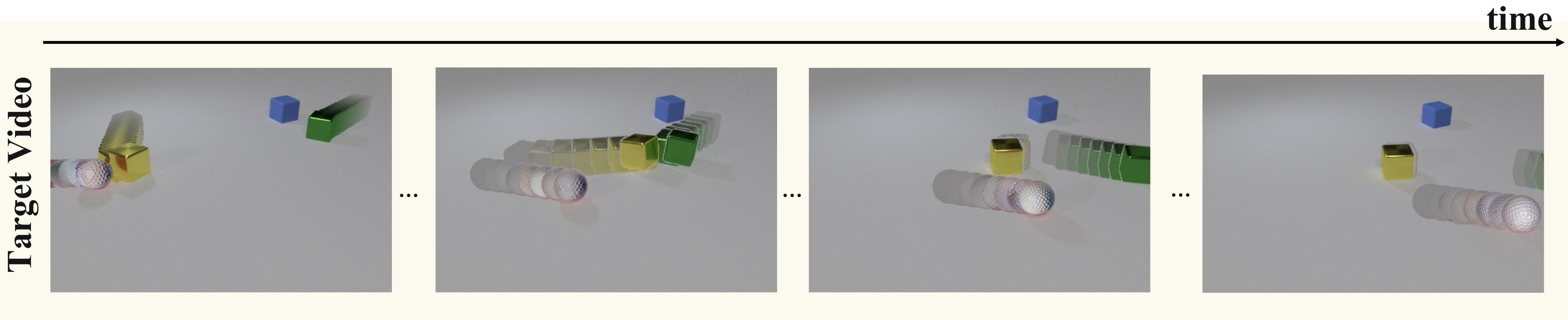}
    \vspace{-1em}
    \caption{\revise{The output video of replacing the \red{red} object with a \textbf{golf ball} by calling the new stable diffusion module (\texttt{edit\_object})~\cite{Rombach_2022_CVPR} to edit the target object region.}}
    \label{fig:video2}
\end{figure*}

\revise{
\noindent\textbf{(3). Handling New Tasks with Modules beyond \modelNew and \dataset.}
Another benefit of LVLMs is that they can be used as a controller to control both the modules in \modelNew and other modules that are learned from other datasets and tasks. As shown in Figure~\ref{fig:video1}, Figure~\ref{fig:code2} and Figure~\ref{fig:video2}, we show how we can achieve the goal of fine-grained video editing by combining \modelNew and LVLMs.
The LVLM first parses the question into an executable python program (Figure \ref{fig:code2}) that calls neural modules from \modelNew (\texttt{get\_color}) to identify the target object and adopts the existing diffusion model module~\cite{Rombach_2022_CVPR} (\texttt{edit\_objects}) to perform fine-grained edits.
}

\begin{figure*}[t]
    \centering
    \includegraphics[width =0.8\textwidth]{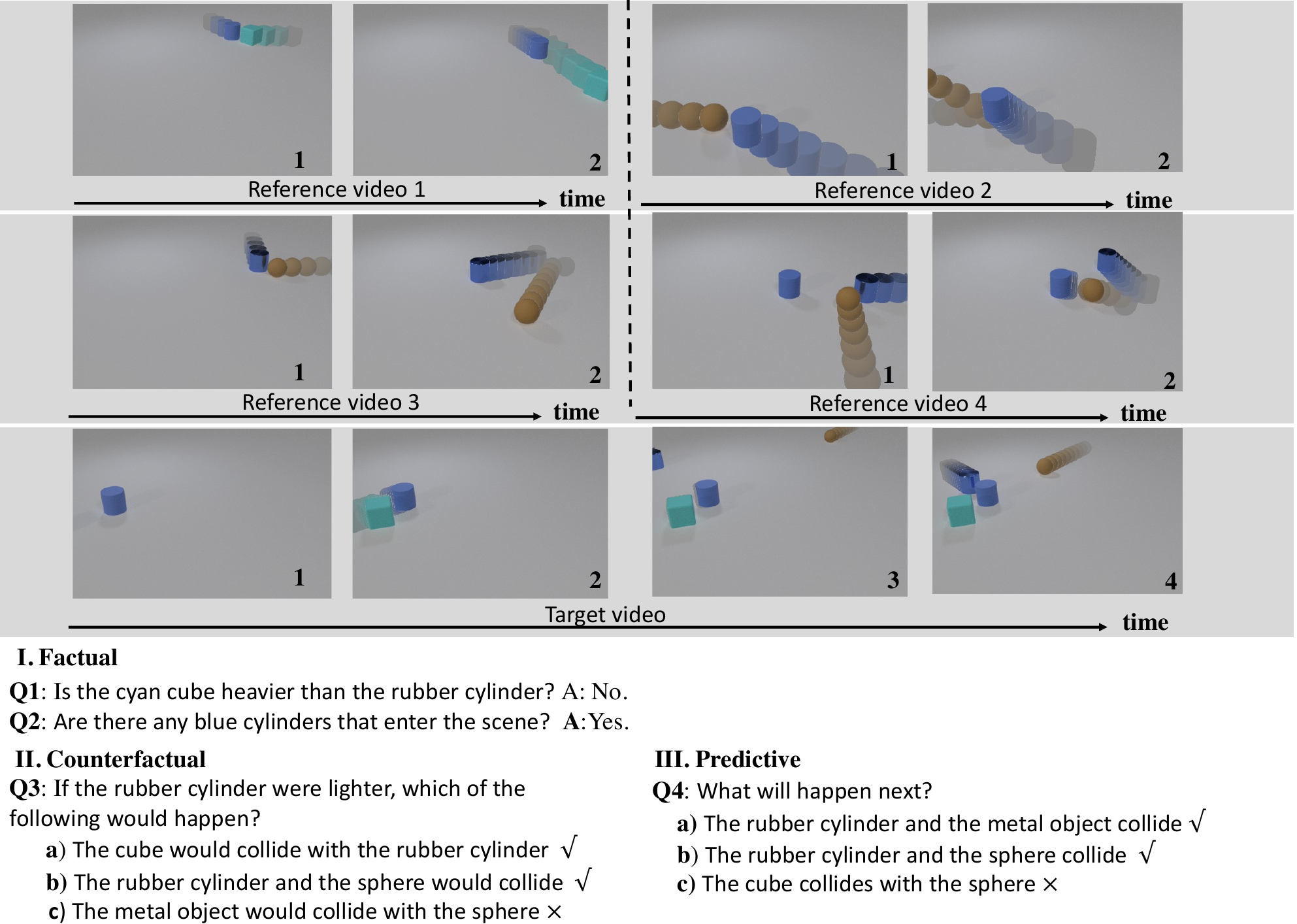}
    \label{fig:data1}
\end{figure*}

\begin{figure*}[t]
    \centering
    \includegraphics[width =0.8\textwidth]{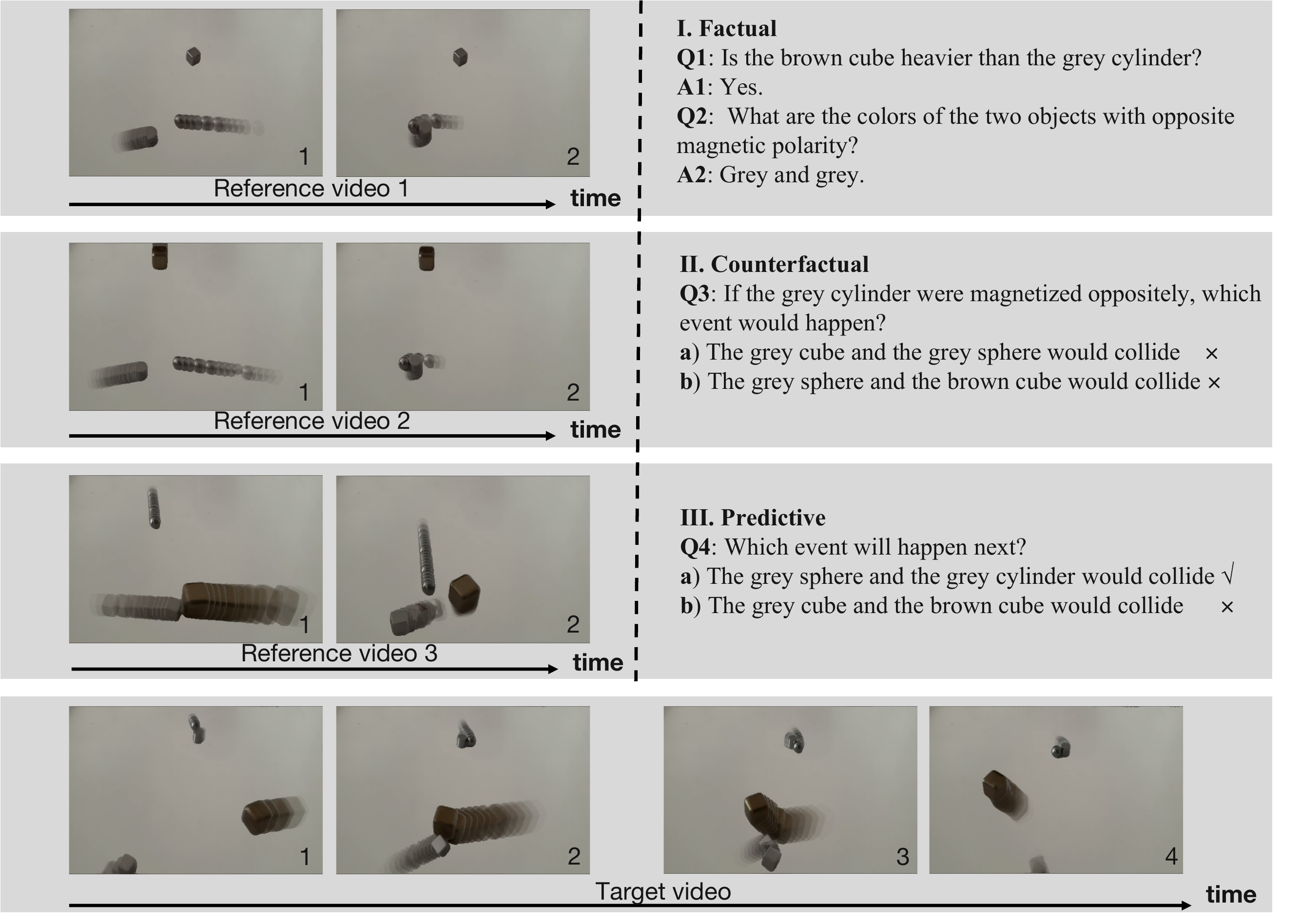}
    \label{fig:real1}
\end{figure*}

\end{document}